\documentclass[journal]{IEEEtran}
\ifCLASSINFOpdf
\else
\fi
\usepackage{lineno,hyperref,amssymb}
\usepackage{graphicx}
\usepackage{algorithm}
\usepackage{algpseudocode}
\usepackage{threeparttable}
\usepackage{subfig}
\usepackage{subfloat}
\usepackage{hyperref}[colorlinks,linkcolor=blue]
\usepackage{listings}
\usepackage{makecell}
\usepackage{color}
\usepackage{booktabs}
\usepackage{amsmath}
\usepackage{amssymb}
\begin{document}
%
\title{BrainCog: A Spiking Neural Network based Brain-inspired Cognitive Intelligence Engine for Brain-inspired AI and Brain Simulation}
%
%
%

%
%
\author{Yi Zeng, Dongcheng Zhao, Feifei Zhao, Guobin Shen, Yiting Dong, Enmeng Lu, Qian Zhang, Yinqian Sun, Qian Liang, Yuxuan Zhao, Zhuoya Zhao, Hongjian Fang, Yuwei Wang, Yang Li, Xin Liu, Chengcheng Du, Qingqun Kong, Zizhe Ruan, Weida Bi
   
\thanks{Yi Zeng is with the Research Center for Brain-inspired Intelligence, the National Laboratory of Pattern Recognition, at the Institute of Automation, Chinese Academy of Sciences, Beijing 100190, China, and University of Chinese Academy of Sciences, Beijing 100049, China, and Center for Excellence in Brain Science and Intelligence Technology, Chinese Academy of Sciences, Shanghai 200031, China. (e-mail: yi.zeng@ia.ac.cn, yi.zeng@braincog.ai).}   
 
\thanks{Dongcheng Zhao, Feifei Zhao, Enmeng Lu, Qian Liang, Yuxuan Zhao, Yuwei Wang, Xin Liu, Zizhe Ruan, and Weida Bi are with the Research Center for Brain-inspired Intelligence, Institute of Automation, Chinese Academy of Sciences, Beijing 100190, China. (e-mail: dongcheng.zhao@braincog.ai, feifei.zhao@braincog.ai)}

\thanks{Guobin Shen, Yiting Dong, Yinqian Sun, Zhuoya Zhao, Hongjian Fang, and Qingqun Kong are with the Research Center for Brain-Inspired Intelligence,  Institute of Automation, Chinese Academy of Sciences, Beijing 100190, China, and also with the School of Future Technology, University of Chinese Academy of Sciences, Beijing 100049, China. (e-mail: guobin.shen@braincog.ai, yiting.dong@braincog.ai)}

\thanks{Qian Zhang, Yang Li and Chengcheng Du are with the Research Center for Brain-Inspired Intelligence,  Institute of Automation, Chinese Academy of Sciences, Beijing 100190, China, and also with the School of Artificial Intelligence, University of Chinese Academy of Sciences, Beijing 100049, China.}

\thanks{Yi Zeng, Dongcheng Zhao, Feifei Zhao, Guobin Shen, and Yiting Dong contributed equally to this work, and serve as co-first authors.}
\thanks{The corresponding author is Yi Zeng.}
}

%



\maketitle


%
\begin{abstract}
Spiking neural networks (SNNs) have attracted extensive attentions in Brain-inspired Artificial Intelligence and computational neuroscience. They can be used to simulate biological information processing in the brain from multiple scales, including membrane potential, neuronal firing, synaptic transmission, synaptic plasticity, and multiple brain areas coordination. More importantly, SNNs serve as an appropriate level of abstraction to bring inspirations from brain and cognition to Artificial Intelligence. Existing software frameworks support SNNs in machine learning, brain simulations, and specific hardware devices from certain perspectives respectively. However, the community requires an open-source platform that can support building and integrating computational models for brain-inspired AI and brain simulation at multiple scales. In this paper, we present the Brain-inspired Cognitive Intelligence Engine (BrainCog) for creating brain-inspired AI and brain simulation models. BrainCog incorporates different types of spiking neuron models, learning rules, brain areas, etc., as essential modules provided by the platform. Based on these easy-to-use modules, BrainCog supports various brain-inspired cognitive functions, including Perception and Learning, Decision Making, Knowledge Representation and Reasoning, Motor Control, and Social Cognition. These brain-inspired AI models have been effectively validated on various supervised, unsupervised, and reinforcement learning tasks, and they can be used to enable AI models to be with multiple brain-inspired cognitive functions. For brain simulation, BrainCog realizes the function simulation of \emph{Drosophila} decision-making and prefrontal cortex working memory, the structure simulation of the Neural Circuit (Microcircuit and Cortical Column), and whole brain structure simulation of Mouse brain, Macaque brain, and Human brain. An AI engine named BORN is developed based on BrainCog, and it demonstrates how the components of BrainCog can be integrated and used to build AI models and applications. To enable the scientific quest to decode the nature of biological intelligence and create Artificial Intelligence, BrainCog aims to provide necessary, easy-to-use, and essential building blocks, and infrastructural support to develop brain-inspired spiking neural network based Artificial Intelligence, and to simulate the cognitive brains at multiple scales. The online repository of BrainCog can be found at  \href{http://www.brain-cog.network}{http://www.brain-cog.network}.

\end{abstract}

\IEEEpeerreviewmaketitle

\section{Introduction}
The human brain can self-organize and coordinate different cognitive functions to flexibly adapt to complex and changing environments. A major challenge for Artificial Intelligence and computational neuroscience is integrating multi-scale biological principles to build biologically plausible brain-inspired intelligent models. As the third generation of neural networks~\cite{Maass1997Networks}, Spiking Neural Networks (SNNs) are more biologically realistic at multiple scales, more biologically interpretable, more energy-efficient, and naturally more suitable for modeling various cognitive functions of the brain and creating biologically plausible AI. 

\begin{figure*}[!htbp]
	\centering
	\includegraphics[scale=0.3]{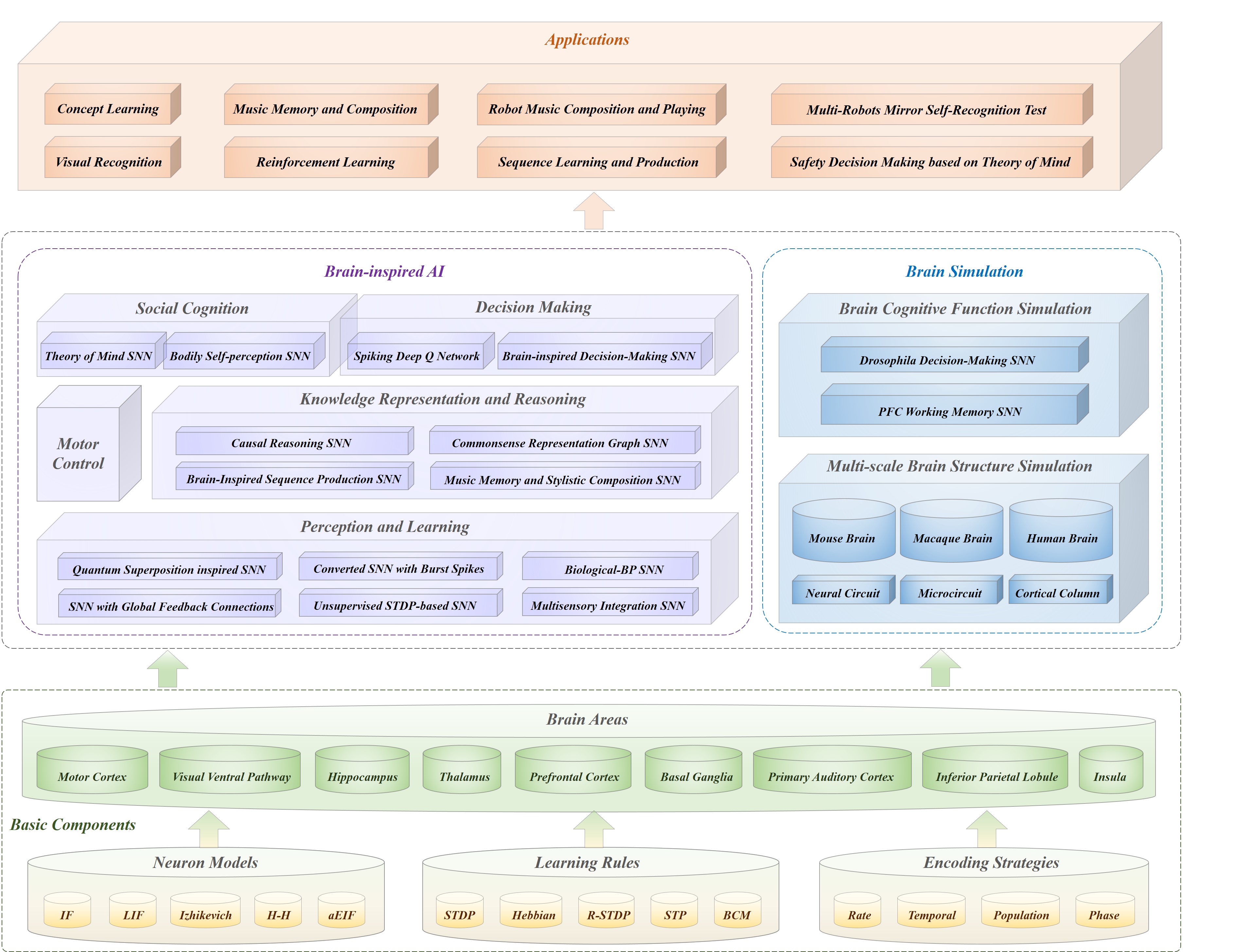}
	\caption{The architecture of Brain-inspired Cognitive Intelligence Engine (BrainCog).}
	\label{braincog}
\end{figure*}

Existing neural simulators attempted to simulate elaborate biological neuron models or build large-scale neural network simulations, neural dynamics models, deep SNNs. Neuron~\cite{carnevale2006neuron} focuses on simulating elaborate biological neuron models. NEST~\cite{gewaltig2007nest} implements large-scale neural network simulations. Brian/Brian2~\cite{stimberg2019brian, goodman2009brian} provides an efficient and convenient tool for modeling spiking neural networks. Shallow SNNs implemented by Brian2 can realize unsupervised visual classification~\cite{diehl2015unsupervised}. Further, BindsNET~\cite{ hazan2018bindsnet} builds SNNs with coordination of various neurons and connections and incorporates multiple biological learning rules for training SNNs. Such learning SNNs can perform machine learning tasks, including simple supervised, unsupervised, and reinforcement learning. However, supporting more complex tasks would be great challenge for these SNNs frameworks, and there is a large gap in performance compared with traditional deep neural networks (DNNs). Deep SNNs trained by surrogate gradient or converting well-trained DNNs to SNNs have achieved great progress in the fields of speech recognition~\cite{dominguez2018deep, loiselle2005exploration}, computer vision~\cite{kim2020spiking, wu2018spatio}, and reinforcement learning~\cite{tan2021strategy}. Motivated by this, SpikingJelly~\cite{SpikingJelly} develops a deep learning SNN framework (trained by surrogate gradient or converting well-trained DNNs to SNNs), which integrates deep convolutional SNNs and various deep reinforcement learning SNNs for multiple benchmark tasks. Platforms as SpikingJelly are relatively more inspired by the field of deep learning and focuse on improving the performance of different tasks. They are currently lack of in-depth inspiration from the brain information processing mechanisms and may not aim at, hence short at simulating large scale functional brains.

BrainPy~\cite{Wang2021} excels at modeling, simulating, and analyzing the dynamics of brain-inspired neural networks from multiple perspectives, including neurons, synapses, and networks. While it focuses on computational neuroscience research, it fails to consider the learning and optimization of deep SNNs or the implementation of brain-inspired functions. SPAUN~\cite{eliasmith2012large}, a large-scale brain function model consisting of 2.5 million simulated neurons and implemented by Nengo~\cite{bekolay2014nengo}, integrates multiple brain areas and realizes multiple brain cognitive functions, including image recognition, working memory, question answering, reinforcement learning, and fluid reasoning. However, it is not suitable for solving challenging and complex AI tasks that can be handled by deep learning models. In summary, there is still a lack of open-source spiking neural network frameworks that could incorporate the ability of simulating brain structures, cognitive functions at large scale, while keep itself effective for creating complex and efficient AI models at the same time.

Considering the various limitations of existing frameworks mentioned above, in this paper, we present the Brain-inspired Cognitive Intelligence Engine (BrainCog), a spiking neural network based open source platform for both brain-inspired AI and brain simulation at multiple scales. As shown in Fig.~\ref{braincog}, some basic modules (such as different types of neuron models, learning rules, encoding strategies, etc.) are provided as building blocks to construct different brain areas and neural circuits to implement brain-inspired cognitive functions. BrainCog is an easy-to-use framework that can flexibly replace different components according to various purposes and needs. BrainCog tries to achieve the vision ``the structure and mechanism are inspired by the brain, and the cognitive behaviors are similar to humans'' for Brain-inspired AI~\cite{2016Retrospect}. BrainCog is developed based on the deep learning framework (currently it is based on PyTorch, while it is easy to migrate to other frameworks such as PaddlePaddle, TensorFlow, etc.). This approach aims at greatly facilitating researchers to quickly familiarize themselves with the platform and implement their own algorithms.

\subsection{Brain-inspired AI}
\begin{figure}[!htbp]
	\centering
	\includegraphics[scale=0.35]{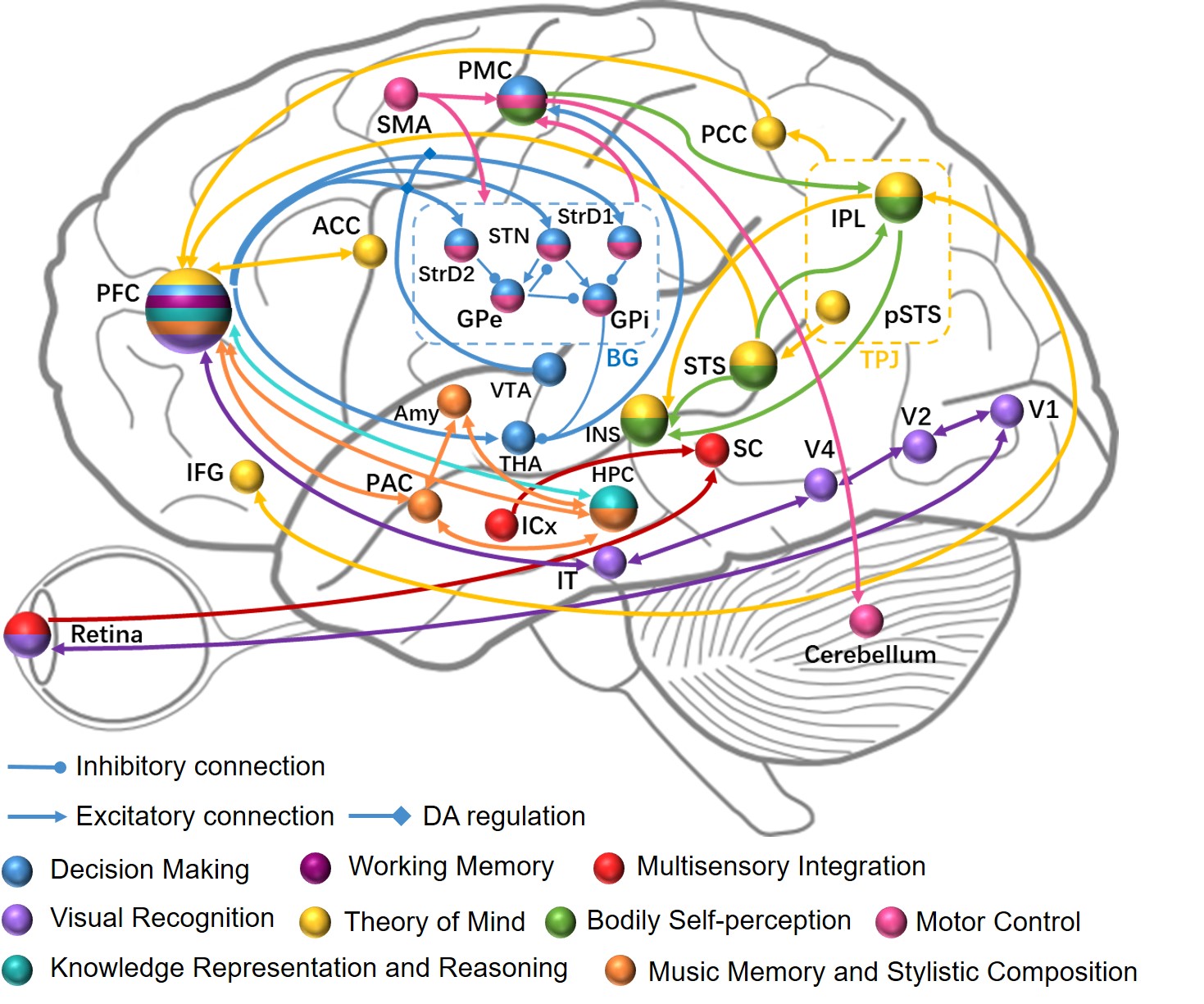}
	\caption{Multiple cognitive functions integrated in BrainCog, their related brain areas and neural circuits.}
	\label{born}
\end{figure}

BrainCog is aimed at providing infrastructural support for Brain-inspired AI. Currently, it provides cognitive functions components that can be classified into five categories: Perception and Learning, Decision Making, Motor Control, Knowledge Representation and Reasoning, and Social Cognition. These components collectively form neural circuits corresponding to 28 brain areas in the mammalian brains, as shown in Fig.~\ref{born}. These brain-inspired AI models have been effectively validated on various supervised and unsupervised learning, deep reinforcement learning, and several complex brain-inspired cognitive tasks.

\subsubsection{Perception and Learning} 
BrainCog provides a variety of supervised and unsupervised methods for training spiking neural networks, such as the biologically-plausible Spike-Timing-Dependent Plasticity (STDP)~\cite{bi1998synaptic}, the backpropagation based on surrogate gradients~\cite{wuSpatioTemporalBackpropagationTraining2018,zhengGoingDeeperDirectlyTrained2021,fangIncorporatingLearnableMembrane2021}, and the conversion-based algorithms~\cite{li2021free, han2020deep, han2020rmp}. In addition to high performance in common perception and learning process, it also shows strong adaptability in small samples and noisy scenarios. BrainCog also provides a multi-sensory integration framework for human-like concept learning~\cite{wywFramework}. Inspired by quantum information theory, BrainCog provides a quantum superposition spiking neural network, which encodes complement information to neuronal spike trains with different frequency and phase~\cite{SUN2021102880}. 

\subsubsection{Decision Making}
For decision making, BrainCog provides multi-brain areas coordinated decision-making spiking neural network~\cite{zhao2018brain}. The biologically inspired decision-making model implemented by BrainCog achieves human-like learning ability on the Flappy Bird game and supports UAVs' online decision-making tasks. In addition, BrainCog combines SNNs with deep reinforcement learning and provides the brain-inspired spiking deep Q network~\cite{sun2022solving}.

\subsubsection{Motor Control}
Embodied cognition is crucial to realizing biologically plausible AI. As part of the embodied cognition modules, and inspired by the motor control mechanism of the brain, BrainCog provides a multi-brain areas coordinated robot motion spiking neural networks model, which includes premotor cortex (PMC), supplementary motor areas (SMA), basal ganglia and cerebellum functions. With proper mapping, the spiking motor network outputs can be used to control various robots.

\subsubsection{Knowledge Representation and Reasoning} 
BrainCog incorporates multiple neuroplasticity and population coding mechanisms for knowledge representation and reasoning. The brain-inspired music memory and stylistic composition model implements the knowledge representation and memory of note sequences and can generate music according to different styles~\cite{LQ2020,LQ2021}. Sequence Production Spiking Neural Network (SPSNN) achieves the memory of the symbol sequence and can reconstruct the symbol sequence in the light of different rules~\cite{fang2021spsnn}. Commonsense Knowledge Representation Graph SNN (CKR-GSNN) realizes the representation of commonsense knowledge through incorporating multi-scale neural plasticity and population coding mechanism into a graph SNN model~\cite{KRRfang2022}.
Causal Reasoning Spiking Neural Network (CRSNN) encodes the causal graph into a spiking neural network and realizes deductive reasoning tasks accordingly~\cite{fang2021crsnn}.

\subsubsection{Social Cognition} 
BrainCog provides a brain-inspired social cognition model with biological plausibility. This model gives the agent a preliminary ability to perceive and understand itself and others and can enable the robots pass the Multi-Robots Mirror Self-Recognition Test~\cite{zeng2018toward} and the AI Safety Risks Experiment~\cite{zhao2022brain}. The former is a classic experiment of self-perception in social cognition, and the latter is a variation and application of the theory of mind experiment in social cognition.
\subsection{Brain Simulation}

\subsubsection{Brain Cognitive Function Simulation}
To demonstrate the capability of BrainCog for cognitive function simulation, we provide \emph{Drosophila} decision-making and prefrontal cortex working memory function simulation~\cite{zhao2020neural,zhangqian2021comparison}. For \emph{Drosophila} nonlinear and linear decision-making simulation, BrainCog verifies the winner-takes-all behaviors of the nonlinear dopaminergic neuron‑GABAergic neuron‑mushroom body (DA‑GABA‑MB) circuit under a dilemma and obtains consistent conclusions with \emph{Drosophila} biological experiments~\cite{zhao2020neural}. For the working memory performance of the prefrontal cortex network implemented by BrainCog, we discover that using human neurons to replace rodent neurons without changing network structure can significantly improve the accuracy and completeness of an image memory task~\cite{zhangqian2021comparison}, and this implies the evolution of the brains are not only on their structures, but also applies to single computational units such as neurons.

\subsubsection{Multi-scale Brain Structure Simulation} 
BrainCog provides simulations of brain structures at different scales, from microcircuits, and cortical columns, to whole-brain structure simulations. Anatomical and imaging multi-scale connectivity data is used to support whole-brain simulations from mouse brain, macaque brain to human brain at different scales.


\section{Basic Components}
BrainCog provides essential and fundamental components to model biological and artificial intelligence. It includes various biological neuron models, learning rules, encoding strategies, and models of different brain areas. One can build brain-inspired SNN models through reusing and refining these building blocks. Expanding and refining the components and cognitive functions included in BrainCog is an ongoing effort. We believe this should be a continuous community effort, and we welcome researchers and practitioners to enrich and improve the work together in a complementary way.

\subsection{Neuron Models}
BrainCog supports various models for spiking neurons, such as the following :

(1) Integrate-and-Fire spiking neuron (IF)~\cite{abbott1999lapicque}:

\begin{equation}
\frac{dV}{dt} = I
\end{equation}

$I$ denotes the input current from the pre-synaptic neurons. Once the membrane potential reaches the threshold $V_{th}$, the neuron $j$ fires a spike~\cite{abbott1999lapicque}.

(2) Leaky Integrate-and-Fire spiking neuron (LIF)~\cite{dayan2005theoretical}:
\begin{equation}
 \tau\frac{dV}{dt} = - V + I
\label{equa_lif1}
\end{equation}
$\tau=RC$ denote the time constant, $R$ and $C$ denotes the membrane resistance and capacitance respectively~\cite{dayan2005theoretical}.  

(3) The adaptive Exponential Integrate-and-Fire spiking neuron (aEIF)~\cite{Fourcaud2003How,brette2005adaptive}:

\begin{equation}
\left\{
\begin{aligned}
C \frac{d V}{d t}&=-g_{L}(V-E_{L})+g_{L} \exp(\frac{V-V_{t h}}{\Delta T})+I-w \\
\tau_{w} \frac{dw}{dt}&=a(V-E_L)-w
\end{aligned}
\right.
\end{equation}

where $g_L$ is the leak conductance, $E_L$ is the leak reversal potential, $V_r$ is the reset potential, $\Delta T$ is the slope factor, $I$ is the  background currents.  $\tau_w$ is the adaptation time constant. When the membrane potential is greater than the threshold $V_{th}$, $V = V_{r}$, and $w = w +b$.  $a$ is the subthreshold adaptation, and $b$ is the spike-triggered adaptation~\cite{Fourcaud2003How,brette2005adaptive}. 

(4) The Izhikevich spiking neuron~\cite{izhikevich2003simple}:
\begin{equation}
\left\{
\begin{aligned}
\frac{dV}{dt} &= 0.04V^2 + 5v + 140 - u + I \\
\frac{du}{dt} &= a(bv-u)\\
\end{aligned}
\right.
\end{equation}
When the membrane potential is greater than the threshold:
\begin{equation}
\left\{
\begin{aligned}
V &= c \\
u &= u + d\\
\end{aligned}
\right.
\end{equation}
$u$ represents the membrane recovery variable, and $a, b, c, d$ are the dimensionless parameters~\cite{izhikevich2003simple}. 

(5) The Hodgkin-Huxley spiking neuron (H-H)~\cite{hodgkin1952quantitative}:
\begin{equation}
I = C \frac{dV}{dt} + \overline{g}_K n^4 (V - V_K) + \overline{g}_{Na} m^3 h (V - V_{Na}) + \overline{g}_L (V - V_L)
\label{equa_hh}
\end{equation}
\begin{equation}
\left\{
\begin{aligned}
\frac{dn}{dt} &= \alpha_n(V)(1-n) - \beta_n(V)n \\
\frac{dm}{dt} &= \alpha_m(V)(1-m) - \beta_m(V)m \\
\frac{dh}{dt} &= \alpha_n(V)(1-h) - \beta_n(V)h \\
\end{aligned}
\right.
\end{equation}

$\alpha_i$  and $\beta_i$ are used to control the $i_{th}$ ion channel, $n$, $m$, $h$ are dimensionless probabilities between 0 and 1. $\overline{g}_i$ is the maximal value of the conductance~\cite{hodgkin1952quantitative}.

The H-H model shows elaborate modeling of biological neurons. In order to apply it more efficiently to AI tasks, BrainCog incorporates a simplified H-H model ($C=0.02\mu F/cm^2, V_r=0, V_{th}=60mV$), as illustrated in~\cite{wang2016spiking}.

\subsection{Learning Rules}
BrainCog provides various plasticity principles and rules to support biological plausible learning and inference, such as:

(1) Hebbian learning theory~\cite{amit1994correlations}:
\begin{equation}
\Delta w_{ij}^t=x_i^t x_j^t
\end{equation}
where $w_{ij}^t$ means the $i$th synapse weight of $j$th neuron at the time $t$. $x_i^t$ is the input of $i$th synapse at time $t$. $x_j^t$ is the output of $j$th neuron at time $t$~\cite{amit1994correlations}.

(2) Spike-Timing-Dependent Plasticity (STDP)~\cite{bi1998synaptic}:

\begin{equation}
	\begin{split}
		&\Delta w_{j}=\sum\limits_{f=1}^{N} \sum\limits_{n=1}^{N} W(t^{f}_{i}-t^{n}_{j})\\
		&W(\Delta t)= 
		\left\{
		\begin{split}
		A^{+}e^{\frac{-\Delta t}{\tau_{+}}}\quad if\; \Delta t>0&\\ 
		-A^{-}e^{\frac{\Delta t}{\tau_{-}}}\quad if\;\Delta t<0&\\ 
		\end{split}
		\right.
	\end{split}
	\label{eq4}
\end{equation}
where $\Delta w_{j}$ is the modification of the synapse $j$, and $W(\Delta t)$ is the STDP function. $t$ is the time of spike. $A^{+},A^{-}$ mean the modification degree of STDP. $\tau_{+}$ and $\tau_{-}$ denote the time constant~\cite{bi1998synaptic}.

(3) Bienenstock-Cooper-Munros theory (BCM)~\cite{bienenstock1982theory}:
\begin{equation}
\Delta w=y\left(y-\theta_M \right)x-\epsilon w
\end{equation}

Where $x$ and $y$ denote the firing rates of pre-synaptic and post-synaptic neurons, respectively, threshold $\theta_M$ is the average of historical activity of the post-synaptic neuron~\cite{bienenstock1982theory}.

(4) Short-term Synaptic Plasticity (STP)~\cite{maass2002synapses}:

Short-term plasticity is used to model the synaptic efficacy changes over time.

\begin{equation}
{a_k} = {u_k} \cdot {R_k}
\end{equation}
\begin{equation}
{u_{k+1}} = U + {u_{k}}(1 - U)\exp \left( { - {{\rm{\Delta }}t_{k}}/{\tau _{{\rm{fac}}}}} \right)
\end{equation}
\begin{equation}
{R_{k+1}} = 1 + \left( {{R_{k}} - {u_{k}}{R_{k}} - 1} \right)\exp \left( { - {{\rm{\Delta }}t_{k}}/{\tau _{{\rm{rec}}}}} \right)
\end{equation}

$a$ denotes the synaptic weight, $U$ denotes the fraction of synaptic resources. $\tau _{{\rm{fac}}}$ and $\tau _{{\rm{rec}}}$ denotes the time constant for recovery from facilitation and depression. The variable $R_n$models the fraction of synaptic efficacy available for the (k)th spike,and $u_n R_n$ models the fraction of synaptic efficacy~\cite{maass2002synapses}.

(5) Reward-modulated Spike-Timing-Dependent Plasticity (R-STDP)~\cite{Eugene2007}:

R-STDP uses synaptic eligibility trace $e$ to store temporary information of STDP. The eligibility trace accumulates the STDP $\Delta w_{STDP}$ and decays with a time constant $\tau_e$~\cite{Eugene2007}. 

\begin{equation}\label{rstdpe}
\Delta e=-\frac{e}{\tau _e}+\Delta w_{STDP}
\end{equation}

Then, synaptic weights are updated when a delayed reward $r$ is received, as Eq.~\ref{rstdpw} shown~\cite{Eugene2007}. 

\begin{equation}\label{rstdpw}
\Delta w=r*\Delta e 
\end{equation}

\subsection{Encoding Strategies}
BrainCog supports a number of different encoding strategies to help encode the inputs to the spiking neural networks.

(1) Rate Coding~\cite{adrian1926impulses}:

Rate coding is mainly based on spike counting to ensure that the number of spikes issued in the time window corresponds to the real value. Poisson distribution can describe the number of random events occurring per unit of time, which corresponds to the firing rate~\cite{adrian1926impulses}. Set $\alpha \sim \mathbb{U}(0,1)$, the input can be encoded as
\begin{equation}
	s(t) =  \begin{cases}
	   1,\quad if \quad x > \alpha\\
	   0, \quad else 
	\end{cases}
\end{equation}

(2) Phase Coding~\cite{kim2018deep}:

The idea of phase coding can be used to encode the analog quantity changing with time. The value of the analog quantity in a period can be represented by a spike time, and the change of the analog quantity in the whole time process can be represented by the spike train obtained by connecting all the periods.
Each spike has a corresponding phase weighting under phase encoding, and generally, the pixel intensity is encoded as a 0/1 input similar to binary encoding. $\gg$ denotes the shift operation to the right, $K$ is the phase period~\cite{kim2018deep}. Pixel $x$ is enlarged to $x^{\prime}=x*(2^{K}-1)$ and shifted $k=K-1-(t \mod K)$ to the right, where $\mod$ is the  remainder operation. If the lowest bit is one, $s$ will be one at time $t$. $\&$ means 
bit-wise AND operation.
\begin{equation}
	s(t) =  \begin{cases} 1, \quad if  \quad (x^{\prime} \gg k) \& 1 = 1\\
	0, \quad else
	\end{cases}
\end{equation}

(3) Temporal Coding~\cite{thorpe1996speed}:

The characteristic of the neuron spike is that the form of the spike is fixed, and there are only differences in quantity and time. The most common implementation is to express information regarding the timing of individual spike. The stronger the stimulus received, the earlier the spike generated~\cite{rueckauer2018conversion}. Let the total simulation time be T, and the input $x$ of the neuron can be encoded as the spike at time $t^s$:
\begin{equation}
	t^{s} = T - \operatorname{round}(Tx)
\end{equation}

(4) Population Coding~\cite{bohte2002error}:

Population coding helps to solve the problem of the ambiguity of information carried by a single neuron. The ambiguity of information carried can be understood as: the original information is input into a neuron, which makes the network hard to distinguish some overlapping or similar related information~\cite{quian2009extracting,grun2010analysis}.
The intuitive idea of population coding is to make different neurons to be with different sensitivity to different types of inputs, which is also reflected in biology. For example, rats' whiskers have different sensitivity to different directions~\cite{quian2009extracting}. The inputs will be transformed into a spike train with a period by population coding. A classical population coding method is the neural information coding method based on the Gaussian tuning curve referred to Eq.~\ref{gaus}. This method is more suitable when the amount of data is small, and the information is concentrated. A Gaussian neuron covers a range of analog quantities in the form of Gaussian function~\cite{bohte2002error}, \cite{li2021online}. Suppose that $m$ ($m > 2$) neurons are used to encode a variable x with a value range of $[I_{min}, I_{max}]$. $f(x)$ can be firing time or voltage.

\begin{equation}
        f(x) = x^{-\frac{(x-\mu)^{2}}{2\sigma^{2}}}
\label{gaus}
\end{equation}

The corresponding mean with adjustable parameters $\beta$ and variance of the $ith$ ($i=1,2,..., m$) neuron are as follows:

\begin{equation}
        \mu = I_{min} + \frac{2i-3}{2}\frac{I_{max}-I_{min}}{m-2}
\end{equation}

\begin{equation}
        \sigma = \frac{1}{\beta}\frac{I_{max}-I_{min}}{m-2}
\end{equation}

\subsection{Brain Areas Models}

Brain-inspired models of several functional brain areas are constructed for BrainCog from different levels of abstraction.

(1) Basal Ganglia (BG): 

Basal ganglia facilitates desired action selection and inhibit competing behavior (making winner-takes-all decisions)~\cite{chakravarthy2010basal,redgrave1999basal}. It cooperates with PFC and the thalamus to realize the decision-making process in the brain~\cite{parent1995functional}. BrainCog models the basal ganglia brain area, including excitatory and inhibitory connections among striatum, Globus pallidus internas (Gpi), Globus pallidus external (Gpe), and subthalamic nucleus (STN) of basal ganglia~\cite{lanciego2012functional}, as shown in the orange areas of Fig.~\ref{dmsnn}. The BG brain area component adopts the \emph{LIF} neuron model in BrainCog, as well as the \emph{STDP} learning rule and \emph{CustomLinear} to build internal connections of the BG. Then, the BG brain area component can be used to build brain-inspired decision-making SNNs (see section 3.2.1 for detail).

(2) Prefrontal Cortex (PFC):

PFC is of significant importance when human high-level cognitive behaviors happen. In BrainCog, many cognitive tasks based on SNN are inspired by the mechanisms of the PFC~\cite{bechara1998dissociation}, such as decision-making, working memory~\cite{rao1999isodirectional,d2000prefrontals,lara2015role}, knowledge representation~\cite{wood2003human}, theory of mind and music processing~\cite{frewen2015healing}. Different circuits are involved to complete these cognitive tasks. In BrainCog, the data-driven PFC column model contains 6 layers and 16 types of neurons. The distribution of neurons, membrane parameters and connections of different types of neurons are all derived from existing biological experimental data. The PFC brain area component mainly employs the \emph{LIF} neuron model to simulate the neural dynamics. The \emph{STDP} and \emph{R-STDP} learning rules are utilized to compute the weights between different neural circuits. 

(3) Primary Auditory Cortex (PAC):

The primary auditory cortex is responsible for analyzing sound features, memory, and the extraction of inter-sound relationships~\cite{Koelsch2012}. This area exhibits a topographical map, which means neurons respond to their preferred sounds. In BrainCog, neurons in this area are simulated by the \emph{LIF} model and organized as minicolumns to represent different frequencies of sounds. To store the ordered note sequences, the excitatory and inhibitory connections are updated by \emph{STDP} learning rule.

(4) Inferior Parietal Lobule (IPL):

The function of IPL is to realize motor-visual associative learning~\cite{macuga2011selective}. The IPL consists of two subareas: IPLM (motor perception neurons in IPL) and IPLV (visual perception neurons in IPL). The IPLM receives information generated by self-motion from ventral premotor cortex (vPMC), and the IPLV receives information detected by vision from superior temporal sulcus (STS). The motor-visual associative learning is established according to the STDP mechanism and the spiking time difference of neurons in IPLM and IPLV.

(5) Hippocampus (HPC):

The hippocampus is part of the limbic system and plays an essential role in the learning and memory processes of the human brain. Epilepsy patients with bilateral hippocampus removed (e.g. the patient H.M.) have symptoms of anterograde amnesia. They are unable to form new long-term declarative memories~\cite{milner1998}. This case study has proved that the hippocampus is in the key process of converting short-term memory to long-term memory and plays a vital role~\cite{smith1981role}. 

Furthermore, through electrophysiological means, it was found that the hippocampal region is also crucial for forming new concepts. Specific neurons in the hippocampus only respond selectively to specific concepts, completing the specific encoding between different concepts. Moreover, it was through the study of the hippocampus that neuroscientists discovered the STDP learning rule~\cite{dan2004spike}, which further demonstrated the high plasticity of the hippocampus.

(6) Insula:

The function of the Insula is to realize self-representation~\cite{craig2009you}, that is, when the agent detects that the movement in the field of vision is generated by itself, the Insula is activated. The Insula receives information from IPLV and STS. The IPLV outputs the visual feedback information predicted according to its motion, and the STS outputs the motion information detected by vision. If both are consistent, the Insula will be activated.

(7) Thalamus (ThA):

Studies have shown that the thalamus is composed of a series of nuclei connected to different brain parts and heavily contributes to many brain processes. In BrainCog, this area is discussed from both anatomic and cognitive perspectives. Understanding the anatomical structure of the thalamus can help researchers to comprehend the mechanisms of the thalamus. Based on the essential and detailed anatomic thalamocortical data~\cite{Izhikevichlargescale}, BrainCog reconstructs the thalamic structure by involving five types of neurons (including excitatory and inhibitory neurons) to simulate the neuronal dynamics and building the complex synaptic architecture according to the anatomic results. Inspired by the structure and function of the thalamus, the brain-inspired decision-making model implemented by BrainCog takes into account the transfer function of the thalamus and cooperates with the PFC and basal ganglia to realize multi-brain areas coordinated decision-making model.

(8) Ventral Visual Pathway:

Cognitive neuroscience research has shown that the brain can receive external input and quickly recognize objects due to the hierarchical information processing of the ventral visual pathway. The ventral pathway is mainly composed of V1, V2, V4, IT, and other brain areas, which mainly process information such as object shape and color~\cite{ishai1999distributed,kobatake1994neuronal}. These visual areas form connections through forward, feedback, and self-layer projections. The interaction of different visual areas enables humans to recognize visual objects. The primary visual cortex V1 is selective for simple edge features. With the transmission of information, high-level brain areas combine with lower-level receptive fields to form more complex large receptive fields to recognize more complex objects~\cite{hubel1962receptive}. Inspired by the structure and function of the ventral visual pathway, BrainCog builds a deep forward SNN with layer-wise information abstraction and a feedforward and feedback interaction deep SNN. The performance is verified on several visual classification tasks. 

(9) Motor Cortex:

The control of biological motor function involves the cooperation of many brain areas. The extra circuits consisting of the PMC, cerebellum, and BA6 motor cortex area are primarily associated with motor control elicited by external stimuli such as visual, auditory, and tactual inputs. Internal motor circuits, including the basal ganglia and SMA, predominate in self-guided, learned movements~\cite{geldberg1985supplementary,mushiake1991neuronal,gerloff1997stimulation}. Population activity of motor cortical neurons encodes movement direction. Each neuron has its preferred direction. The more consistent the target movement direction is with its preferred direction, the stronger the neuron activities are~\cite{georgopoulos1995motor,kakei1999muscle}. The cerebellum receives input from motor-related cortical areas such as PMC, SMA, and the prefrontal cortex, which are important for the completion of fine movements, maintaining balance and coordination of movements~\cite{strick2009cerebellum}. Inspired by the organization of the brain's motor cortex, we use spiking neurons to construct a motion control model, and apply it to the iCub robot, which enables the robot to play the piano according to music pieces.

\section{Brain-inspired AI}
Computational units (different neuron models, learning rules, encoding strategies, brain area models, etc.) at multiple scales, provided by BrainCog serve as a foundation to develop functional networks. To enable BrainCog for Brain-inspired AI, cognitive function centric networks need to be built and provided as reusable functional building blocks to create more complex brain-inspired AI. This section introduces various functional building blocks developed based on BrainCog.

\subsection{Perception and Learning}
In this subsection, we will introduce the supervised and unsupervised perception and learning spiking neural networks based on the fundamental components of BrainCog. Inspired by the global feedback connections, the neural dynamics of spiking neurons, and the biologically plausible STDP learning rule, we improve the performance of the spiking neural networks. We show great adaption in the small training sample scenario. In addition, our model has shown excellent robustness in noisy scenarios by taking inspiration from the multi-component spiking neuron and the quantum mechanics. The burst spiking mechanism is used to help our converted SNNs with higher performance and lower latency. Based on this engine, we also present a human-like concept learning framework to generate representations with five types of perceptual strength information.

\textbf{1. Learning Model}

\emph{(1) SNN with global feedback connections}

The spiking neural network transmits information in discrete spike sequences, which is consistent with the information processing in the human brain~\cite{Maass1997Networks}. The training of spiking neural networks has been widely concerned by researchers. Most researchers take inspiration from the mechanism of synaptic learning and updating between neurons in the human brain, and propose biologically plausible learning rules, such as Hebbian theory~\cite{amit1994correlations}, STDP~\cite{bi1998synaptic}, and STP~\cite{zucker2002short}, which have been adopted into the training of spiking neural networks~\cite{tavanaei2016bio,tavanaei2017multi,diehl2015unsupervised,falez2019multi,zhang2018plasticity,zhang2018brain}. However, most SNNs are based on feedforward structures, while the importance of the brain-inspired structures has been ignored. The anatomical and physiological evidences show that in addition to feedforward connections, numerous feedback connections exist in the brain, especially among sensory areas~\cite{felleman1991distributed,sporns2004small}. The feedback connections will carry out the predictions from the top layer to cooperate with the local plasticity rules to formulate the learning and inference in the brain. 

Here, we introduce the global feedback connections and the local differential learning rule~\cite{zhao2020glsnn} in the training of SNNs. 
\begin{figure}[!htbp]
	\centering
	\includegraphics[scale=0.6]{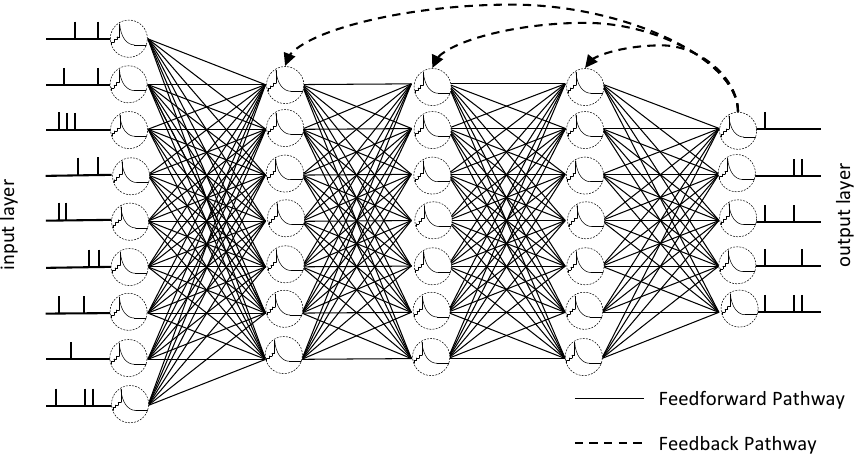}
	\caption{The feedforward and feedback pathway in the SNN model. The global feedback pathway propagates the target of the hidden layer, modified from~\cite{zhao2020glsnn}.}
	\label{snn}
\end{figure}

We use the LIF spiking neuron model in the BrainCog to simulate the dynamical process of the membrane potential $V(t)$ as shown in Eq.~\ref{equa_lif1}. We use the mean firing rates $S_{l}$ of each layer to denote the representation of the $l_{th}$ layer in the forward pathway, and the corresponding target is denoted as $\hat{S}_{l}$. Here we use the mean squared loss (MSE) as the final loss function.

$\hat{S}_{L-1}$ denotes the target of the penultimate layer, and is calculated as shown in Eq.~\ref{equal_per}, $W_{L-1}$ denotes the forward weight between the $(L-1)_{th}$ layer and the $L_{th}$. $\eta_t$ represents the learning rate of the target~\cite{zhao2020glsnn}.
\begin{equation}
	\hat{S}_{L-1} = S_{L-1} - \eta_t\Delta S = S_{L-1} - \eta_t W_{L-1}^T(S_{out} - S^T)
	\label{equal_per}
\end{equation}

The target of the other hidden layer can be obtained through the feedback connections:

\begin{equation}
	\hat{S}_{l} = S_{l} - G_{l}(S_{out} - S^T)
	\label{equal_all}
\end{equation}

\begin{figure}[!htbp]
	\centering
	\includegraphics[scale=0.35]{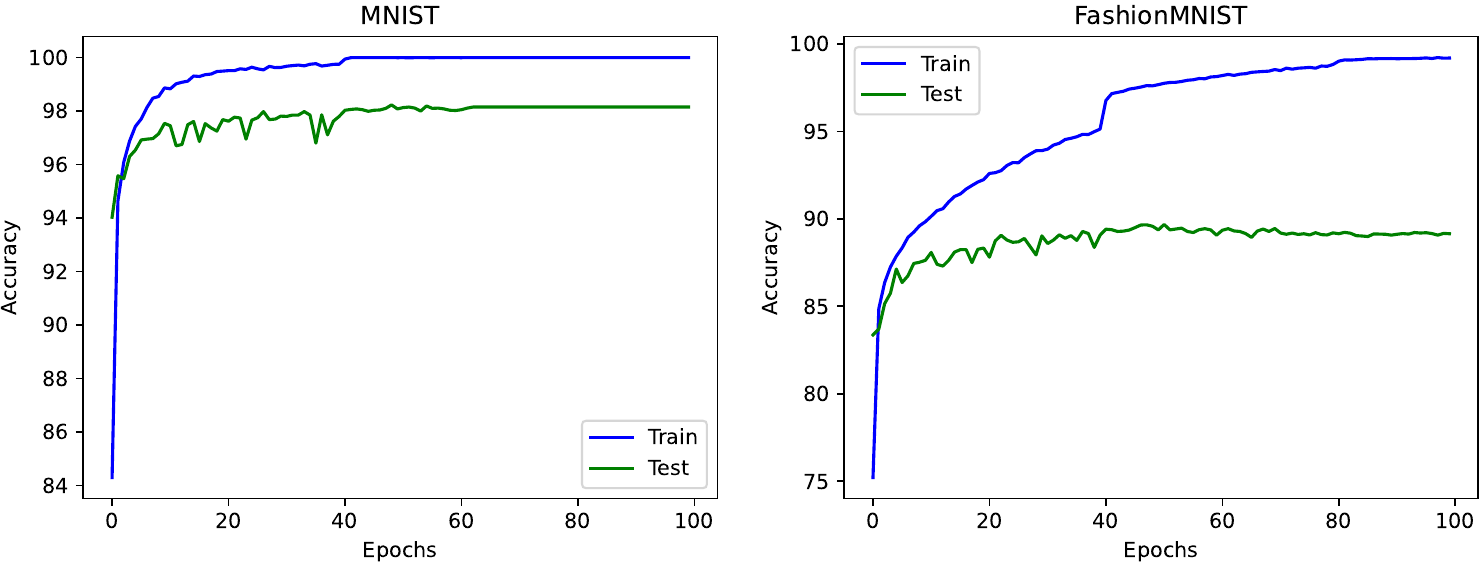}
	\caption{The test accuracy on MNIST and Fashion-MNIST datasets of the SNNs with global feedback connections.}
	\label{zdc_result}
\end{figure}

By combining the feedforward representation and feedback target, we compute the local MSE loss. We can compute the local update of the parameters with the surrogate gradient. We have conducted experiments on the MNIST and Fashion-MNIST datasets, and achieved 98.23\% and 89.68\% test accuracy with three hidden layers. Each hidden layer is set with 800 neurons. The details are shown in Fig.~\ref{zdc_result}.

\emph{(2) Biological-BP SNN}

The backpropagation algorithm is an efficient optimization method that is widely used in deep neural networks and has promoted the great success of deep learning. However, due to the non-smoothness of neurons in SNNs, the backpropagation optimization is difficult to be applied to train SNNs directly. To solve the above problem, Bengio et al.~\cite{bengioEstimatingPropagatingGradients2013} proposed four gradient approximation methods, including Straight-Through Estimator to enable the application of the backpropagation algorithm to neural networks containing nonsmooth neurons. Wu et al.~\cite{wuSpatioTemporalBackpropagationTraining2018} proposed the Spatio-Temporal Backpropagation (STBP) algorithm to train SNNs by using a differentiable function to approximate spiking neurons through a surrogate gradient method. Based on the spatio-temporal dynamic of SNNs, they achieved the backpropagation of SNNs in both time and space dimensions. Zheng et al.~\cite{zhengGoingDeeperDirectlyTrained2021} proposed threshold-dependent batch normalization (tdBN) to improve the performance of deep SNNs. Fang et al.~\cite{fangIncorporatingLearnableMembrane2021} proposed a parametric LIF (PLIF) model to further improve the performance of SNNs by optimizing the time constants of neurons during the training process. 

\begin{figure}[!htbp]
	\centering
	\includegraphics[scale=0.48]{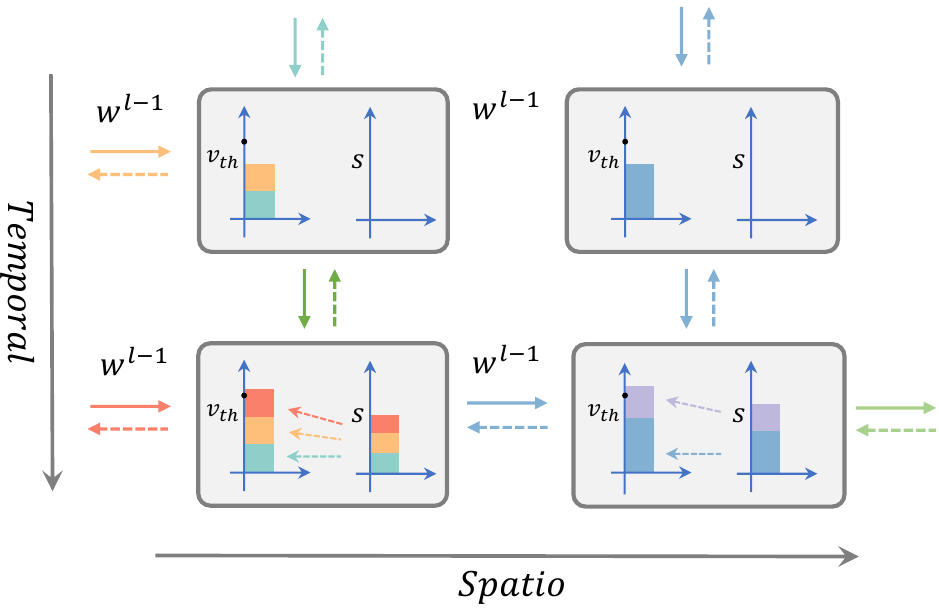}
	\caption{The forward and backward process of biological BP-based SNNs for BrainCog.}
	\label{bpsta}
\end{figure}

Most of these BP-based SNNs often simply regard SNN as a substitute for RNN, but ignore the dynamic of spiking neurons. To solve the problem of non-differentiable neuronal models, Bohte et al.~\cite{bohte2011error} approximated the backpropagation process of neuronal models using the surrogate gradient method. However, this method results in gradient leakage and does not allow for proper credit assignment in both temporal and spatial dimensions. Also, due to the reset operation after the spikes are emitted, the errors in the backward process can not propagate across spikes. To solve the above problem, we propose Backpropagation with biologically plausible spatio-temporal adjustment~\cite{shen2022backpropagation}, as shown in Fig.~\ref{bpsta}, which can correctly assign credit according to the contribution of the neuron to the membrane potential at each moment.

Based on LIF spiking neuron, the direct input encoding strategy, the MSE loss function and the surrogate gradient function supplied in BrainCog, we propose a Biologically Plausible Spatio-Temporal Adjustment (BPSTA) to help BP algorithm with more reasonable error adjustment in the spatial temporal dimension~\cite{shen2022backpropagation}.  The algorithm realizes the reasonable adjustment of the gradient in the spatial dimension, avoids the unnecessary influence of the neurons that do not generate spikes on the weight update, and extracts more important features. By applying the temporal residual pathway, our algorithm helps the error to be transmitted across multiple spikes, and enhances the temporal dependency of the BP-based SNNs. Compared with SNNs and ANNs with the same structure that only used the BP algorithm, our model greatly improves the performance of SNNs on the DVS-CIFAR10 and DVS-Gesture datasets, while also greatly reducing the energy consumption and decay of SNNs, as shown in Tab.~\ref{compare2}.
\begin{table}[!htbp]
 \centering
 \caption{The energy efficiency study. The former represents our method, the latter represents the baseline, adopted from~\cite{shen2022backpropagation}.}
 \begin{tabular}{cccc}
  \toprule[2pt]
   Dataset   &  Accuracy    &  Firing-rate  &EE = $\frac{E_{ANN}}{E_{SNN}}$ \\
  \midrule[2pt]
   MNIST     &  99.58\%/99.42\% & 0.082/0.183 & 35.1x/15.7x  
  \\
  
    N-MNIST   &  99.61\%/99.32\%  & 0.097/0.176   & 29.6x/16.3x      \\
  
   CIFAR10   &  92.33\%/89.49\%  & 0.108/0.214 & 26.6x/13.4x    \\ 
  
   DVS-Gesture  &  98.26\%/93.92\%  & 0.083/0.165   & 34.6x/17.4x    \\ 
       
        DVS-CIFAR10  &  77.76\%/71.40\%  & 0.097/0.177  & 29.5x/16.2x    \\
  \bottomrule[2pt]
 \end{tabular}
 \label{compare2}
\end{table} 

\emph{(3) Unsupervised STDP-based SNN}
 
\begin{figure}
\centering
\includegraphics[width=0.48\textwidth]{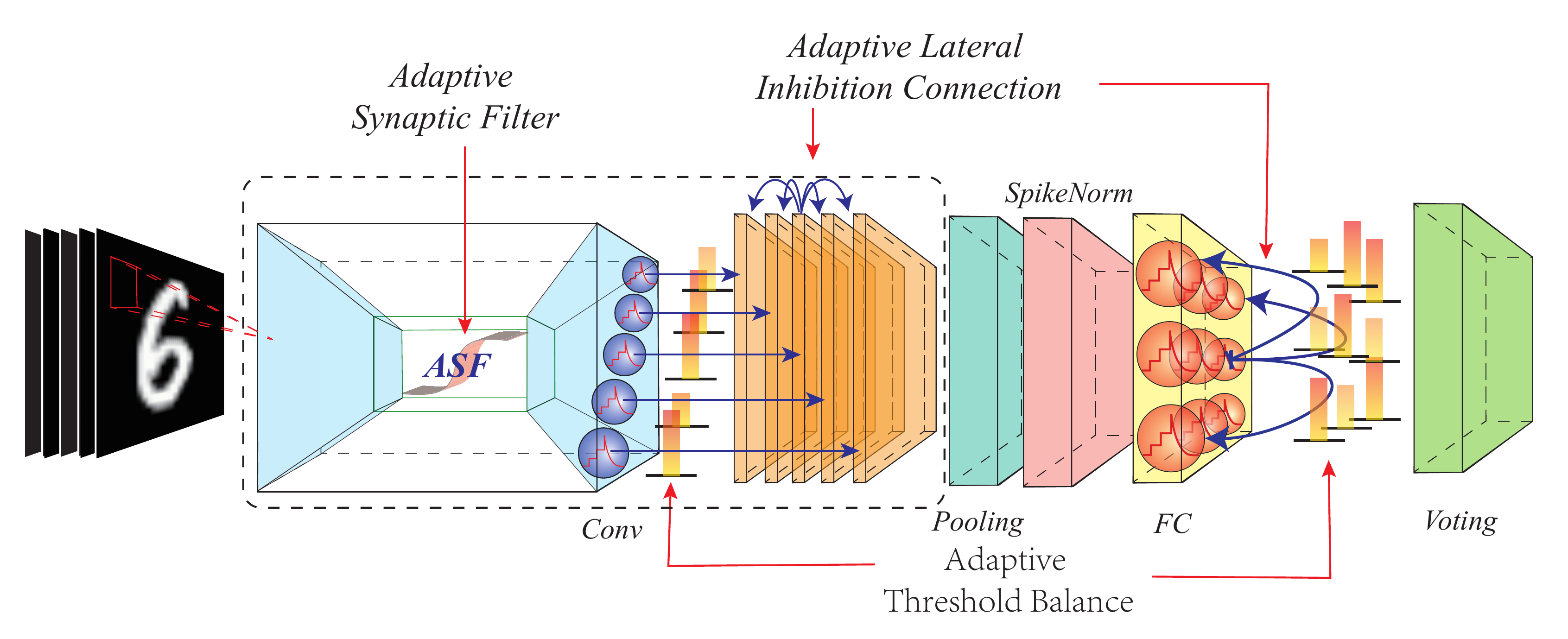}
\caption{The framework of the unsupervised STDP-based spiking neural network model, which introduces the adaptive synaptic filter (ASF), the adaptive threshold balance (ATB), and the adaptive lateral inhibitory connection (ALIC) mechanisms to improve the information transmission and feature extraction of STDP-based SNNs. This figure is from \cite{dong2022unsupervised}.}
\label{frame}
\end{figure}

Unsupervised learning is an important cognitive function of the brain. The brain can complete the task of object recognition by summarizing the characteristics and features of objects. Unsupervised learning does not require explicit labels, but extracts the features of samples adaptively in learning process. Modeling of this ability of the brain is critical. There are multiple learning rules in the brain to accomplish various learning tasks. STDP is a widespread rule of synaptic weight modification in the brain. It updates the synaptic weights according to the temporal relationship of the pre- and post-synaptic spikes. Compared with the backpropagation algorithm widely used in DNN with gradient calculation, STDP is more biological plausible. However, unlike the backpropagation algorithm that relies on a large number of sample labels, STDP is a local optimization algorithm. Due to the lack of global information, the ability of self-organization and coordination between neurons is insufficient. In the SNN model, it will lead to disorder and uncertainty of spikes discharge, and it is difficult to achieve a stable release balance state of neurons. To this end, we design an unsupervised STDP-based spiking neural network model based on BrainCog, and bring unsupervised learning to BrainCog as a functional module.
As shown in Fig.~\ref{frame}.

\begin{figure}
\centering
\includegraphics[width=0.49\textwidth]{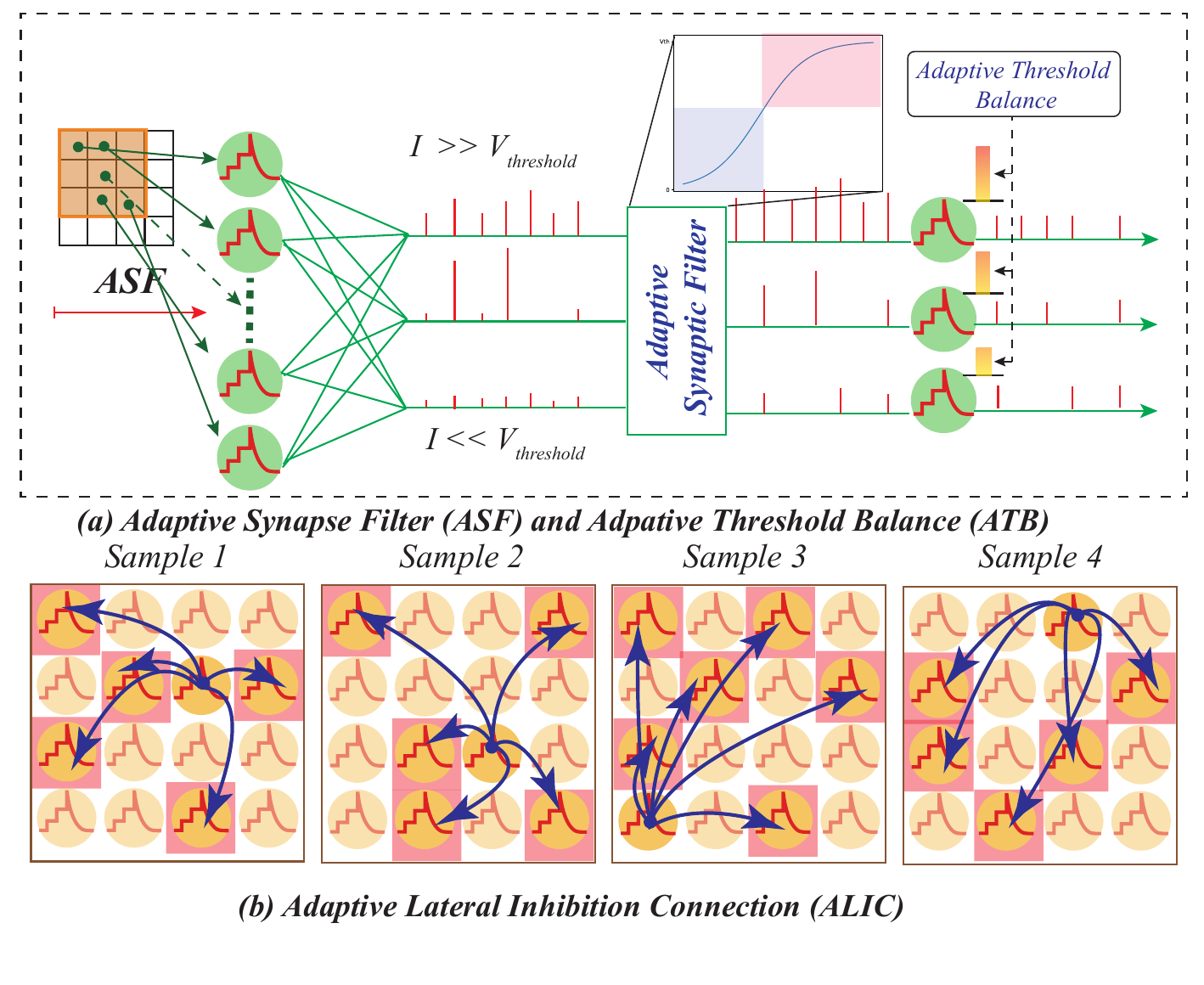}
\caption{(a) The adaptive synaptic filter and the adaptive threshold balance jointly regulate the neuron spikes balance of neuron. (b) The adaptive lateral inhibitory connection has different connections for different samples, preventing neurons from learning the same features. This figure is from~\cite{dong2022unsupervised}.}
\label{stp}
\end{figure}

To solve the above problems, we introduced various adaptive mechanisms to improve the self-organization ability of the overall network. STP is another synaptic learning mechanism that exists in the brain. Inspired by STP, we designed an adaptive synaptic filter (ASF) that integrates input currents through nonlinear units, and an adaptive threshold balance (ATB) that dynamically changes the threshold of each neuron to avoid excessively high or low firing rates. The combination of the two controls the firing balance of neurons. As the Fig.~\ref{stp} shows. We also address the problem of coordinating neurons within a single layer with an adaptive lateral inhibitory connection (ALIC). The mechanism have different connection structures for different input samples. Finally, in order to solve the problem of low efficiency of STDP training, we designed a sample temporal batch STDP. It combines the information between temporal and samples to uniformly update the synaptic weights, as shown by the following formula.\begin{equation}
	\begin{split}
		&\frac{dw_{j}^{(t)}}{dt}^{+}=\sum \limits_{m=0}^{N_{batch}}\sum \limits_{n=0}^{T_{batch}}\sum\limits_{f=1}^{N} W(t^{f,m}_{i}-t^{n,m}_{j})\\
	\end{split}
	\label{eq5}
\end{equation} where $W(x)$ is the function of STDP, $N_{batch}$ is the batchsize of the input, $T_{batch}$ is the batch of time step, N is the number of neurons. We verified our model on MNIST and Fashion-MNIST, achieving 97.9\% and 87.0\% accuracy, respectively. To the best of our knowledge, these are the state-of-the-art results for unsupervised SNNs based on STDP.

\textbf{2. Adaptability and Optimization}

\emph{(1) Quantum superposition inspired SNN}

In the microscopic size, quantum mechanics dominates the rules of operation of objects, which reveals the probabilistic and uncertainties of the world. New technologies based on quantum theory like quantum computation and quantum communication provide an alternative to information processing. Researches show that biological neurons spike at random and the brain can process information with huge parallel potential like quantum computing. 

\begin{figure}[!htb]
	\centering
	\includegraphics[width=1.0\linewidth]{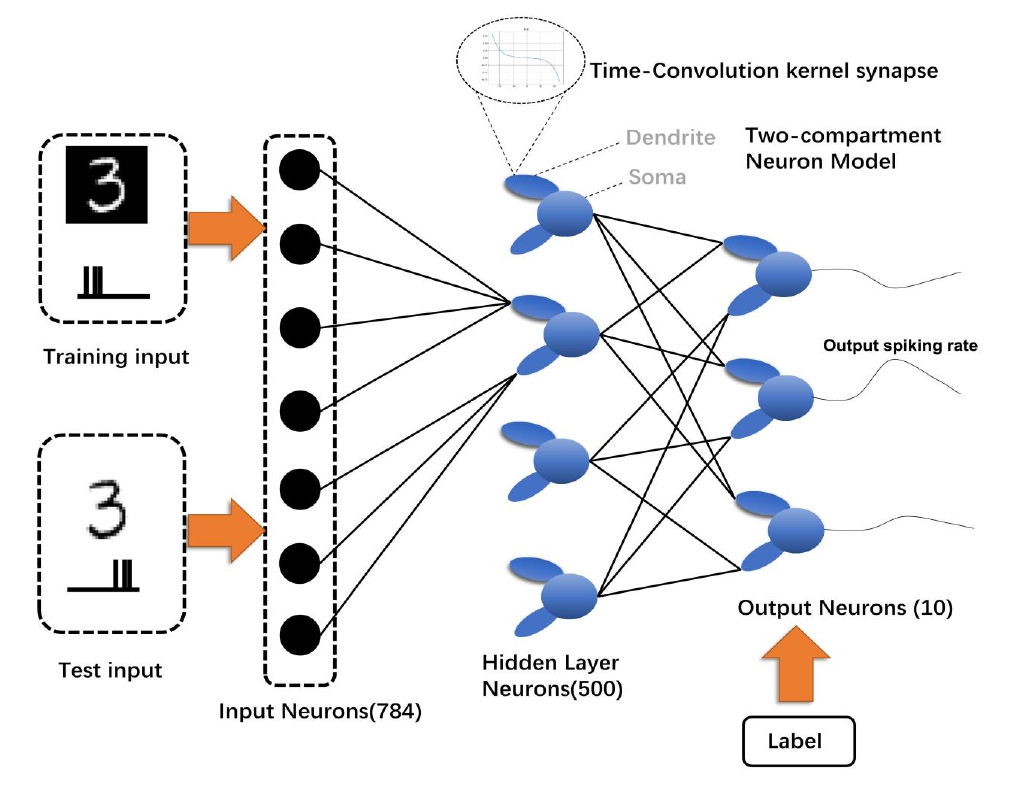}  
	\caption{Quantum superposition inspired spiking neural network, adopted from \cite{SUN2021102880}.}
	\label{Fig:QSSNN}
\end{figure}

Inspired from this, we propose the Quantum Superposition Inspired Spiking Neural Network (QS-SNN)~\cite{SUN2021102880}, complementing quantum image (CQIE) method to represent image in the form of quantum superposition state and then transform this state to spike trains with different phase. Spiking neural network with time differential convolution kernel (TCK) is used to do further classification shown in Fig.~\ref{Fig:QSSNN}.

%

\begin{figure}[!htb]
	\centering
	\includegraphics[width=1.0\linewidth]{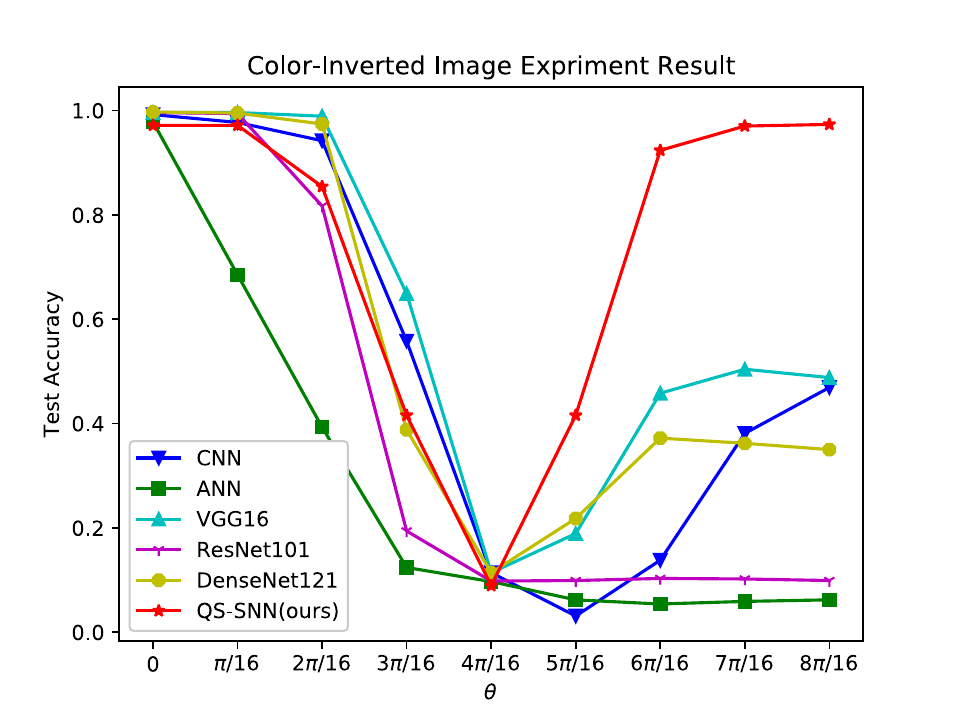}  
	\caption{Performance of QS-SNN on background MNIST inverse image, adopted from \cite{SUN2021102880}.}
	\label{Fig:mnist-res}
\end{figure}

The effort tries to incorporate the quantum superposition mechanism to SNNs as a new form of encoding strategy for BrainCog, and the model finally shows its capability on robustness for learning. The proposed QS-SNN model is tested on color inverted MNIST datasets. The background-inverted picture is encoded in the quantum superposition form as shown in Eq.~\ref{QS-SNN} and~\ref{QS-SNN_limitation}.
\begin{equation}
	\mathinner{|I(\theta)\rangle}=\frac{1}{2^n}\sum\limits_{i=0}^{2^{2n}-1}(cos(\theta_{i})\mathinner{|x_{i}\rangle}+sin(\theta_{i})\mathinner{|\bar{x}_{i}\rangle})\otimes\mathinner{|i\rangle}, \\
	\label{QS-SNN}
\end{equation}

\begin{equation}
	\theta_{i} \in [0, \frac{\pi}{2}], i= 1, 2, 3, \dots, 2^{2n}-1.
	\label{QS-SNN_limitation}
\end{equation}

Spike sequences of different frequencies and phases are generated from the picture information of the quantum superposition state. Furthermore we use two-compartment spiking neural networks to process these spike trains.

We compare the QS-SNN model with other convolutional models. The result in Fig.~\ref{Fig:mnist-res} shows that our QS-SNN model overtakes other convolutional neural networks in recognizing background-inverted image tasks.

\emph{(2) Unsupervised SNN with adaptive learning rule and structure}

Brain can accomplish specific tasks by adaptively learning to organize the features of a small number of samples. Few-shot learning is an important ability of the brain. In the above section, an unsupervised STDP-based spiking neural network is introduced~\cite{dong2022unsupervised}. To better illustrate the power of our model on small sample training, we tested the model with small samples and find that this model has stronger small sample processing ability than ANN with similar structures, as shown in Tab.~\ref{small}~\cite{dong2022unsupervised}.

\begin{table}[h]
	\caption{The performance of unsupervised SNN compared with ANN on MNIST dataset with different number of training samples~\cite{dong2022unsupervised}.}
	\centering
	\resizebox{0.8\linewidth}{!}{
		\begin{tabular}{lrrrr}
			\toprule 
			samples&200  &100&50&10 \\
			\midrule	
			ANN&79.77\%&71.40\%&68.72\%&47.12\%\\
			Ours&81.45\%&75.44\%&72.88\%&51.45\%\\
			\midrule
			&1.68\%&4.04\%&4.16\%&4.33\%\\
			\bottomrule
		\end{tabular}
	}
	\label{small}
	
\end{table}

\emph{(3) Efficient and Accurate Conversion of SNNs}

	SNNs have attracted attention due to their biological plausibility, fast inference, and low energy consumption. However the training methods based on the plasticity~\cite{zeng2017improving} and surrogate gradient algorithms~\cite{wu2018spatio} need much memory and perform worse than ANN on large networks and complex datasets. For users of BrainCog, we are certain there will be clear need to use SNNs while keep the benifit from ANNs. As an efficient method, the conversion method combines the characteristics of backpropagation and low energy consumption and can achieve the same excellent performance as ANN with lower power consumption~\cite{li2021free, han2020deep, han2020rmp}. However, the converted SNNs typically suffer from severe performance degradation and time delays. 
	
	\begin{figure}[htb]
    	\centering
    	\includegraphics[scale=0.3]{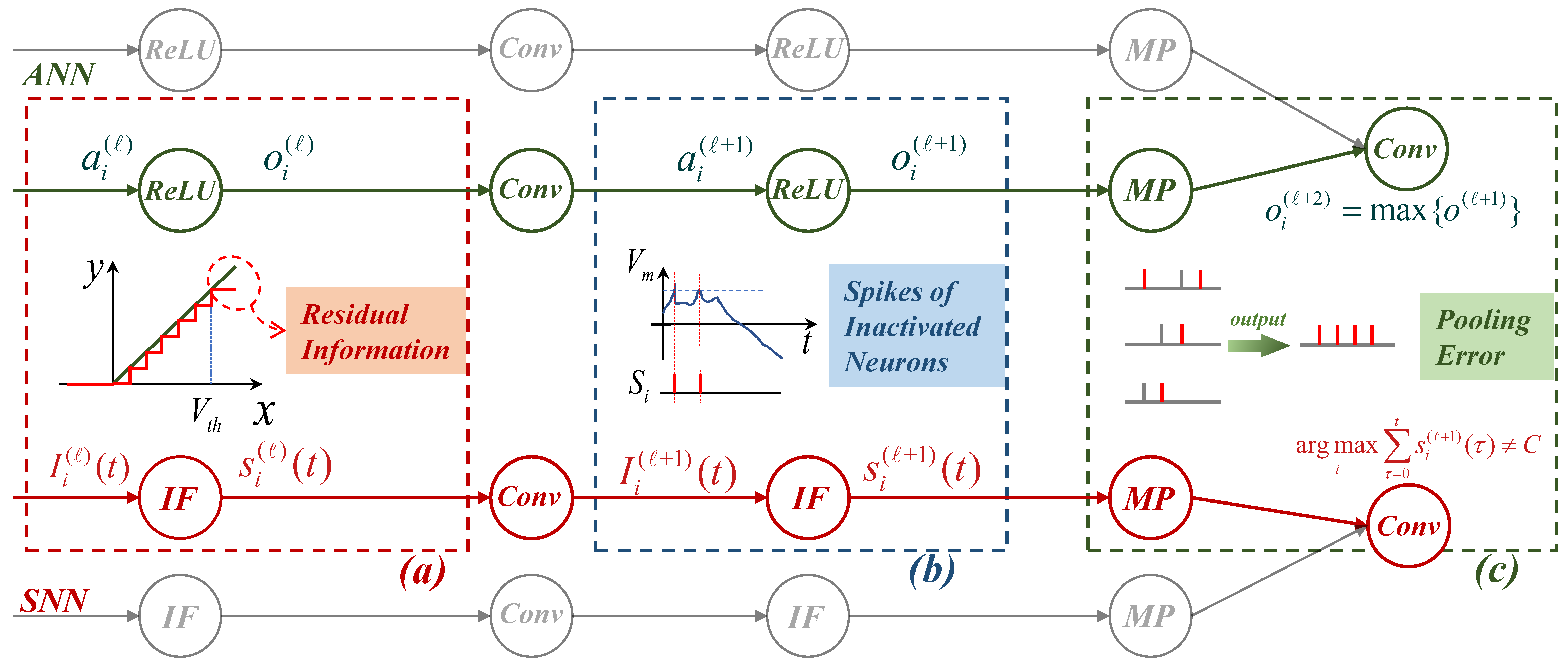}
    	\caption{The conversion errors from IF neuron, time dimension, and MaxPooling, adopted from~\cite{li2022efficient}.}
    	\label{error}
    \end{figure}

    We divide the performance loss to IF neurons, time dimension, and MaxPooling layer~\cite{li2022efficient}, as shown in Fig.~\ref{error}. In SNN, the neuron can only send one spike at most in each time step, so the maximum firing rate of the neuron is 1. After normalizing the weight of trained ANN, some activation values greater than 1 cannot be effectively represented by IF neurons. Therefore, the residual membrane potential in neurons will affect the performance of the conversion to some extent. In addition, since IF receives pre-synaptic neuron spikes for information transmission, the synaptic current received by the neuron at each time is unstable, but its sum can be approx to the converted value by increasing the simulation time. However, the total number of spikes is the time-varying extremum of the sum of the synaptic currents received. When the activation value corresponding to the IF neuron is negative and the total synaptic current received by the neuron in a short time exceeds the threshold, the neuron that should be resting will issue the spike, and the influence of the spike on the later layers cannot be eliminated by increasing the simulation time. Finally, in the conversion of the MaxPooling layer, the previous work has enabled spikes from the neuron with the maximum firing rate to pass through. However, due to the instability of synaptic current, neurons with the maximum firing rate are often not fixed, which makes the output of the converted MaxPooling layer usually larger.

    \begin{figure}[!h]
    	\centering
    	\includegraphics[scale=0.5]{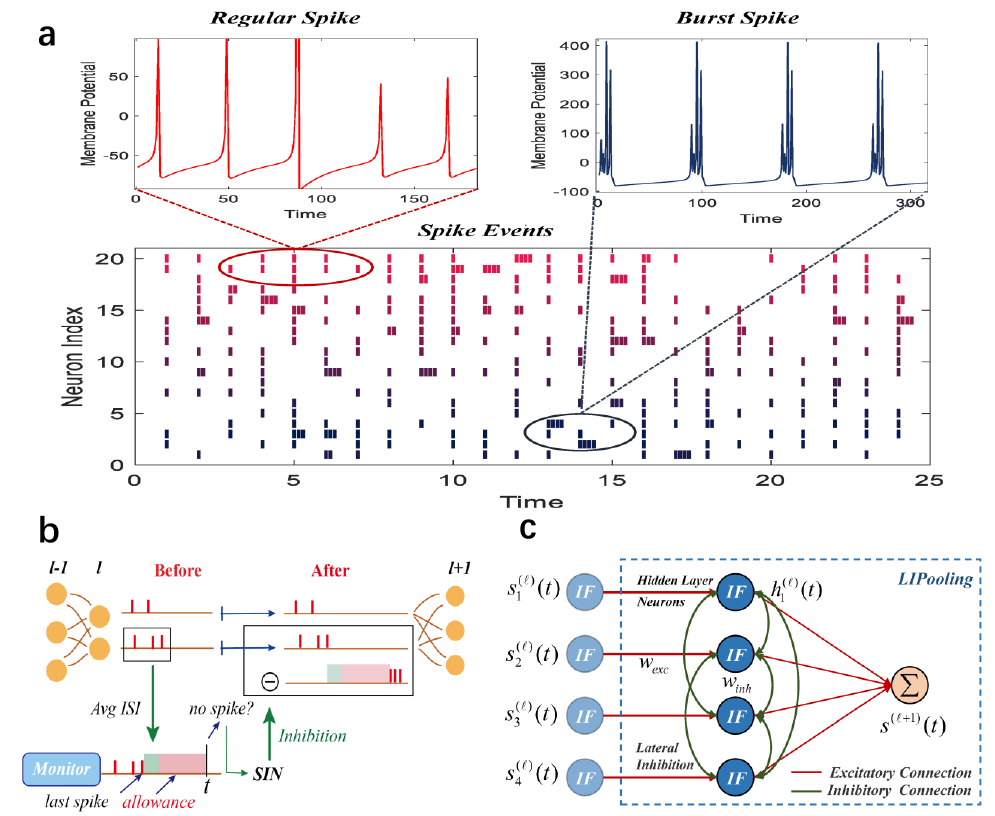}
    	\caption{The conversion methods. (a) Burst spikes increase the upper limit of firing rates; (b) spike calibration corrects the effect of faulty spikes on conversion, and (c) LIPooling uses lateral inhibition mechanisms to achieve accurate conversion of the MaxPooling layers. Refined based on~\cite{li2022efficient,li2022spike}.}
    	\label{conver}
    \end{figure}

    To solve the problem of the residual membrane potential of neurons, we introduce the burst mechanism, as shown in Fig.~\ref{conver} (a), which enables neurons to send more than one spike between two time steps, depending on the current membrane potential. Once some neurons have residual information remaining, they can send spikes between two time steps. In this way, the firing rate of SNN can be increased, and the membrane potential remaining in the neuron can be transmitted to the neuron of the next layer. 
    
    For classification problems, SNN only needs to ensure that the index of the maximum output is correct, but for more demanding conversion tasks, such as object detection, the solution of SIN problem is worth exploring. Note that the spikes emitted by hidden layer neurons are unstable, but the mean of its inter-spike interval distribution is related to its corresponding activation value~\cite{li2022spike}. We monitor each neuron's spiking time in the forward propagation process and update its average inter-spike interval, as shown in Fig.~\ref{conver} (b). Under a certain time allowance, the neuron that does not emit spikes will be determined to be Inactivated Neuron. Then, the twin weights emit spikes to suppress the influence of historical errors and calibrate the influence of wrong spikes to a certain extent to ensure accurate conversion.

    Inspired by the lateral inhibition mechanism~\cite{blakemore1970lateral}, we propose LIPooling for converting the maximum pooling layer, as shown in Fig.~\ref{conver} (c). From the operation perspective, the inhibition of other neurons by the winner in LIPooling is -1, so the neuron with the largest firing rate at the current time step may not spike due to the inhibition of other neurons in history. From the output perspective, LIPooling sums up the output of all neurons during simulation. So the key is that LIPooling uses competition between neurons to get an accurate sum (equal to the actual maximum), instead of picking the winner.

\textbf{3. Multi-sensory Integration}

One can build SNN models through BrainCog to process different types of sensory inputs, while the human brain learns and makes decisions based on multi-sensory inputs. When information from various sensory inputs is combined, it can lead to increased perception, quicker response times, and better recognition. Hence, enabling BrainCog to process and integrate multi-sensory inputs are of vital importance.

In this section, we focus on concept learning with multi-sensory inputs. We present a multi-sensory concept learning framework based on BrainCog to generate integrated vectors with the multi-sensory representation of the concept.

\begin{figure}[h!]
  \centering
  \includegraphics[width=8cm]{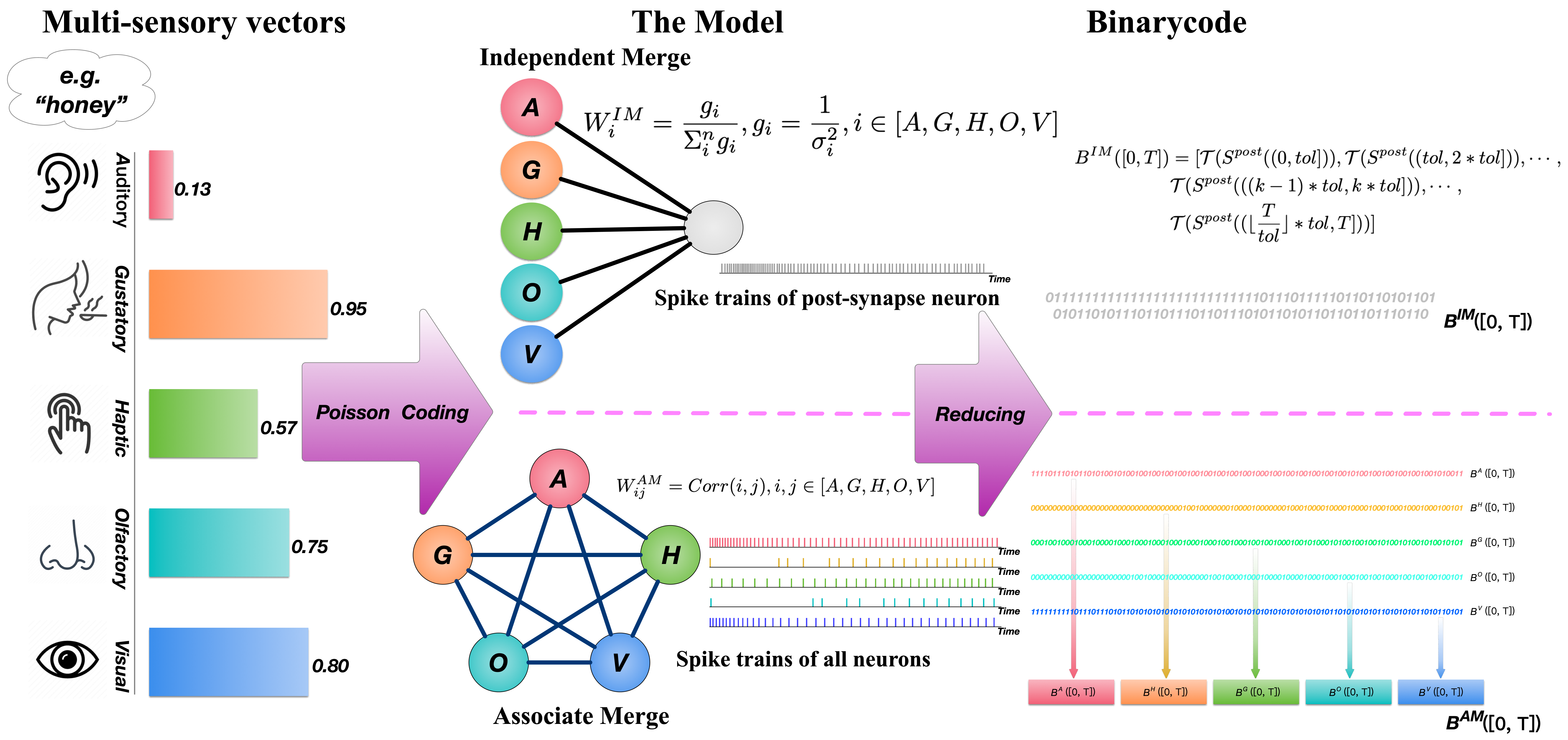}\\
  \caption{The Framework of Concept Learning Based on SNNs with multi-sensory Inputs.}
  \label{MultisensoryIntegrationFramework}
\end{figure}

Embodied theories, which emphasize that meaning is rooted in our sensory and experiential interactions with the environment, supports multi-sensory representations.
Based on SNNs, we present a human-like framework to learn concepts which can generate integrated representations with five types of perceptual strength information~\cite{wywFramework}.
The framework is developed with two distinct paradigms: Associate Merge (AM) and Independent Merge (IM), as Fig.~\ref{MultisensoryIntegrationFramework} shows.

IM is based on the widely accepted cognitive psychology premise that each type of sense for the concept is independent before integration~\cite{wywFramework}.
As the input to the model, we will employ five common perceptual strength: visual, auditory, haptic, olfactory and gustatory.
During the data preparation step, we min-max normalize all kinds of perceptual strength of the concept in the multi-sensory dataset so that each value of the vector is in $[0, 1]$.
We regard them as stimuli to the presynaptic neurons.
It's a 2-layer SNN model, with 5 neurons in the first layer matching the concept's 5 kinds of perceptual strength, and 1 neuron in the second layer representing the neural state following multi-sensory integration.
In this paradigm, we use perceptual strength based presynaptic Poisson neurons and LIF or Izhikevich as the postsynaptic neural model.

The weights between the neurons are $W^i= \frac{g_i}{\Sigma_i^n g_i} $ where  $g_i = \frac{1}{\sigma_i^2}$,$\sigma_i^2$ is the variance of each kind of perceptual strength.
We convert the postsynaptic neuron's spiking train $S^{post}([0, T])$ in $[0, T]$ into integrated representations $B^{IM}([0,T])$ for the concept in this form:
\begin{equation}
\begin{aligned}
B^{IM}([0,T]) &= [\mathcal{T}(S^{post}((0, tol])), \mathcal{T}(S^{post}((tol, 2*tol])), \cdots , \\
& \mathcal{T}(S^{post}(((k-1)*tol, k*tol])) , \cdots,  \\
& \mathcal{T}(S^{post}((\lfloor \frac{T}{tol} \rfloor * tol, T]))]
\end{aligned}
\end{equation}
Here if the interval has any spikes, the bit is 1. Otherwise it is 0, according to the $\mathcal{T} (interval)$ function.

The AM paradigm presupposes that each kind of modality is associated before integration~\cite{wywFramework}. 
It includes 5 neurons, matching the concept's 5 distinct modal information sources. 
They are linked with each other and are not self-connected. 
The input spike trains to the network are generated using a Poisson event-generation algorithm based on perceptual strength.
For each concept, we turn the spike trains of these neurons into the ultimate integrated representations.

The weight value is defined by the correlation between each two modalities, i.e. $W = Corr(i, j)$, where $i, j \in [A, G, H, O, V]$.
We convert the spike trains $S^i([0, T])$ of all neurons into binarycode $B^i([0,T])$ and conjoin them as the ultimate vector $B([0,T])$ as follows:
\begin{equation}
\begin{aligned}
B^i([0,T]) &= [\mathcal{T}(S^i((0, tol])), \mathcal{T}(S^i((tol, 2*tol])), \cdots , \\
& \mathcal{T}(S^i(((k-1)*tol, k*tol])) , \cdots, \\
& \mathcal{T}(S^i((\lfloor \frac{T}{tol} \rfloor * tol, T]))]
\end{aligned}
\end{equation}

\begin{equation}
\begin{aligned}
B^{AM}([0,T]) &= [B^A([0,T]) \oplus B^H([0,T]) \oplus  B^G([0,T]) \\
& \oplus B^O([0,T]) \oplus B^V([0,T])]
\end{aligned}
\end{equation}

To test our framework, we conducted experiments with three multi-sensory datasets (LC823~\cite{wywLC423,wywLC400}, BBSR~\cite{wywBBSR}, Lancaster40k~\cite{wywLan40k}) for the IM and AM paradigms, respectively. 
We used WordSim353~\cite{wywWS353} and SCWS1994~\cite{wywSCWS} as metrics~\cite{wywFramework}. 
The resuls show that integrated representations are closer to human beings than the original ones based on our framework, according to the overall results: 37 submodels outperformed a total of 48 tests for both AM and IM paradigm~\cite{wywFramework}.
Meanwhile, to compare the two paradigms, we introduce concept feature norms datasets which represent concepts with systematic and standardized feature descriptions. 
In this study, we use the datasets McRae~\cite{wywMcRae} and CSLB~\cite{wywCSLB} as criteria.
The findings show that the IM paradigm performs better at multi-sensory integration for concepts with higher modality exclusivity. 
The AM paradigm benefits the concept of uniform perceptual strength distribution. 
Furthermore, we present perceptual strength-free metrics to demonstrate that both paradigms of our framework have excellent generality~\cite{wywFramework}.

\begin{figure}[h!]
  \centering
  \includegraphics[width=8cm]{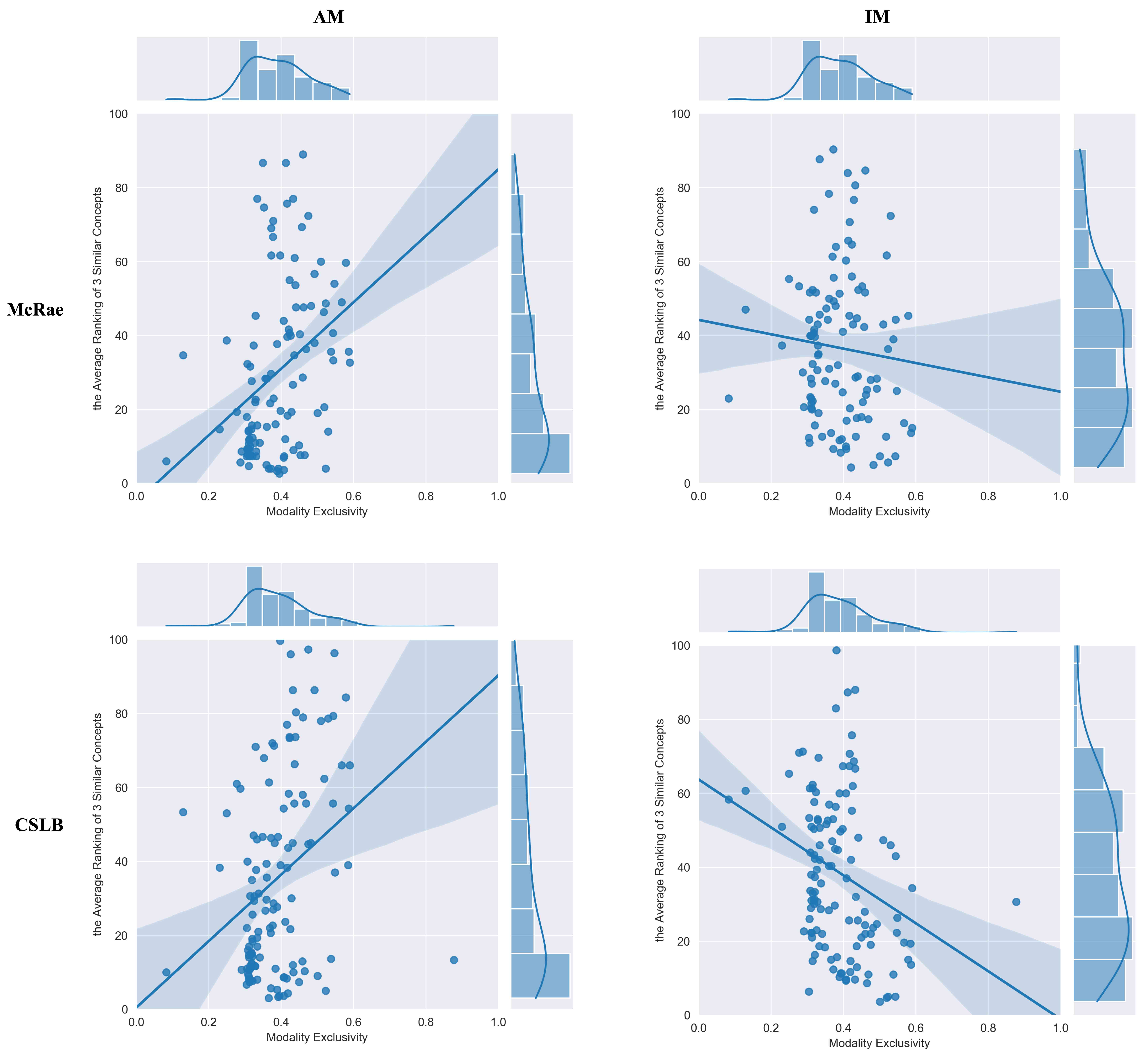}\\
  \caption{The Correlation Results Between Modality Exclusivity and Average of 3 Neighbors' Rankings.}
  \label{MEFS}
\end{figure}

\subsection{Decision Making}
This subsection introduces how BrainCog implements decision-making functions from the perspective of brain neural mechanism modeling and deep SNN-based reinforcement learning models. Using BrainCog, we build a multi-brain areas coordinated SNN model and a spiking deep Q-network to solve decision-making and control problems.

\textbf{1. Brain-inspired Decision-Making SNN}

\begin{figure}[htbp]
\begin{center}
\includegraphics[width=8cm]{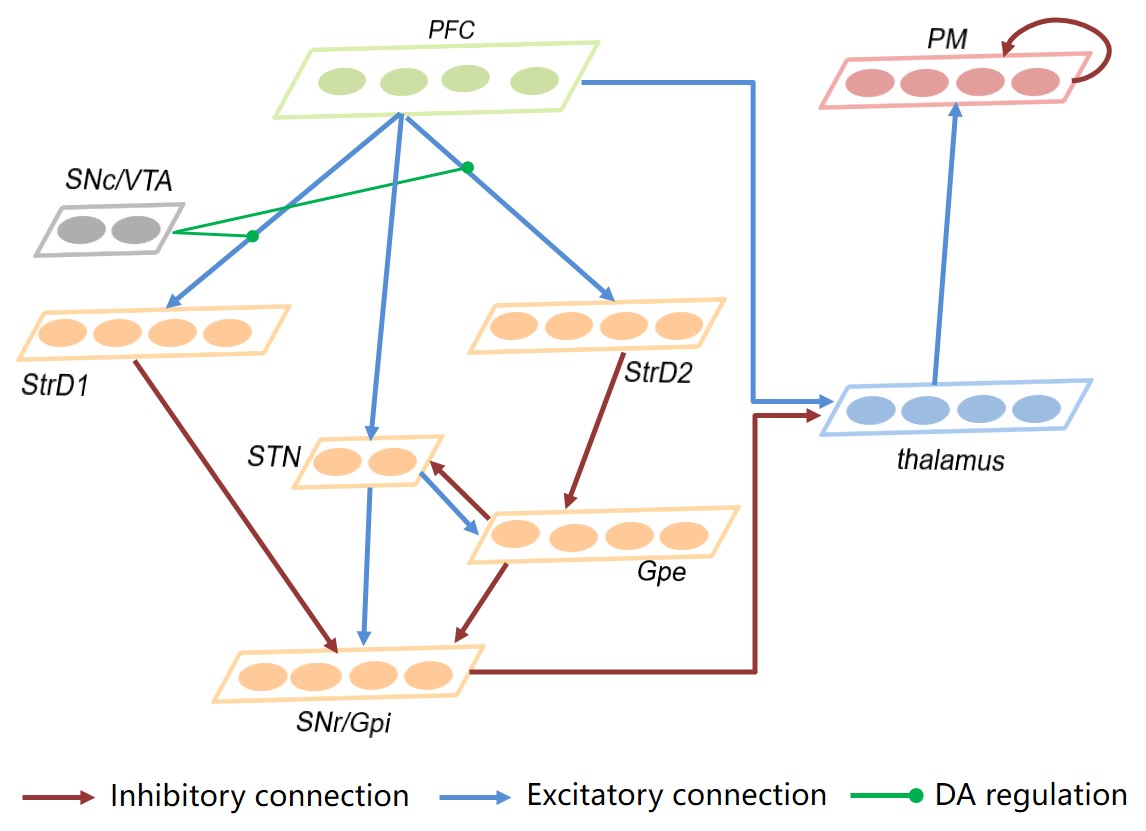}
\end{center}
\caption{The architecture of DA-regulated BDM-SNN, refined based on~\cite{zhao2018brain}.}\label{dmsnn}
\end{figure}

For mammalian brain-inspired decision-making, we take inspiration from the PFC-BG-ThA-PMC neural circuit and build brain-inspired decision-making spiking neural network (BDM-SNN) model~\cite{zhao2018brain} by BrainCog as shown in Fig. ~\ref{dmsnn}. BDM-SNN contains the excitatory and inhibitory connections within the basal ganglia nuclei and direct, indirect, and hyperdirect pathways from the PFC to the BG~\cite{frank2006anatomy,silkis2000cortico}. This BDM-SNN model incorporates biological neuron models (LIF and simplified H-H models), synaptic plasticity learning rules, and interactive connections among multi-brain areas developed by BrainCog. On this basis, we extend the dopamine (DA)-regulated BDM-SNN, which modulates synaptic learning for PFC-to-striatal direct and indirect pathways via dopamine. Different from the DA regulation method in~\cite{zhao2018brain} which uses multiplication to modulate the specified connections, we improve it by introducing R-STDP~\cite{Eugene2007} (from Eq.~\ref{rstdpe} and Eq.~\ref{rstdpw}) to modulate the PFC-to-striatal connections. 

\begin{figure}[htbp]
\begin{center}
\includegraphics[width=8.8cm]{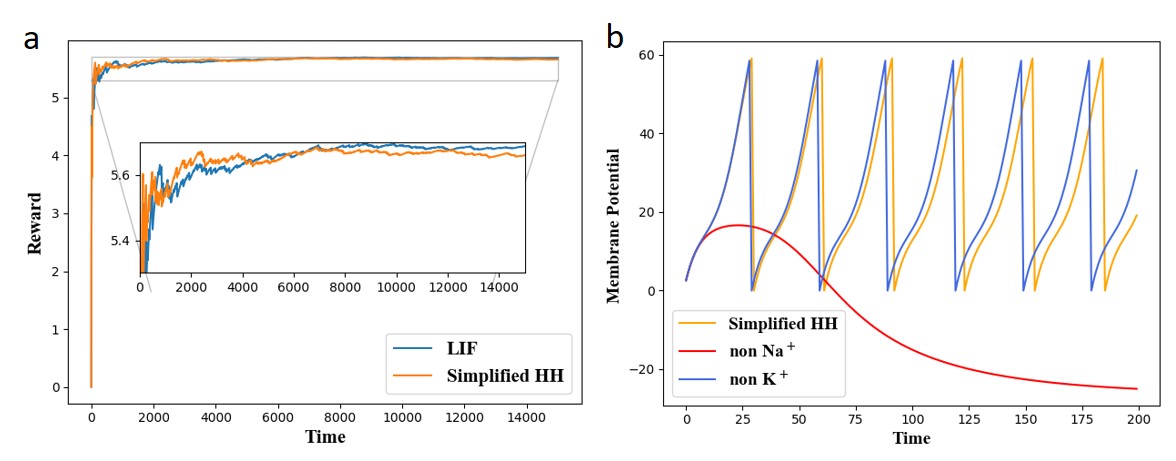}
\end{center}
\caption{(a) Experimental result of BrainCog based BDM-SNN on Flappy Bird. The y-axis is the mean of cumulative rewards. (b) Effects of different ion channels on membrane potential for simplified H-H model.}\label{fb}
\end{figure}

The BDM-SNN model implemented by BrainCog could perform different tasks, such as the Flappy Bird game and has the ability to support UAV online decision-making. For the Flappy Bird game, our method achieves a performance level similar to humans, stably passing the pipeline on the first try. Fig.~\ref{fb}a illustrates the changes in the mean cumulative rewards for LIF and simplified H-H neurons while playing the game. The simplified H-H neuron achieves similar performances to that of LIF neurons. BDM-SNN with different neurons can quickly learn the correct rules and keep obtaining rewards. We also analyze the role of different ion channels in the simplified H-H model. From Fig.~\ref{fb}b, we find that sodium and potassium ion channels have opposite effects on the neuronal membrane potential. Removing sodium ion channels will make the membrane potential decay, while the membrane potential rises faster and fires earlier when removing potassium ion channels. These results indicate that sodium ion channels can help increase the membrane potential, and potassium ion channels have the opposite effect. Experimental results also indicate that BDM-SNN with simplified H-H model that removes sodium ion channels fails to learn the Flappy Bird game.

In addition, for the UAV decision-making tasks in the real scene, our model could perform potential applications including flying over doors and windows and obstacle avoidance, which have been realized in~\cite{zhao2018brain}. Users only need to divide the state space and action space according to different tasks, call the DA-regulated BDM-SNN decision-making model, and combine the UAV's action control instructions to complete the UAV's decision-making process.

This part of the work mainly draws on the neural structure and learning mechanism of brain decision-making and proposes the multi-brain areas coordinated decision-making spiking neural network constructed by BrainCog, and verifies the ability of reinforcement learning in different application scenarios.

\textbf{2. Spiking Deep Q Network with potential based layer normalization}

Deep Q network is widely used for decision-making tasks, and it is required to have SNN based deep Q network for BrainCog so that it can be used for SNN based decision making. We propose potential-based layer normalization spiking deep Q network (PL-SDQN) model to combine SNN with deep reinforcement learning~\cite{sun2022solving}. We use the LIF neuron model in BrainCog to simulate neurodynamics. Deep spiking neural networks are difficult to be applied to reinforcement learning tasks. On the one hand, it is due to the complexity of reinforcement learning task itself, on the other hand, it is challenging to train spiking neural networks and transmit spiking signal characteristics in deep layers. We find that the spiking deep Q network quickly dissipates the spiking signal in the convolutional layer. Inspired by how local environmental potentials influence brain neurons, we propose the potential-based layer normalization (pbLN) method. The $x_t$, the postsynaptic potential of convolution layers, are normalized as 
\begin{equation}
	\hat{x_t} = \frac{x_t-\bar{x}_t}{\sqrt{\sigma_{x_t}+\epsilon}}
	\label{Eq:norm}
\end{equation}	

\begin{equation}
	\bar{x}_t = \frac{1}{H} \sum_{i=1}^Hx_{t, i}
\end{equation}	

\begin{equation}
	\sigma_{x_t} = \frac{1}{H}\sqrt{\sum_{i=1}^H(x_{t, i}-\bar{x}_t)}
\end{equation}

\begin{figure}[h!]
	\centering
	\includegraphics[width=8cm]{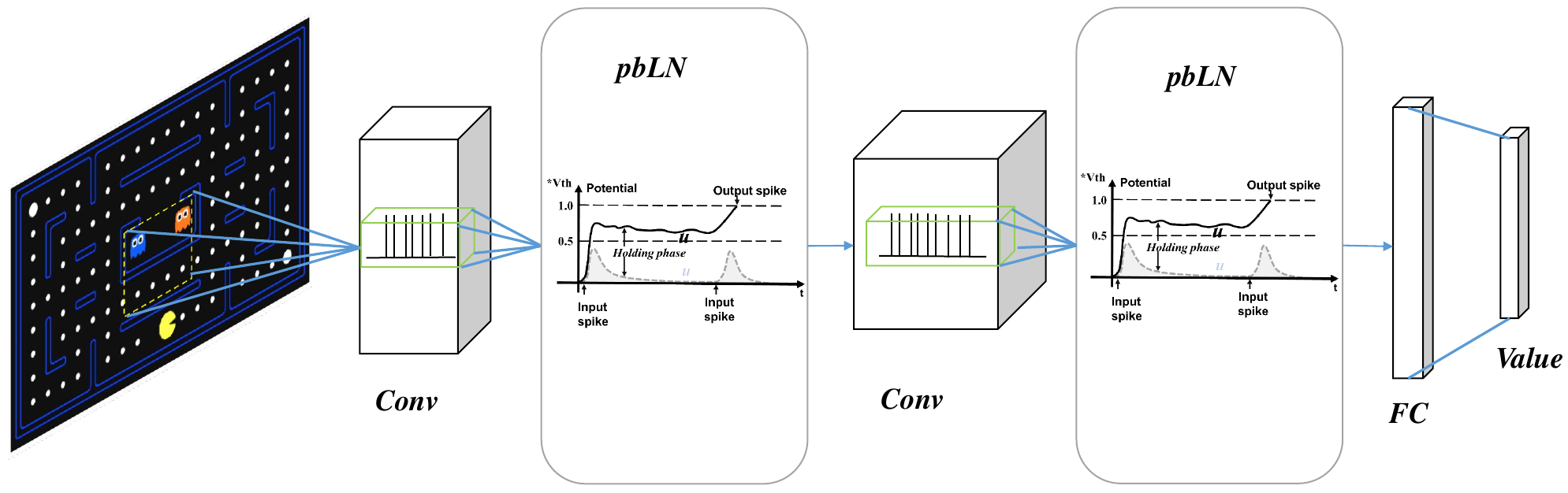}
	\caption{The framework of PL-SDQN. Refined from~\cite{sun2022solving}.}
	\label{Fig:sdqn}
\end{figure}

We construct PL-SDQN model as shown in Fig.~\ref{Fig:sdqn}. Atari game images are processed by spiking convolution network and pbLN method and input into a fully connected LIF neural network. The spiking output of PL-SDQN is weighted and summed to continue state-action values. 

\begin{figure}[h!]
	\centering
	\includegraphics[width=8cm]{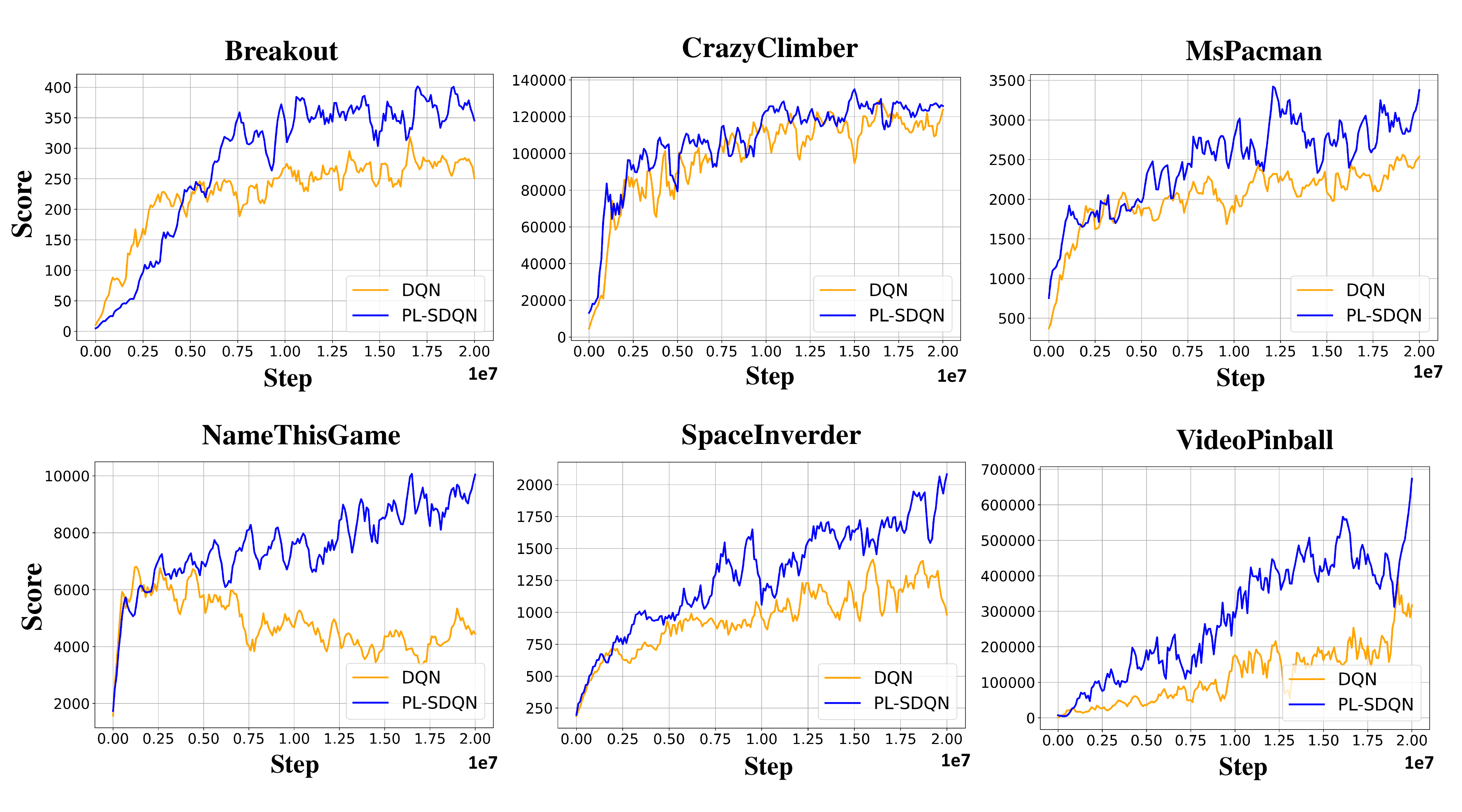}
	\caption{PL-SDQN performance on Atari games, adopted from~\cite{sun2022solving}.}
	\label{Fig:game_res}
\end{figure}

We compared our model with the original ANN-based DQN model, and the results are shown in Fig.~\ref{Fig:game_res}. It shows that our model achieved better performance compared with vanilla DQN model.

\subsection{Motor Control}

Neuromorphic models on robot control can achieve more robust and energy efficient effects than conventional methods. Spiking neural networks have been used in robot control studies like navigation~\cite{wang2014mobile} and robot arm control~\cite{Tieck2018}. Inspired by the brain motor circuit, we construct a multi-brain areas coordinated SNN robot motor control model, to extend BrainCog to control various robots to model embodied intelligence.

\begin{figure}[h!]
	\centering
	\includegraphics[width=8cm]{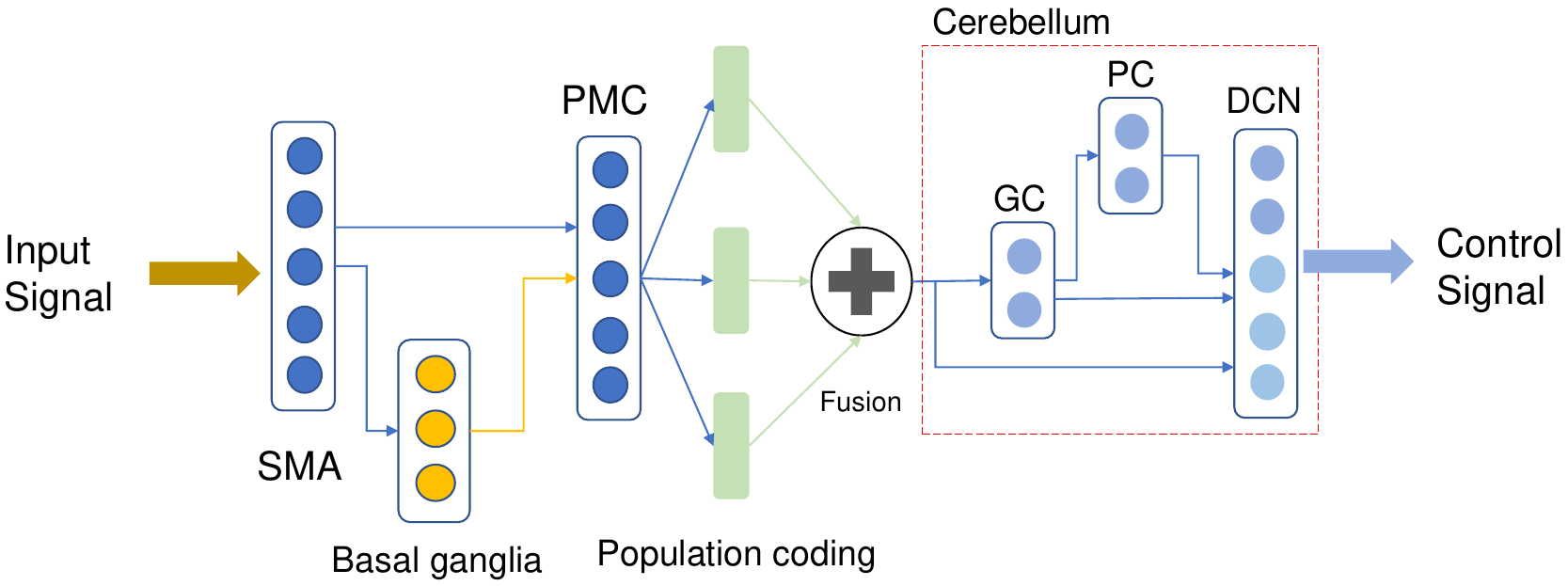}
	\caption{A spiking neural network for motor control based on BrainCog.}
	\label{Fig:motor}
\end{figure}

We construct the brain-inspired motor control model with LIF neuron provided by BrainCog. The whole network model architecture is shown in Fig.~\ref{Fig:motor}. The high-level motion information is produced by SMA and  PMC modules. As discussed above, the function of SMA is to process internal movement stimuli and is responsible for the planning and abstraction of advanced actions. The SMA model contains LIF neurons and receives input signals. One part of the output pulse of the SMA module stimulates the PMC module, and the other part is received by the BG module. The output of the BG module is used as a supplementary signal for action planning, and serves as the input to the PMC together with the SMA signal.

In order to expand the dimension of neuron direction representation, we use neuron population coding to output the high-level action abstraction. Population-coded spiking neural network has been used for energy-efficient continuous control~\cite{tang2021deep}, showing population coding can increase the ability of spiking neurons to represent precise continuous values. In our work, we are inspired by neural mechanisms of population encoding of motion directions in the brain and use LIF neuron groups to process the output spikes from PMC module. 

The cerebellum plays an important role in motor coordination and fine regulation of movements. We built spiking neural network based cerebellum model to process the high-level motor control population embedding. The outputs of populations are fused to encode motor control information generated by high-level cortex area inputs to a three-layer cerebellum spiking neural network, including GCs, PCs and DCN modules. And the cerebellum takes pathway connections like DenseNet~\cite{huang2017densely}. The DCN layer generates the final joint control outputs.

\subsection{Knowledge Representation and Reasoning}
This subsection shows how the BrainCog platform achieve the ability of knowledge representation and reasoning. Via neuroplasticity and population coding mechanisms, spiking neural networks acquire music and symbolic knowledge. Moreover, on this basis, cognitive tasks such as music generation, sequence production, deductive reasoning and inductive reasoning are realized.

\textbf{1. Music Memory and Stylistic Composition SNN}

Music is part of human nature. Listening to melodies involves sensory perception, personal memory, action, emotion, and even creative behaviors, etc~\cite{Koelsch2012}. Music memory is a fundamental part of musical behaviors, and humans have strong abilities to store a sequence of notes in the brain. Learning and creating music are also essential processes. A musician engages his memory, emotion, musical knowledge and skills to write a beautiful melody. Actually, neuroscientists have found that many brain areas need to collaborate to complete the cognitive behaviors with music. Inspired by brain mechanisms, this paper focuses on the two key issues of music memory and composition, which are modeled by spiking neural networks based on the BrainCog platform.

\subsubsection{Musical Memory Spiking Neural Network} \label{secA}
A musical melody is composed of a sequence of notes. Pitch and duration are two essential attributes of a note. Scientists have found that the primary auditory cortex provides a tonotopic map to encode the pitches, which means that neurons in this region have their preferences of pitches~\cite{Kalat2015}. Meanwhile, neural populations in the medial premotor cortex have the preferences of all the time intervals covered in hundreds of milliseconds~\cite{merchant2013a}. Besides, researchers have emphasized the contribution of the hippocampus in sequence memory~\cite{Fortin2002}. Inspired by these mechanisms, this work proposes a spiking neural model, which contains collaborated subnetworks to encode, store and retrieve the music melodies~\cite{LQ2020}.

\begin{figure}[!htbp]
    \centering
    \includegraphics[scale=0.4]{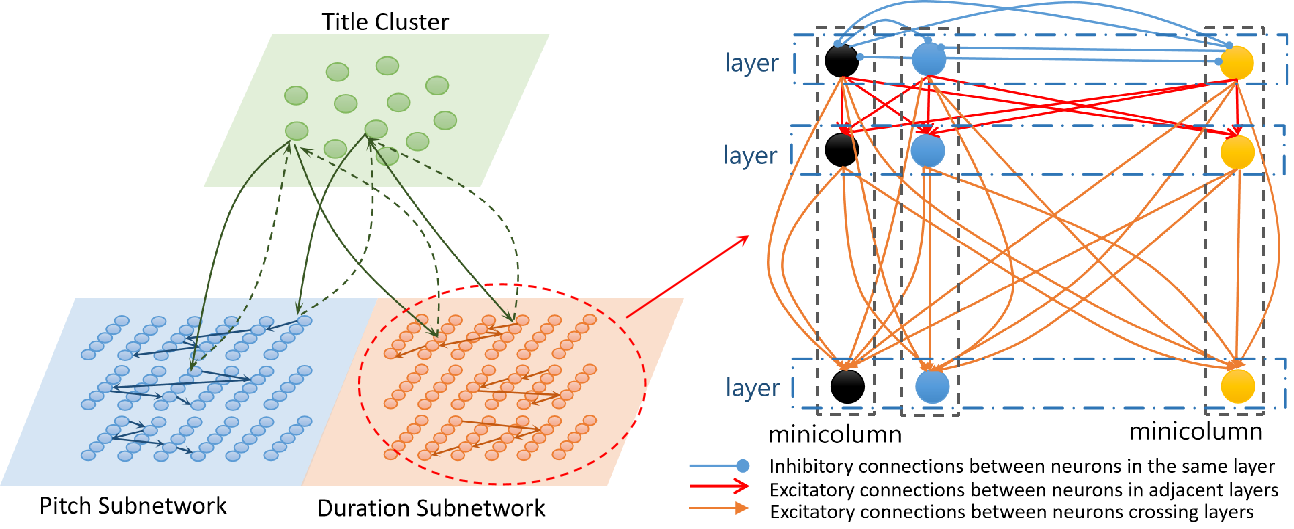}
    \caption{The architecture of the music memory model, refined based on~\cite{LQ2020}.}
    \label{01}
\end{figure}

\emph{Encoding:} As is shown in Fig.~\ref{01}, this work defines pitch subnetwork and duration subnetwork to encode pitches and durations of musical notes respectively. These two subnetworks are composed of numbers of minicolumns with different preferences. Synaptic connections with transmission delays exist between neurons from different layers. Besides, a cluster that represents the title of a musical melody is composed of numerous individual neurons. This cluster has the feedforward and feedback connections with pitch and duration subnetworks. Since the BrainCog platform supports various neural models, this work takes LIF model to simulate neural dynamics.
 
\emph{Storing:} Based on the encoding process, as the notes input sequently, the neurons in pitch and duration subnetworks with different preferences respond to these sequential notes and fire orderly. Meanwhile, connections between these neurons are computed and updated by the STDP learning rule. It is important to indicate that the neurons and synapses are grown dynamically. Besides, synaptic connections between the title cluster and other two subnetworks are generated and updated by the STDP learning rule simultaneously. The details of note sequence memorizing can be found in our previous work~\cite{LQ2020}.

\emph{Retrieving:} Given the title of a musical work, the ordered notes can be recalled accurately. Since the weights of connections are updated in storing process, neural activities in the title cluster lead to the excitations of neurons in pitch and duration subnetworks. Then, the notes are retrieved in order. We use a public corpus that contains 331 classical piano works~\cite{Krueger2018} recorded by MIDI standard format to evaluate the model. The experiments have shown that our model can memorize and retrieve the melodies with an accuracy of 99\%. The details of the experiments have been discussed in our previous work~\cite{LQ2020}.

\subsubsection{Stylistic Composition Spiking Neural Network}

How to learn and make music are quite complex processes for humans. Scientists have found that the memory system and knowledge experience participate in human creative behaviors~\cite{Dietrich2004}. Many brain areas like the prefrontal cortex are engaged in human creativity~\cite{Jung2013}. However, the details of brain mechanisms are still unclear. Inspired by the current neuroscientific findings, the BrainCog introduces a spiking neural network for learning musical knowledge and creating melodies with different styles~\cite{LQ2021}. 

\begin{figure}[!htbp]
    \centering
    \includegraphics[scale=0.28]{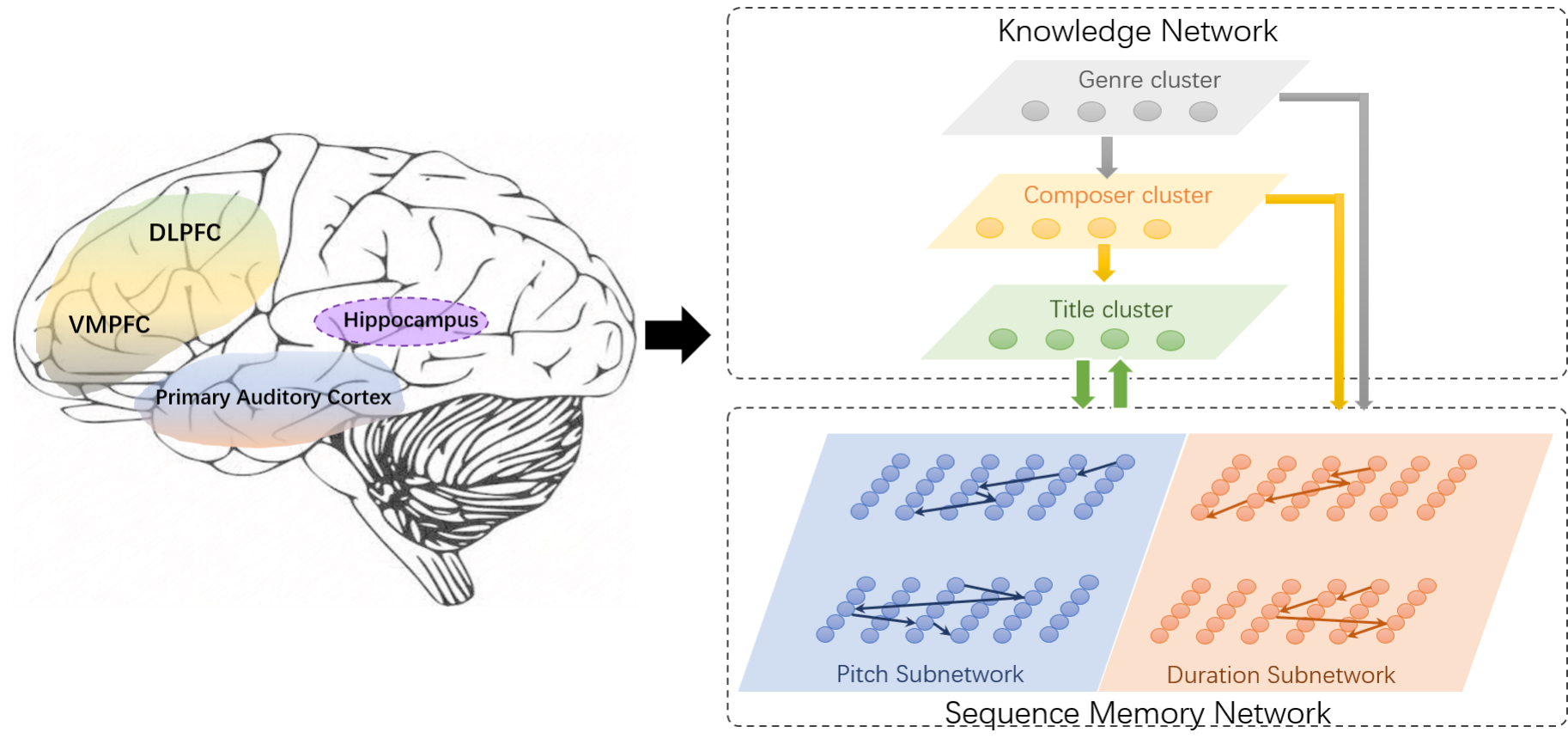}
    \caption{Stylistic composition model inspired by brain mechanisms. Refined based on~\cite{LQ2021}.}
    \label{02}
\end{figure}

\emph{Musical Learning:} This work proposes a spiking neural model which is composed of a knowledge network and a sequence memory network. As is shown in Fig.~\ref{02}, the knowledge network is designed as a hierarchical structure for encoding and learning musical knowledge. These layers store the genre (such as Baroque, Classical, and Romantic), the names of famous composers and the titles of musical pieces. Neurons in the upper layers project their synapses to the lower layers. The sequence memory network stores the ordered notes which have been discussed in section~\ref{secA}. During the learning process, synaptic connections are also projected from the knowledge network to the sequence memory network. This work also takes LIF model which is supported by the BrainCog platform to simulate neural dynamics. Furthermore, all the connections are generated and updated dynamically by the STDP learning rule. 

\begin{figure}[htbp]
    \centering
    \includegraphics[scale=0.6]{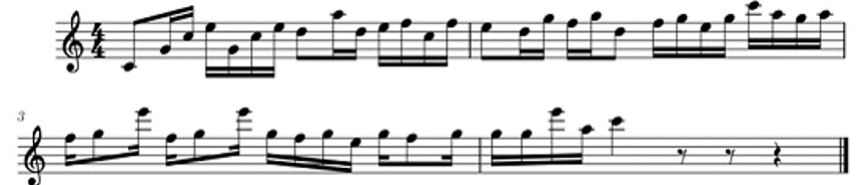}
    \caption{A sample of a generated melody with Bach's characteristic.}
    \label{03}
\end{figure}

\emph{Musical Composition:} Based on the learning process, genre-based and composer-based melody compositions are discussed in this paper. Given the beginning notes and the length of the melody to be generated, the genre-based composition can produce a single-part melody with a specific genre style. This task is achieved by the neural circuits of genre cluster and sequential memory system. Similarly, the composer-based composition can produce melodies with composers' characters. The composer cluster and sequential memory system circuits contribute to this process~\cite{LQ2021}. We also use a classical piano dataset including 331 musical works recorded by MIDI format~\cite{Krueger2018} to train the model. Fig.~\ref{03} shows a sample of the generated melody with Bach's style. The details of stylistic composition can be referred to our previous work~\cite{LQ2021}.

A total of 41 human listeners are invited to evaluate the quality of the generated melodies, and they are divided into two groups, one of which has a musical background. Experiments have shown that the pieces produced by the model have strong characteristics of different styles and some of them sound nice.

\textbf{2. Brain-Inspired Sequence Production SNN}

Sequence production is an essential function for AI applications. Components in BrainCog enable the community to build SNN models to handle this task. In this paper, we introduce the brain-inspired symbol sequences production spiking neural network (SPSNN) model that has been incorporated in BrainCog~\cite{fang2021spsnn}. SPSNN incorporates multiple neuroscience mechanisms including Population Coding~\cite{xie2022geometry}, STDP~\cite{dan2004spike}, Reward-Modulated STDP~\cite{fremaux2016neuromodulated}, and Chunking Mechanism~\cite{pammi2004chunking}, mostly covered and provided by BrainCog. After reinforcement learning, the network can complete the memory of different sequences and production sequences according to different rules.

\begin{figure}[!htbp]
\centering
\includegraphics [width=8.9cm]{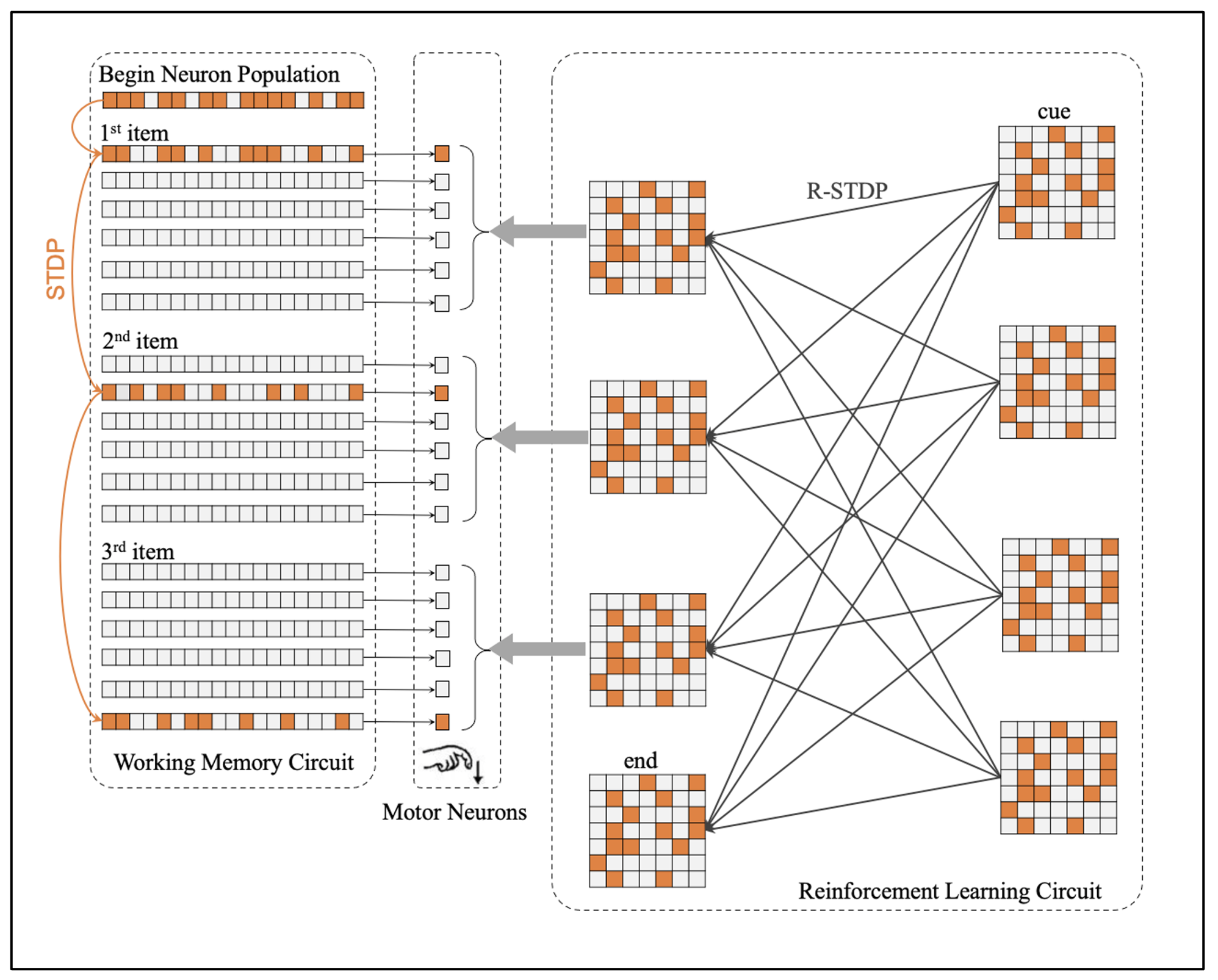}
\caption{The architecture of SPSNN, adopted from~\cite{fang2021spsnn}.}
\label{SPSNN}
\end{figure}

For Population Coding, this model utilizes populations of neurons to represent different symbols. The whole neural loop of SPSNN is divided into Working Memory Circuit, Reinforcement Learning Circuit, and Motor Neurons~\cite{fang2021spsnn}, shown in Fig.~\ref{SPSNN}. The Working Memory Circuit is mainly responsible for completing the memory of the sequence. The Reinforcement Learning Circuit is responsible for acquiring different rules during the reinforcement learning process. The Motor Neurons can be regarded as the network's output. 

In the working process of the model, the Working Memory Circuit and the Reinforcement Learning Circuit cooperate to complete the memory and production of different sequences~\cite{fang2021spsnn}. It is worth mentioning that with the increase of background noise, the recall accuracy of symbols at different positions in a sequence gradually decreases, and the overall change trend follows the "U-shaped accuracy", which is consistent with experiments in psychology and neuroscience~\cite{jiang2018production}. The results are highly consistent due to the superposition of primacy and recency effects. Our model provides a possible explanation for both effects from a computational perspective.

\textbf{3. Commonsense Knowledge Representation Graph SNN}

Commonsense knowledge representation and reasoning are important  cornerstones  on the way to realize human-level general  AI~\cite{minsky2007emotion}. In this module, we build Commonsense Knowledge Representation SNN(CKR-SNN) to explore whether SNN can complete these cognitive function.

The hippocampus plays a critical role in the formation of new knowledge memory~\cite{schlichting2017hippocampus}. Inspired by the population coding mechanism found in hippocampus~\cite{ramirez2013creating},  this module encodes the entities and relations of commonsense knowledge graph into different  populations of neurons. Via spiking timing-dependent plasticity (STDP) learning principle, the synaptic connections between neuron populations are formed after guiding the sequential firings of corresponding  neuron populations~\cite{KRRfang2022}.

\begin{figure}
\centering
\includegraphics [width=8.9cm]{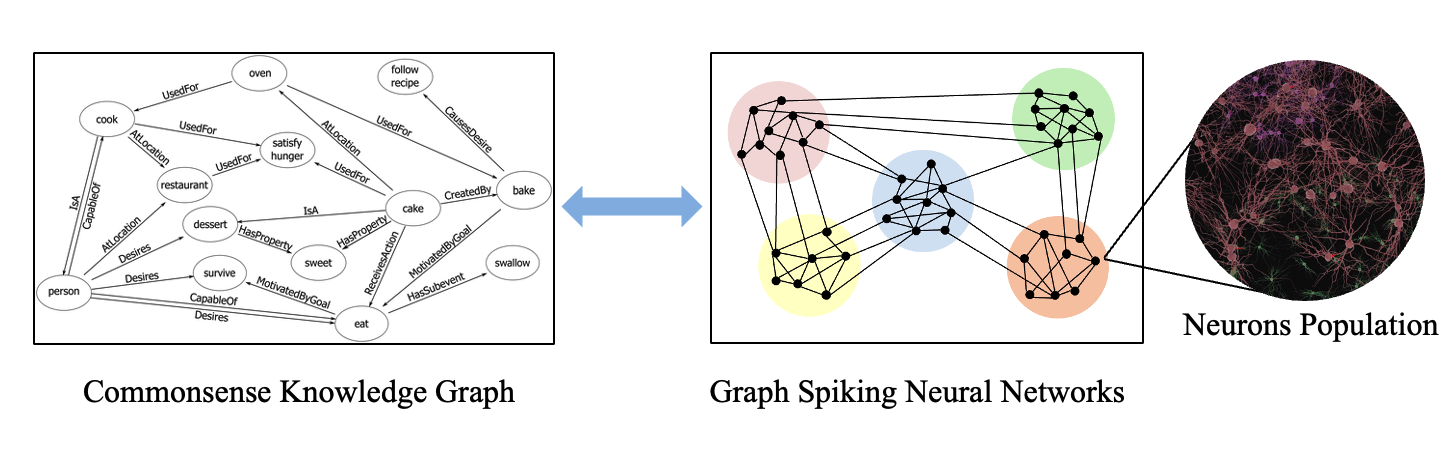}
\caption{Graph Spiking Neural Networks for Commonsense Representation, adopted from~\cite{KRRfang2022}.}
\label{GSNN}
\end{figure}

As Fig.~\ref{GSNN} shows, neuron populations together constructed the giant graph spiking neural networks, which contain the commonsense knowledge. In this module, Commonsense Knowledge Representation SNN(CKR-SNN) represents a subset of Commonsense Knowledge Graph ConceptNet~\cite{Conceptnet55}. After training, CKR-SNN can complete conceptual  knowledge generation and other cognitive tasks~\cite{KRRfang2022}.

\textbf{4. Causal Reasoning SNN}

In BrainCog, we constructed causal reasoning SNN, as an instance to verify the feasibility of spiking neural networks to realize deductive and inductive  reasoning. Specifically, Causal Reasoning Spiking Neural Network (CRSNN) module contains a brain-inspired causal reasoning spiking neural network model~\cite{fang2021crsnn}.

\begin{figure}
\centering
\includegraphics [width=8.8cm]{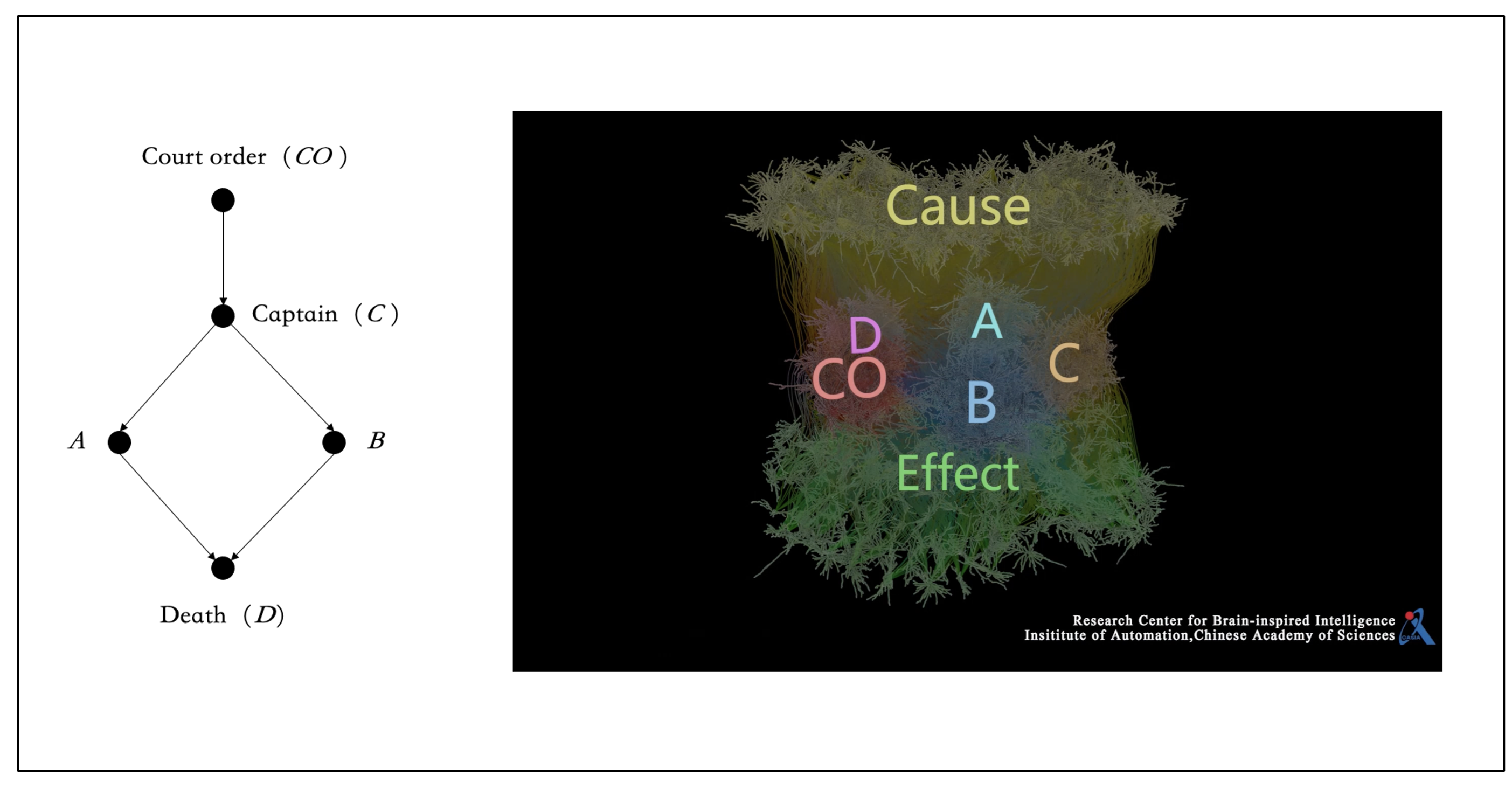}
\caption{CRSNN 3D model, adapted from~\cite{fang2021crsnn}.}
\label{CRSNN}
\end{figure}

This model explores how to encode a static causal graph into a spiking neural network and implement subsequent reasoning based on a spiking neural network. Inspired by the causal reasoning process of the human brain~\cite{pearl2018book}, we try to explore how causal reasoning can be implemented based on spiking neural networks. The 3D model of CRSNN is shown in Fig.~\ref{CRSNN}.

Inspired by neuroscience, the CRSNN module adopts the population coding mechanism and uses neuron populations to represent nodes and relationships in the causal graph. Each node indicates different events in the causal graph, as shown in Fig.~\ref{CRSNN}. By giving current stimulation to different neuron populations in the spiking neural networks and combining the STDP learning rule~\cite{dan2004spike}, CRSNN can encode the topology between different nodes in a causal graph into a spiking neural network. Furthermore, according to this network, CRSNN completes the subsequent deductive reasoning tasks.

Then, by introducing an external evaluation function, we can grasp the specific reasoning path in the working process of the network according to the firing patterns of the model, which gives the CRSNN more interpretability compared to traditional ANN models~\cite{fang2021crsnn}.

\subsection{Social Cognition}

The nature and neural correlates of social cognition is an advanced topic in cognitive neuroscience. In the field of artificial intelligence and robotics, there are few in-depth studies that take the neural correlation and brain mechanisms of biological social cognition seriously. Although the scientific understanding of biological social cognition is still in a preliminary stage~\cite{zeng2018toward}, we integrate the biological findings of social cognition into a model to construct a brain-inspired model for social cognition to extend the functions of BrainCog.

Understanding ourselves and other people is a prerequisite for social cognition. 

An individual's perception of the world is realized through his own body, and the importance of  knowing oneself is the perception of self-body. Neuroscientific researches show that the inferior parietal lobule (IPL) is activated when the subjects see self-generated actions~\cite{macuga2011selective} and their own faces~\cite{sugiura2015neural}. Similar to the IPL, the Insula is activated in bodily ownership and self-recognition tasks~\cite{craig2009you}. 

Understanding the mental states of others plays an important role for understanding other people. Theory of mind is an ability to distinguish between self and others and to infer others' mental states (such as desires, goals, beliefs, etc.) in the social context~\cite{shamay-tsoory_dissociation_2007,sebastian_neural_2012,dennis2013cognitive}. This ability can help us reasonably infer other people's policies and goals. Inspired by this, we believe that applying theory of mind to the agent's decision-making process will improve the agent's inference of other agents, so as to take more reasonable actions. Neuroscientific researches~\cite{abu-akel_neuroanatomical_2011,hartwright_multiple_2012,hartwright_special_2015, koster-hale_theory_2013} show that the brain areas related to theory of mind are mainly TPJ, part of PFC, ACC and IFG. The IPL contained in the TPJ is mainly used to represent self-relevant information, while the pSTS is used to represent information related to others. The insula, representing the abstract self, can be stimulated with self-related information~\cite{zeng2018toward}. When theory of mind is going on, the IFG will suppress self-relevant information. Therefore, the TPJ will input other-relevant information into the PFC. The ACC evaluates the value of others' states, so as to help the PFC to infer others. The process of inferring others' goals or behaviors can be understood as simulating other people's decision-making~\cite{suzuki_learning_2012}. Therefore, this process will be regulated by dopamine from substantia nigra compacta/ventral tegmental area (Snc/VTA).



\begin{figure}[htbp]
\begin{center}
\includegraphics[width=8cm]{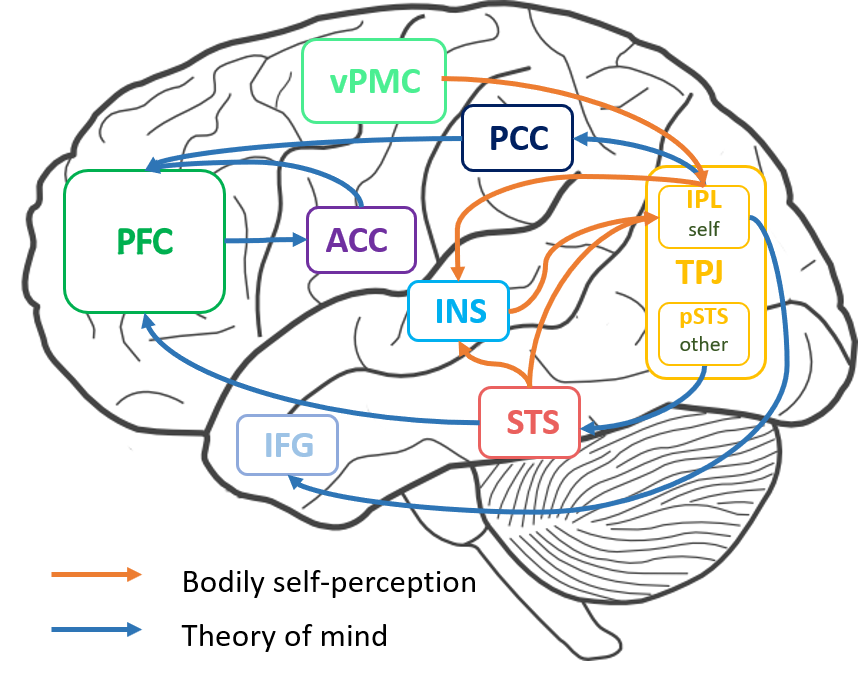}
\end{center}
\caption{Brain-inspired social cognition model.}
\label{BISC}
\end{figure}

With the neuron model and STDP function provided by the BrainCog framework, a brain-inspired social cognition model is constructed, as shown in Fig.~\ref{BISC}.


The brain-inspired social cognition model contains two pathways: the bodily self-perception pathway and the theory of mind pathway.

\begin{figure}[htbp]
\begin{center}
\includegraphics[width=8cm]{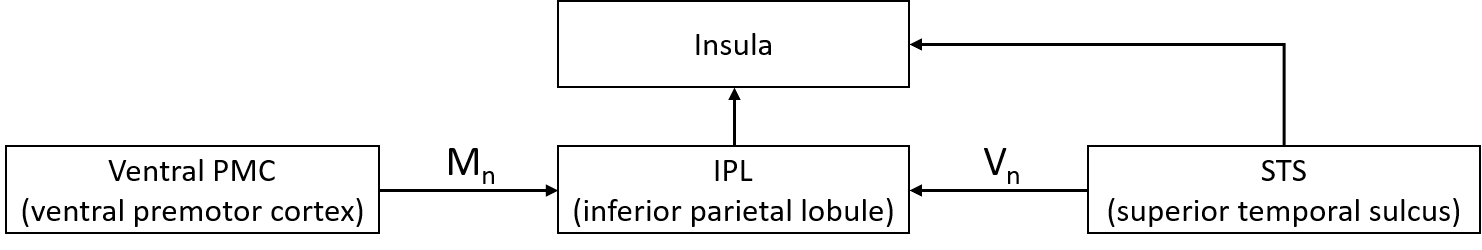}
\end{center}
\caption{The architecture and pathway of bodily self-perception in the brain-inspired social cognition model.}
\label{Self}
\end{figure}

The bodily self-perception pathway (shown in Fig.~\ref{Self}) consists of inferior parietal lobule spiking neural network (IPL-SNN) and Insula spiking neural network (Insula-SNN). The IPL-SNN realizes motor-visual associative learning. The Insula-SNN realizes the abstract representation of oneself, that is, when the detected movement's visual results match the expected results of its own movement, the Insula will be activated, and the robot considers that the moving part in the field of vision belongs to itself.

\begin{figure}[htbp]
\begin{center}
\includegraphics[width=8cm]{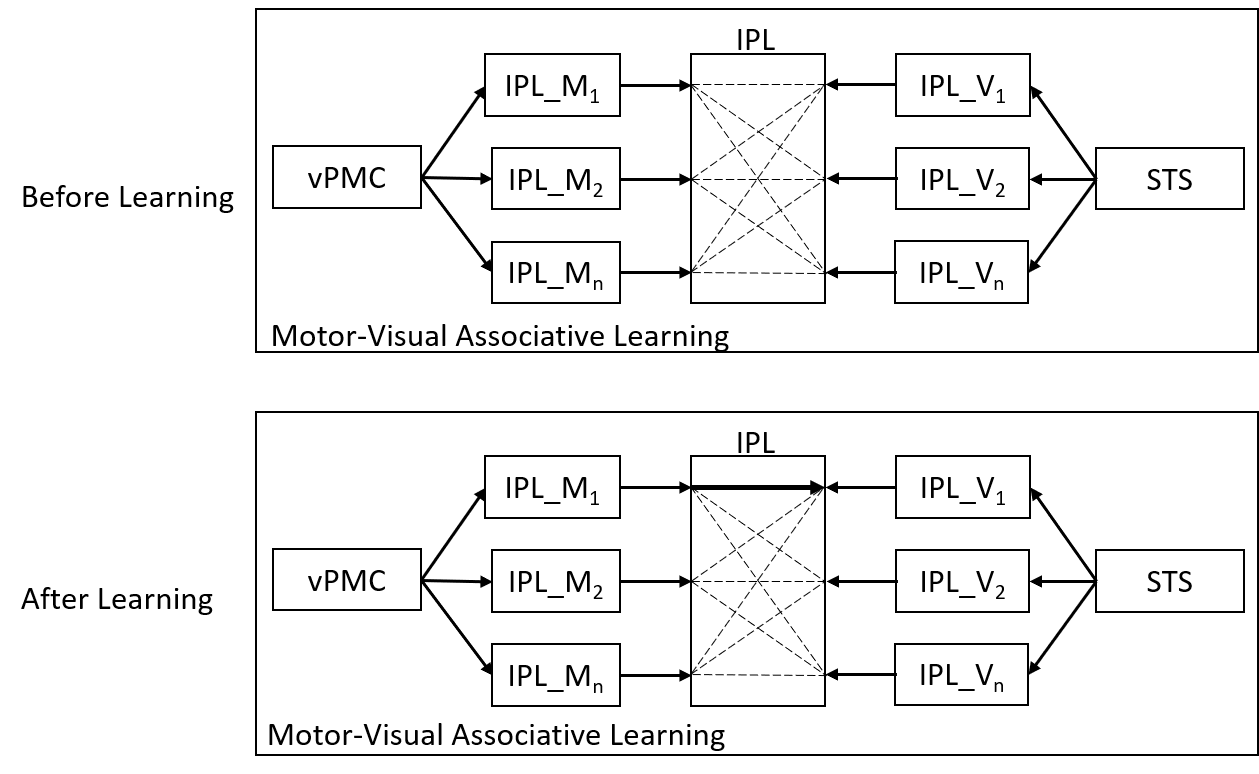}
\end{center}
\caption{Motor-visual associative learning in IPL, adapted from~\cite{zeng2018toward}.}
\label{IPL}
\end{figure}

The architecture of IPL-SNN and the process of motor-visual associative learning is shown in Fig.~\ref{IPL}. The vPMC generates its own motion angle information, and the STS outputs the motion angle information detected by vision. According to the STDP mechanism and the spiking time difference of neurons in IPLM and IPLV, the motor-visual associative learning is established. 

\begin{figure}[htbp]
\begin{center}
\includegraphics[width=8cm]{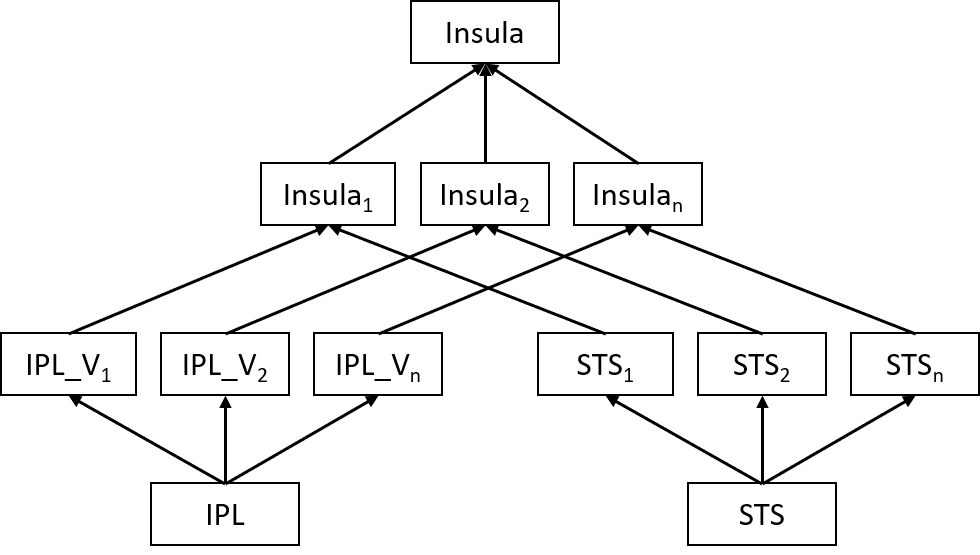}
\end{center}
\caption{The architecture of Insula-SNN.}
\label{Insula}
\end{figure}

The architecture of Insula-SNN is shown in Fig.~\ref{Insula}. The Insula receives angle information from IPLV and STS. After the motor-visual associative learning in IPL, IPLV outputs the visual feedback angle information predicted according to its own motion, and STS outputs the motion angle information detected by vision. If the two are consistent, the Insula will be activated.

\begin{figure}[!htbp]
\centering
\includegraphics[width=0.48\textwidth]{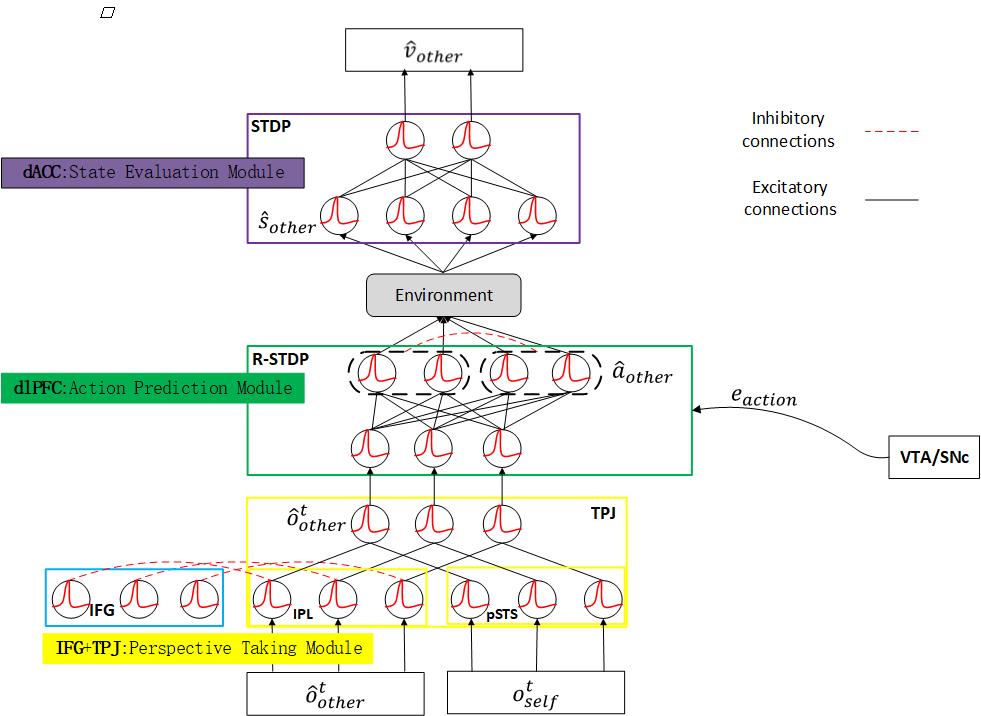}
\caption{The architecture of theory of mind in the brain-inspired social cognition model, refined based on~\cite{zhao2022brain}.}
\label{ToM}
\end{figure}

The theory of mind pathway~\cite{zhao2022brain} is mainly composed of three modules: the perspective taking module, the policy inference module, the action prediction module, and the state evaluation module (shown in Fig.~\ref{ToM}).

The perspective taking (also called self-perspective inhibition~\cite{zeng2018toward}) module simulates the function of suppressing self-relevant information in the process of distinguish self and others. The information related to self can stimulate an representation of abstract self. When we infer others, the information related to self can be suppressed. Assuming that the agent knows the environment. When the agent with the theory of mind (ToM) infers the observation of others, it only needs to bring its own observation into the position of others. A matrix is used to represent the observed environment, where 1 indicates that the area can be observed at the location, and 0 indicates that the area cannot be observed at the location. Another matrix is used to represent the position of objects. The position which is occupied by objects is represented by 1, otherwise it is represented by 0. By taking the intersection of the two matrices, the estimation of others' states are obtained. According to the fact that the IFG helps the brain to suppress the performance of self-relevant information in the process of ToM, agent with the ToM will inhibit its own representation of states, and further infer the behavior of others by these estimation of others' states. In summary, the input of this module is the observation vector and the matrix of the environment. The output is the observation vector of others' perspective.

The dorsolateral prefrontal cortex (DLPFC) has the function of storing working memory and predicting others' behaviors. The action prediction module is used to simulate the function of the DLPFC. The input is formed by the observation value of others' state output by the perspective taking module. The module is a single layer spiking neural network. There is lateral suppression in the output layers. The network is trained by R-STDP. The source of reward is the difference between the predicted value and the real value. When the predicted value is consistent with the real value, the reward is positive. When the predicted value is not equal to the real value, the reward is negative.

The state evaluation module composed of a single layer spiking neural network simulates the function of the ACC brain area. The inputs of the module are the predicted state and the output is safe or unsafe.

Finally, we conducted  two experiments to test the brain-inspired social cognition model.

\textbf{1. Multi-Robots Mirror Self-Recognition Test }

%
The mirror test is the most representative test of social cognition. Only a few animals passed the test, including chimpanzees~\cite{RN826}, orangutans~\cite{RN827}, bonobos~\cite{RN828}, gorillas~\cite{RN829,RN830}, Asiatic elephant~\cite{RN831}, dolphins~\cite{RN832}, orcas~\cite{RN833}, macaque monkeys~\cite{RN835}, etc. Based on the mirror test, we proposed the Multi-Robots Mirror Self-Recognition Test~\cite{zeng2018toward}, in which three robots with identical appearance move their arms randomly in front of the mirror at the same time, and each robot needs to determine which mirror image belongs to it. The experiment includes training stage and test stage.

\begin{figure}[htbp]
\begin{center}
\includegraphics[width=8cm]{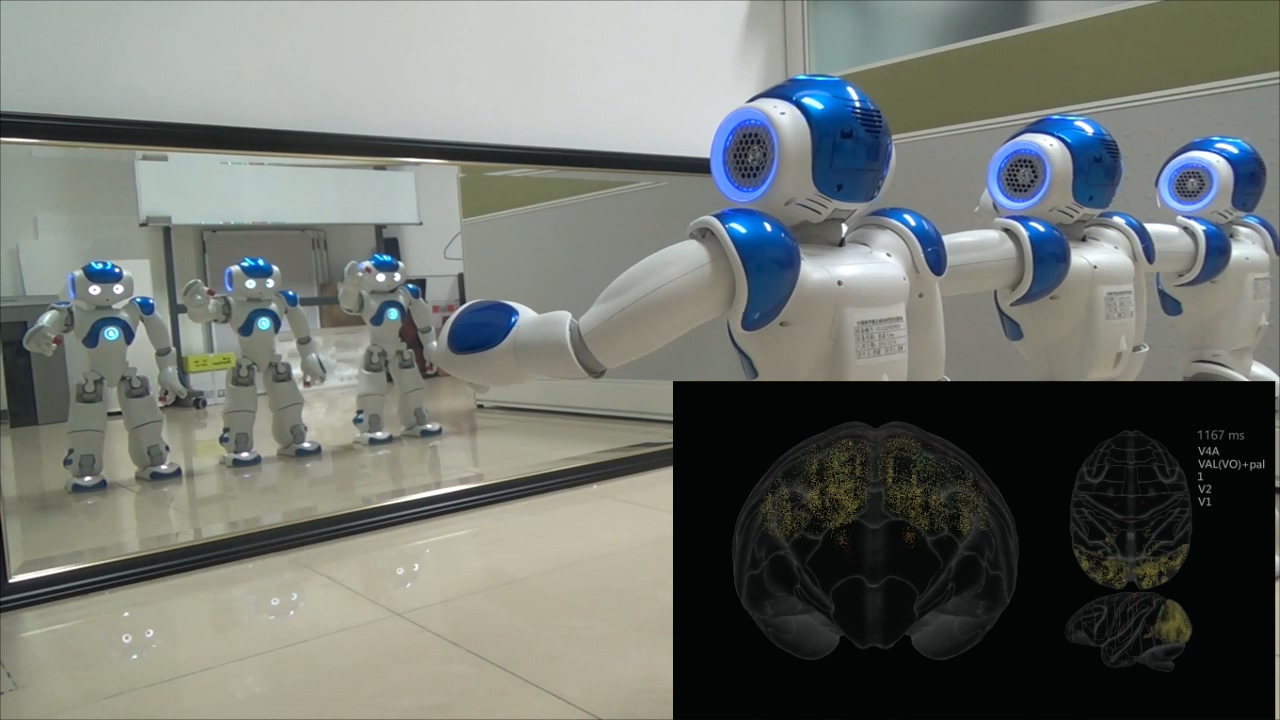}
\end{center}
\caption{Training stage in multi-robots mirror self-recognition test, adapted from~\cite{zeng2018toward}.}
\label{train}
\end{figure}

The training stage is shown in Fig.~\ref{train}. Three blue robots with identical appearance move randomly in front of the mirror at the same time. Each robot establishes the motor-visual association according to the motion angle of its own arm and the angle of visual detection.

\begin{figure}[htbp]
\begin{center}
\includegraphics[width=8cm]{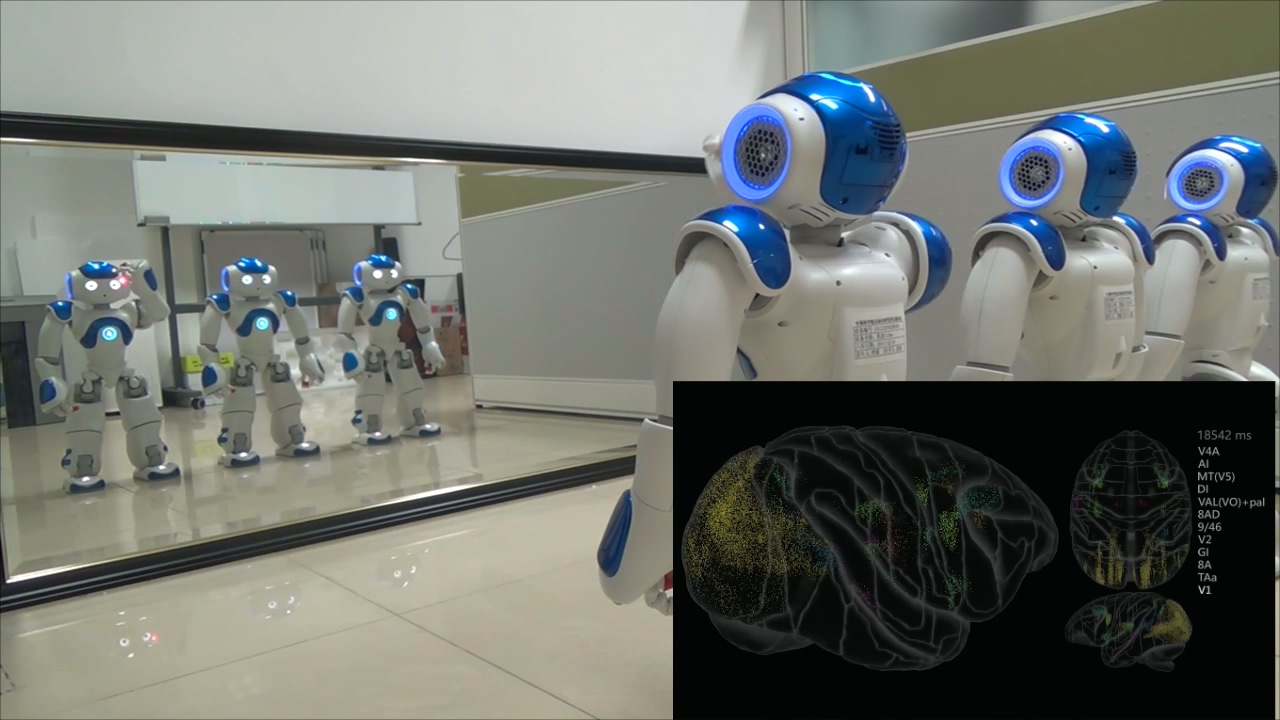}
\end{center}
\caption{Test stage in multi-robots mirror self-recognition test, adapted from~\cite{zeng2018toward}.}
\label{test}
\end{figure}

The test stage is shown in Fig.~\ref{test}. In the test stage, the robot can predicts the visual feedback generated by its arm movement according to the training results. By comparing the similarity between the predicted visual feedback and the detected visual results, the robot can identify which mirror image belongs to it. 


\begin{figure}[htbp]
\begin{center}
\includegraphics[width=8cm]{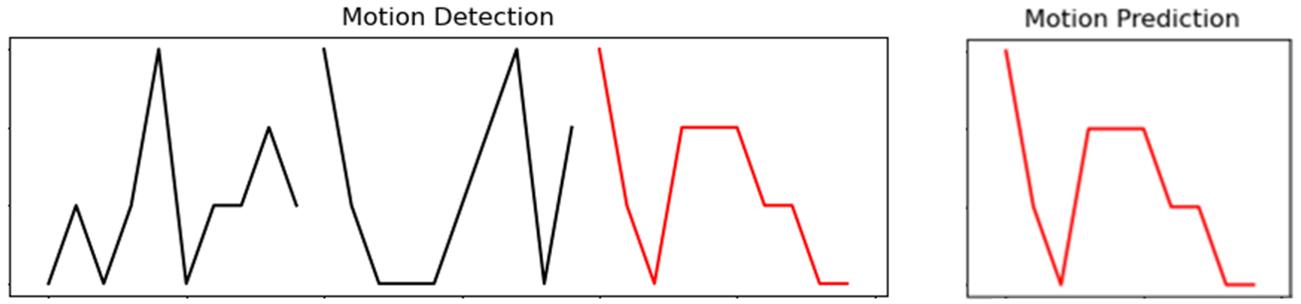}
\end{center}
\caption{The result of IPL-SNN.}
\label{result}
\end{figure}

In the bodily self-perception pathway, the input is the angle of the robot's random motion and the angle detected by the robot's vision. After training and testing, the output is an image, which is the result of visual motion detection and the result of self motion prediction. The motion track corresponding to the red line in the visual motion detection result is generated by itself. The result is shown in Fig.~\ref{result}.

\textbf{2. AI Safety Risks Experiment }

\begin{figure}[!htbp]
\centering
\subfloat[Green agent observes others' behaviors (example 1)]{\includegraphics[height=0.65in]{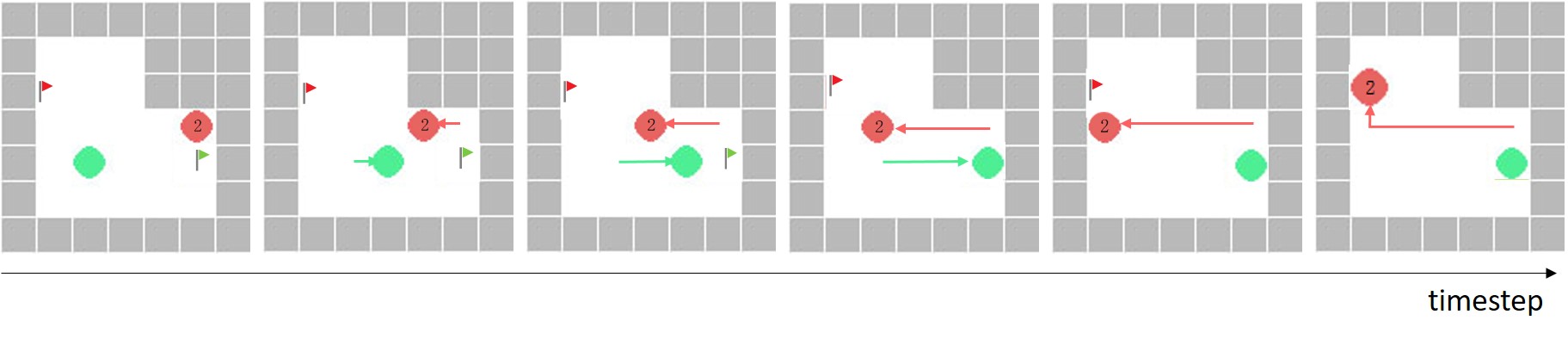}%
\label{1}}
\hfil
\subfloat[Green agent observes others' behaviors (example 2)]{\includegraphics[height=0.65in]{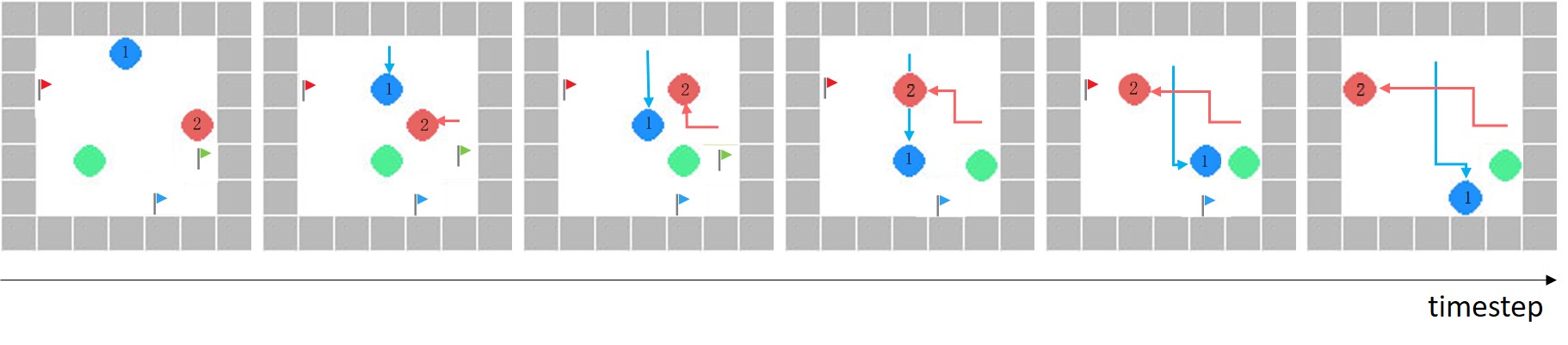}%
	\label{2}}
\hfil
\subfloat[Test example 1 (with ToM)]{\includegraphics[height=0.65in]{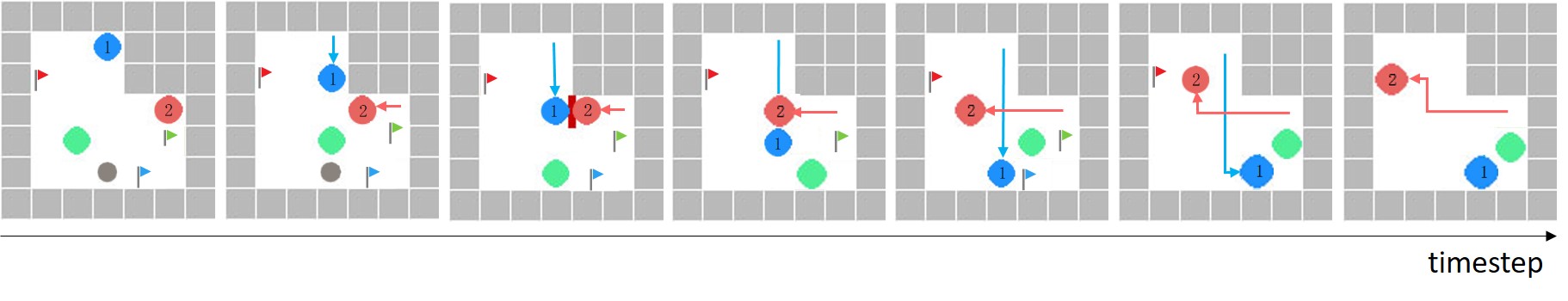}%
	\label{3}}
\hfil
\subfloat[Test example 2 (without ToM)]{\includegraphics[height=0.65in]{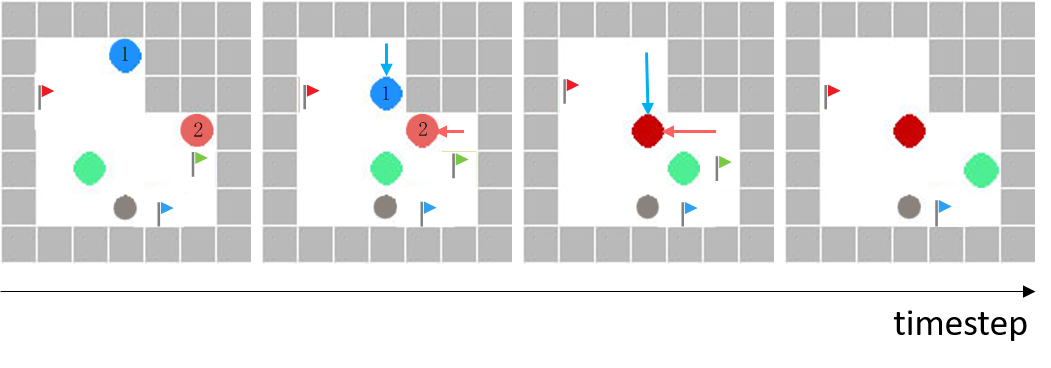}%
	\label{4}}
\caption{Comparison diagram of experimental results. (a) Example 1. The green agent observes others’ behaviors. (b) Example 2. The green agent observes others’ behaviors. (c) The green agent with ToM can help other agents avoid risks. (d) The green agent without ToM is unable to help other agents avoid risks. Similar results can be found in~\cite{zhao2022brain}.}
\label{fig_sim}
\end{figure}

AI safety risks experiment is shown in Fig.~\ref{fig_sim}. After observing the behavior of the other two agents, the green agent can infer the behaviors of others by utilizing its ToM ability when safety risks may arise due to environmental changes. The experiment was conducted in the environments with some simulated types of safety risks (e.g., the intersection will block the view of agents and may cause agents to collide in the crossing).

The experiment shows that the agent can infer others when they have different perspectives. In the first two environments, the agent observes the movement of others, and in the third environment, the agent predicts others' actions. In the experiment, we verify the effectiveness of the model by taking the rescue behavior as the standard when other agents might be in danger. The experimental results show that the agent with the ToM can predict the danger of the other agent in a slightly changed environment after watching other agents move in the previous environments.

\section{Brain Simulation}
Brain simulation includes two parts: brain cognitive function simulation and multi-scale brain structure simulation. We incorporate as much published anatomical data as possible to simulate cognitive functions such as decision-making and working memory. Anatomical and imaging multi-scale connectivity data is used to make whole-brain simulations from mouse, macaque to human more biologically plausible.

\subsection{Brain Cognitive Function Simulation}

\textbf{1. \emph{Drosophila}-inspired Decision-Making SNN}

\begin{figure}[htbp]
\begin{center}
\includegraphics[width=8cm]{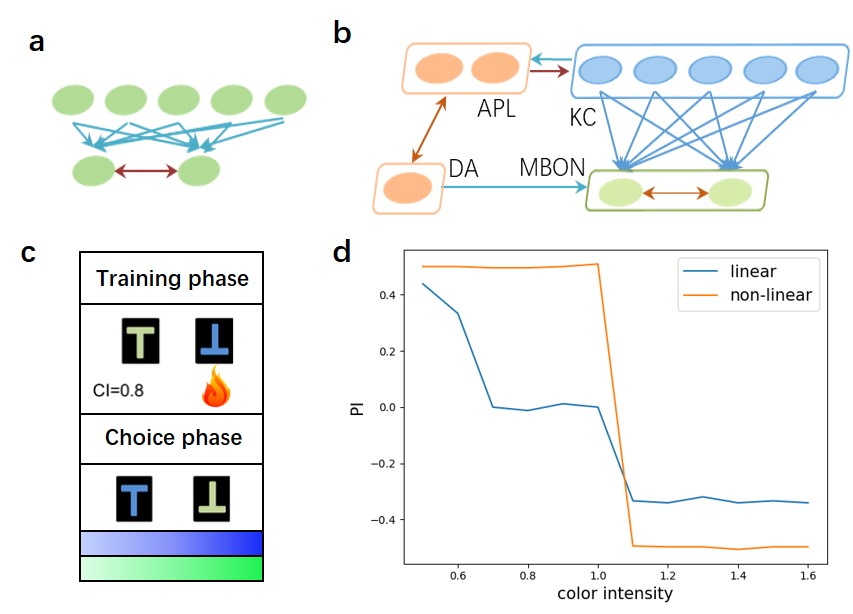}
\end{center}
\caption{(a) Linear Pathway. (b) Nonlinear Pathway. (c) Experiments for training and choice phases. (d) Experimental results of linear and nonlinear networks under the dilemma. The X-axis refers to the color density, and the Y-axis represents the PI values. Refined based on~\cite{zhao2020neural}.}
\label{pi}
\end{figure}

\emph{Drosophila} decision-making consists of value-based nonlinear decision and perception-based linear decision, where the nonlinear decision could help to amplify the subtle distinction between conflicting cues and make winner-takes-all choices~\cite{tang2001choice}. In this paper, the BrainCog framework is used to build \emph{Drosophila} nonlinear and linear decision-making pathways as shown in Fig.~\ref{pi}a-b. The entire model consists of a training phase and a testing phase as same as~\cite{zhao2020neural}. In the training phase, a two-layer SNN with LIF neurons is trained by reward-modulated STDP, which combines local STDP synaptic plasticity with global dopamine regulation. The training phase learns the safe pattern (upright-green T) and the punished pattern (inverted-blue T)~\cite{zhao2020neural}. Therefore, it is safe for green color and upright T shape factors, while blue color and inverted T shape are dangerous.

Two cues (color and shape) are restructured during the testing phase, requiring linear and nonlinear pathways to make a choice between inverted-green T and upright-blue T, as shown in Fig.~\ref{pi}c. The linear decision directly uses the knowledge acquired during the training phase to make decisions. The nonlinear network models the recurrent loop of the DA‑GABA‑MB circuit~\cite{tang2001choice,zhang2007dopamine,zhou2019suppression}: KC activates the anterior posterior lateral (APL) neurons, which in turn releases GABA transmitter to inhibit the activity of KC. KC also provides mushroom body output neuron (MBON) with exciting input in order to generate behavioral choices. When faced with conflicting cues, the level of DA increases rapidly and produces mutual inhibition with APL, thereby producing a disinhibitory effect on KC. The excitatory connection between DA and MBON also helps speed up decision-making.

To verify the consistency of \emph{drosophila}-inspired decision-making SNN with the conclusions from neuroscience~\cite{tang2001choice}, we count the behavior paradigm of our model under different color intensities over a period of time. First, we run the network for 500 steps to count the time $t_1$ of selecting behavior 1 (avoiding) and the time $t_2$ of selecting behavior 2 (approaching). Then we calculate prefer index (PI) values under different color intensity: $PI=\frac{\left | t_1-t_2 \right | }{\left | t_1+t_2 \right | } $. From Fig.~\ref{pi}d, we find that nonlinear circuits could achieve a gain-gating effect to enhance relative salient cue and suppress less salient cue, thereby displaying the nonlinear sigmoid-shape curve~\cite{zhao2020neural}. However, the linear network couldn't amplify the difference between conflicting cues, thus making an ambiguous choice (linear-shape curve)~\cite{zhao2020neural}. This work proves that drawing on the neural mechanism and structure of the nonlinear and linear decision-making of the \emph{Drosophila} brain, the brain-inspired computational model implemented by BrainCog could obtain consistent conclusions with the~\emph{Drosophila} biological experiment~\cite{tang2001choice}.

\textbf{2. PFC Working Memory }

Understanding the detailed differences between the brains of humans and other species on multiple scales will help illuminate what makes us unique as a species~\cite{zhang2021comparison}. The neocortex is associated with many cognitive functions such as working memory, attention and decision making~\cite{miller2000prefontral,nieder2003coding,
bishop2004prefrontal,koechlin2003architecture,
wood2003human}. Based on the human brain neuron database of the Allen Institute for Brain Science, the key membrane parameters of human neurons are extracted \footnote{http://alleninstitute.github.io/AllenSDK/cell\_types.html}. Different types of human brain neuron models and rodent neuron models are established based on adaptive Exponential Integrate-and-Fire (aEIF) model~\cite{Fourcaud2003How,brette2005adaptive}, which is supported by BrainCog.

We refined the model of a  single PFC proposed by Haas and colleagues \footnote{http://senselab.med.yale.edu/ModelDB/}~\cite{hass2016detailed}. Subsequently, a 6-layer PFC column model based on biometric parameters was established~\cite{shapson2021connectomic}. The pyramidal cells and interneurons were proportionally distributed from the literature~\cite{beaulieu1993numerical,defelipe2011evolution} and connected with different connection probabilities for different types of neurons based on previous studies~\cite{gibson1999two,hass2016detailed,
gao2003dopamine}. Firstly, the accuracy of information maintenance was tested on  rodent PFC network model.
%
%

\begin{figure}[!htbp]
\centering
\includegraphics[width=0.5\textwidth]{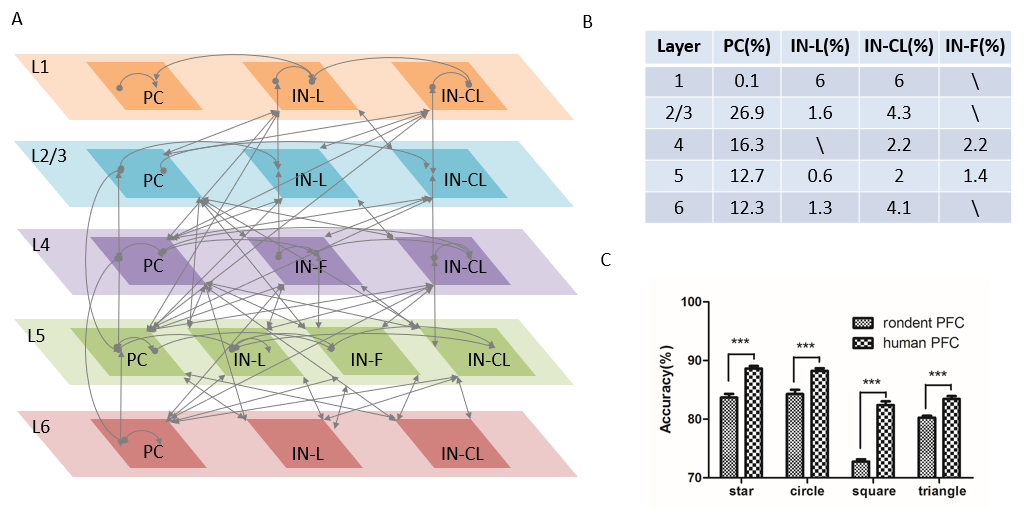}
\caption{Anatomical and network stimulation diagram. (a) The connection of a single PFC column. (b) The distribution proportion of different types of neurons in each column layer. (c) Network persistent activity performance. Refined based on~\cite{zhangqian2021comparison}.}
\end{figure}

Keeping the network structure and other parameters unchanged, only using human neurons to replace rodent neurons can significantly improve the accuracy and integrity of image output. From an evolutionary perspective, the lower membrane capacitance of human neurons facilitates firing. This change improves the efficiency of information transmission, which is consistent with the results of biological experiments~\cite{eyal2016unique}. This data-driven PFC column model provides an effective simulation-validation platform to study other high-level cognitive functions~\cite{2020Computational}.

\subsection{Multi-scale Brain Structure Simulation}

\textbf{1. Neural Circuit}

\subsubsection{Microcircuit}
\begin{figure}[htbp]
\begin{center}
\includegraphics[width=8cm]{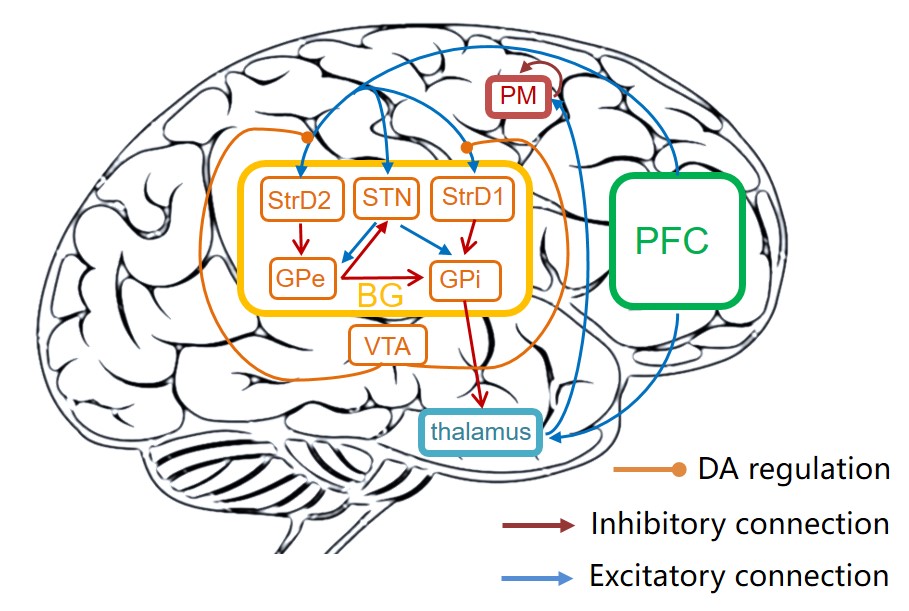}
\end{center}
\caption{The microcircuit of PFC-BG-ThA-PMC. Refined based on~\cite{zhao2018brain}.}
\label{dm}
\end{figure}

BrainCog implements a BDM-SNN model inspired by the decision-making neural circuit of PFC-BG-ThA-PMC in the mammalian brain (as shown in Fig.~\ref{dm})~\cite{zhao2018brain}. The BDM-SNN models the excitatory and inhibitory reciprocal connections between the basal ganglia nucleus~\cite{lanciego2012functional}: (1) Excitatory connections: STN-Gpi, STN-Gpe. (2) Inhibitory connections: StrD1-Gpi, StrD2-Gpe, Gpe-Gpi, Gpe-STN. Direct pathway (PFC-StrD1), indirect pathway (PFC-StrD2), and hyperdirect pathway (PFC-STN) from PFC to BG are further constructed. The output from BG transmits an inhibitory connection to the thalamus and finally excites PMC~\cite{parent1995functional}. In addition, excitatory connections are also formed between PFC and thalamus, and lateral inhibition exists in PMC. Such brain-inspired neural microcircuit, consisting of connections among different cortical and subcortical brain areas, and incorporating DA-regulated learning rules, enable human-like decision-making ability.

\subsubsection{Cortical Column}

\begin{figure}[!htbp]
\centering
\includegraphics[width=0.48\textwidth]{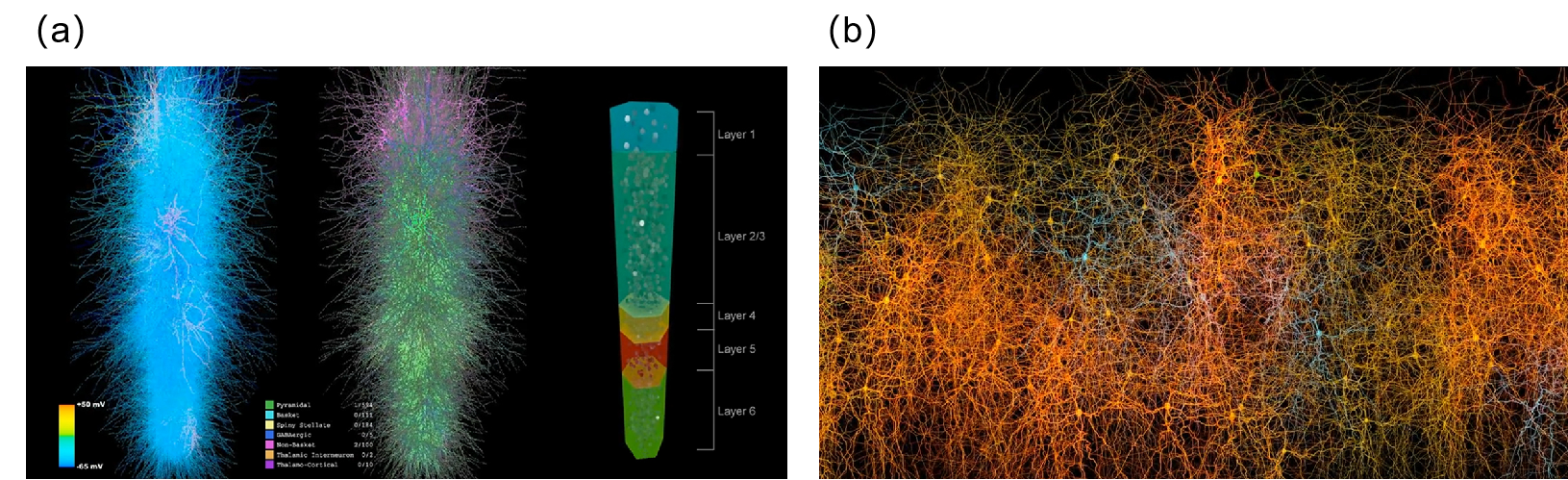}
\caption{The thalamocortical column. (a) shows the structure of the column and (b) describes the running activities of the unfold column when the neurons in Layer 4 receive the external stimulus.}
\label{minicolumn}
\end{figure}

A mammalian thalamocortical column is constructed in BrainCog, which is based on detailed anatomical data~\cite{Izhikevichlargescale}. This column is made up of a six-layered cortical structure consisting of eight types of excitatory and nine types of inhibitory neurons. Thalamic neurons cover two types of excitatory neurons, inhibitory neurons and GABAergic neurons in the reticular thalamic nucleus (RTN). Neurons are simulated by \emph{Izhikevich} model, which BrainCog applies to exhibit their specific spiking patterns depending on their different neural morphologies. For example, excitatory neurons (pyramidal and spiny stellate cells) always exhibit RS (Regular Spiking) or Bursting modes, while inhibitory neurons (basket and non-basket interneuron) are of FS (Fasting Spiking) or LTS (Low-threshold Spiking) patterns. Each neuron has a number of dendritic branches to accommodate a large number of synapses. The synaptic distribution and the microcircuits are reconstructed in BrainCog based on the previous works~\cite{Izhikevichlargescale, Binzegger2004A}. Fig.~\ref{minicolumn}(a) describes the details of the minicolumn. The column contains 1,000 neurons and more than 4,200,000 synapses. To understand the network further, we stimulate the spiny stellate cells in layer 4 to observe the activities of the whole network, Fig.~\ref{minicolumn}(b) shows the result of neural activities after the cells in layer 4 receive the external stimulation. 

\textbf{2. Mouse Brain}

The BrainCog mouse brain simulator is a spiking neural network model covering 213 brain areas of the mouse brain, which are classified according to the Allen Mouse Brain Connectivity Atlas~\cite{richardson2003subthreshold}~\footnote{http://connectivity.brain-map.org}. Each neuron was modeled by conductance-based spiking neuron model and simulated with a resolution of $dt$= 1 ms. A total of 6 types of neurons are included in this model, which are excitatory neurons (E), interneuron-basket cells (I-BC), interneuron-Matinotti cell (I-MC), thalamocortical relay neurons (TC), thalamic interneurons (TI) and thalamic reticular neurons (TRN).

$$
J \in\{E, I\underline{B} C, I\underline{M} C, T C, T I, T R N\}
$$

We use the aEIF neuron model referring to previous work~\cite{jiang2015principles,izhikevich2008large,tchumatchenko2014oscillations}, and obtain the parameters of this study, which are summarized in Tab.~\ref{mouse1}.

\begin{table}[h]
	\caption{Main parameters of different types of neuron models.}
	\centering
	\resizebox{0.8\linewidth}{!}{
		\begin{tabular}{lllllll}
			\toprule 
		 &$V_{th,j}(mV)$ &$ V_{r,j}(mV)$ & $\tau_{v,j}$ &$ \tau_{w,j}$ &$ \alpha _{j}$ & $\beta_{ j} $\\
			\midrule	
		 E & -50 & -110 & 100 & - & 0 & 0 \\
			
			\midrule
		 I-BC & -44 & -110 & 100 & 20 & -2 & 4.5 \\
		 \midrule
		  I-MC & -45 & -66 & 85 & 20 & -2 & 4.5 \\
		 
		  \midrule
		  TC & -50 & -60 & 200 & - & 0 & 0 \\
		   \midrule
		    TI & -50 & -60 & 20 & 20 & -2 & 4.5 \\
		      \midrule
		       TRN & -45 & -65 & 40 & 20 & -2 & 4.5 \\ 
			\bottomrule
		\end{tabular}
	}
	\label{mouse1}
	
\end{table}

The connections between brain areas are based on the quantitative anatomical dataset Allen Mouse Brain Connectivity Atlas. Methods for data generation have been previously described in~\cite{oh2014mesoscale}. The proportions of the different types of neurons were adopted from a previous study~\cite{izhikevich2008large, markram2004interneurons}.

The numbers of each type of neuron in the network are shown in Tab.~\ref{mouse2}.
\begin{table}[h]
	\caption{Number of different types of neurons in the BrainCog mouse brain simulator.}
    \centering
    \resizebox{0.8\linewidth}{!}{
		\begin{tabular}{lllllll}
			\toprule 
        Neuron Type & E & I\_BC & I\_MC & TC & TI & TRN \\ 
        	\midrule
        Neuron Number & 56100 & 14960 & 7480 & 1300 & 260 & 520 \\ 
        	\bottomrule
    \end{tabular}
    }
    \label{mouse2}
\end{table}

\begin{figure}
\centering
\includegraphics[width=0.48\textwidth]{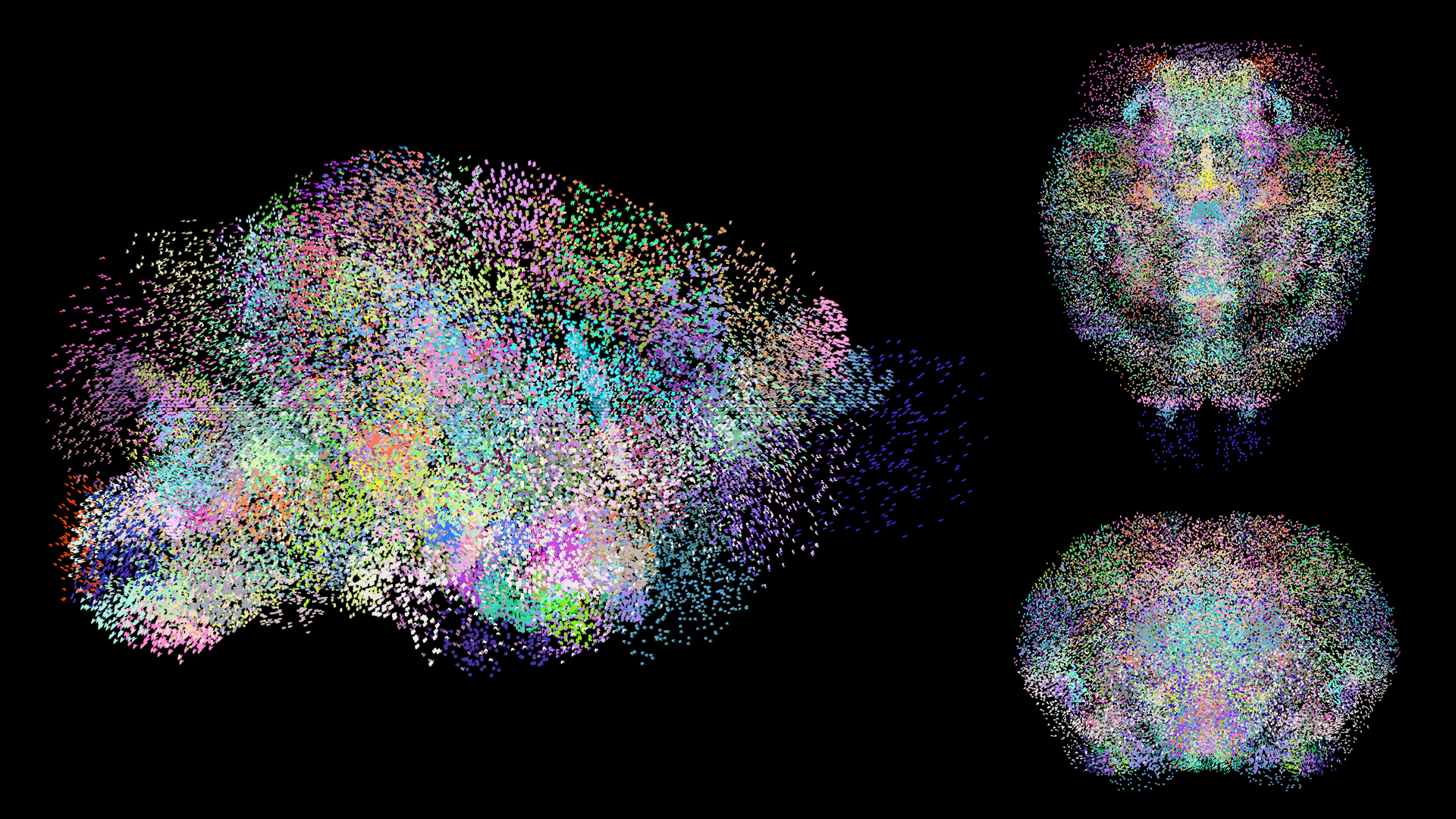}
\caption{Running of the BrainCog mouse brain simulator. The shining point is the spiking neuron at the time t and the point color represents the neuron belong to respective brain area.}
\label{figmouse1}
\end{figure}

The spontaneous discharge of the model without external stimulation is shown in Fig.~\ref{figmouse1}.
  
This is an open platform, and both the parameters of the neuron model and the number of different types of neurons can be set flexibly.

\textbf{3. Macaque Brain}

\begin{figure}[!htbp]
\centering
\includegraphics[width=0.48\textwidth]{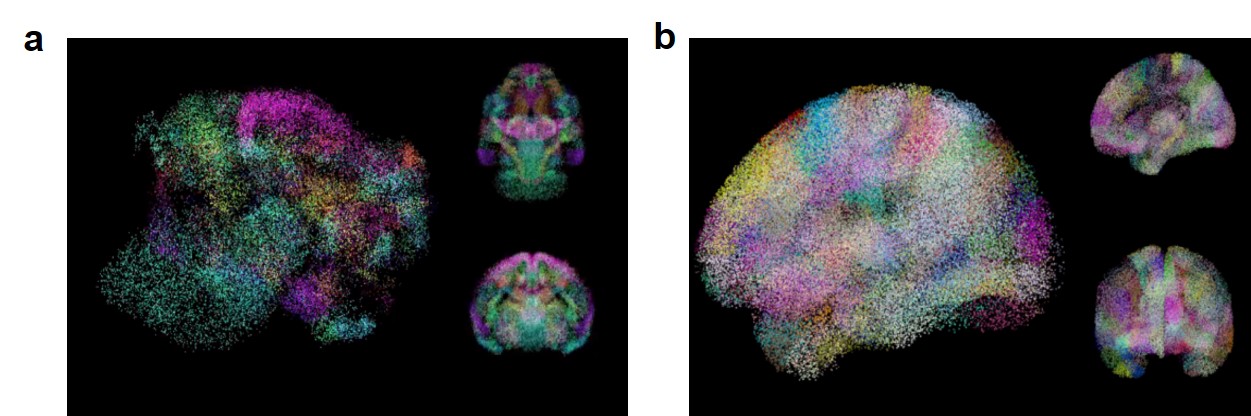}
\caption{Running of the macaque brain (a) and the human brain (b) model. The shining point is the spiking neuron at the time t and the point color represents the region which the neuron belongs to.}
\label{mac}
\end{figure}

The BrainCog macaque brain simulator is a large-scale spiking neural network model covering 383 brain areas~\cite{Dharmendra2010}. We used the multi-scale connectome transformation method~\cite{a2017Zhang} on the EGFP (enhanced green fluorescent protein) results~\cite{Bakker2012, Chaudhuri2015ALC, Christine2010} to obtain the approximate amount of cells per region and the approximate number of synaptic connections between two connected regions~\cite{Liu2016}. The final macaque model includes 1.21 billion spiking neurons and 1.3 trillion synapses, which is 1/5 of a real macaque brain. 
Specifically, the details of the brain micro-circuit are also considered in the simulation. The types of neurons in the micro-circuit include excitatory neurons (90 \% of the neurons are of this type in the simulation) and inhibitory neurons (10\% of the neurons are of this type in the simulation)~\cite{Davis1979}. The spiking neuron follows Hodgkin–Huxley model, which is supported by BrainCog. The running demo of the model is shown in Fig.~\ref{mac}(a). To use the macaque model in the platform, the parameters of the neuron number in each region, the connectome power between regions, and the proportion between the excitatory and inhibitory neurons can be set flexibly.

\textbf{4. Human Brain}

The BrainCog human brain simulator is built with the approach similiar to the BrainCog macaque brain. By using the EGFP results of the
human brainnetom atlas~\cite{Fan2016, Klein2012}, the BrainCog human brain simulator consists of 246 brain areas. It should be noted that since there is no directed human brain connectome available until the release of this paper, the BrainCog human brain simulator keeps bidirectional connections among brain areas. The details of the micro-circuit, including the excitatory neuron and the inhibitory neuron, are also considered. The final model (Fig.~\ref{mac}(b) includes 0.86 billion spiking neurons and 2.5 trillion synapses, which is 1/100 of a real human brain. To use this model in the platform, the neuron number per region, the connectome power, and the proportion between the excitatory and inhibitory neurons can be set flexibly.
Morover, all the simulations were performed on distributed memory clusters~\cite{Liu2016} at the super computing center affiliated to Institute of Automation, Chinese Academy of Sciences, Beijing, China. The cluster named the fat cluster is composed of 16 blade nodes and 2 “fat” computing nodes. In order to imporve the network communication efficiency and the simulation performance, the most connected areas were simulated in the 'fat' computing nodes to minimize the inter-node communications, while the other areas is randomly distributed in the blade nodes. The simulation shows the ability of the framework to deploy on the supercomputer or large-scale computer clusters.

\section{BORN: A spiking neural network driven Artificial Intelligence Engine based on BrainCog}

\begin{figure*}[htb]
\centering
\includegraphics[width=1\textwidth]{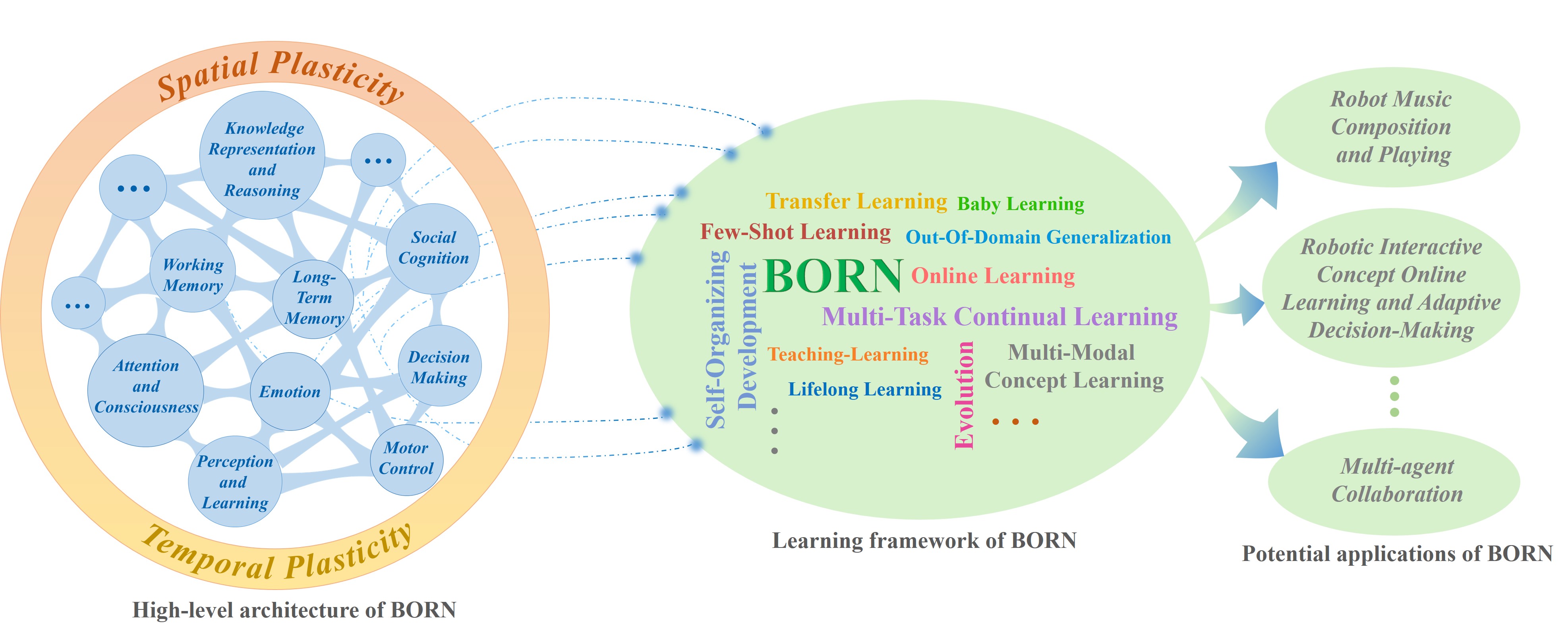}
\caption{The functional framework and vision of BORN.}
\label{bornvision}
\end{figure*}

BrainCog is designed to be an open source platform to enable the community to build spiking neural network based brain-inspired AI models and brain simulators. Based on the essential components developed for BrainCog, one can develop their own domain specific or general purpose AI engines. To further demonstrate how BrainCog can support building Brain-inspired AI engine, here we introduce BORN, an ongoing SNN driven Brain-inspired AI engine, ultimately designed for general purpose living AI. As shown in Fig.~\ref{bornvision}, the high-level architecture of BORN is to integrate spatial and temporal plasticities to realize perception and learning, decision-making, motor control, working memory, long-term memory, attention and consciousness, emotion, knowledge representation and reasoning, social cognition and other brain cognitive functions. Spatial plasticity incorporates multi-scale neuroplasticity principles at micro, meso and macro scales. Temporal plasticity considers learning, developmental and evolutionary plasticity at different time scales.

As an essential component for BORN, we propose a developmental plasticity-inspired adaptive pruning (DPAP) model, that enables the complex deep SNNs and DNNs to gradually evolve into a brain-inspired efficient and compact structure, and eventually improves learning speed and accuracy in the extremely compressed networks~\cite{Han2022Developmental}. The evolutionary process for the brain includes but not limited to searching for the proper connectome among different building blocks of the brain at multiple scales (e.g. neurons, microcircuits, brain areas). BioNAS for BORN uses brain-inspired neural architecture search to construct SNNs with diverse motifs in the brain and experimentally verify that SNNs with rich motif types perform better than plain feedforward SNNs~\cite{shenBrainInspired2022}.

How the human brain selects and coordinates various learning methods to solve complex tasks is crucial for understanding human intelligence and inspiring future AI. BORN is dedicated to address critical research issues like this. The learning framework of BORN consists of multi-task continual learning, few-shot learning, multi-modal concept learning, online learning, lifelong learning, teaching-learning, and transfer learning, etc. 

To demonstrate the ability and principles of BORN, we provide a relatively complex application on emotion dependent robotic music composition and playing. This application requires a humanoid robot perform music composition and playing depending on visual emotion recognition. The application requires BORN to provide cognitive functions such as visual emotion recognition, sequence learning and generation, knowledge representation and reasoning, and motor control, etc. This application of BORN starts with perception and learning, and ends with motor output. 

It includes three modules implemented by BrainCog: the visual (emotion) recognition module, the emotion-dependent music composition module, and the robot music playing module. As shown in Fig.~\ref{rmp}, the visual emotion recognition module enables robots to recognize the emotions expressed in images captured by the humanoid robot eyes. The emotion-dependent music composition module can generate music pieces according to various emotional inputs. When a picture is shown to the robot, the Visual Emotion Recognition network can firstly identify the emotions expressed in the picture, such as joy or sadness. The robot then selects or compose the music piece that best matches the emotions in the picture. And finally, with the help of the robot music playing module, the robot controls its arms and fingers in a series of movements, thus playing the music on the piano. Some details are introduced as follows:

\begin{figure*}[!htbp]
\centering
\includegraphics[width=1\textwidth]{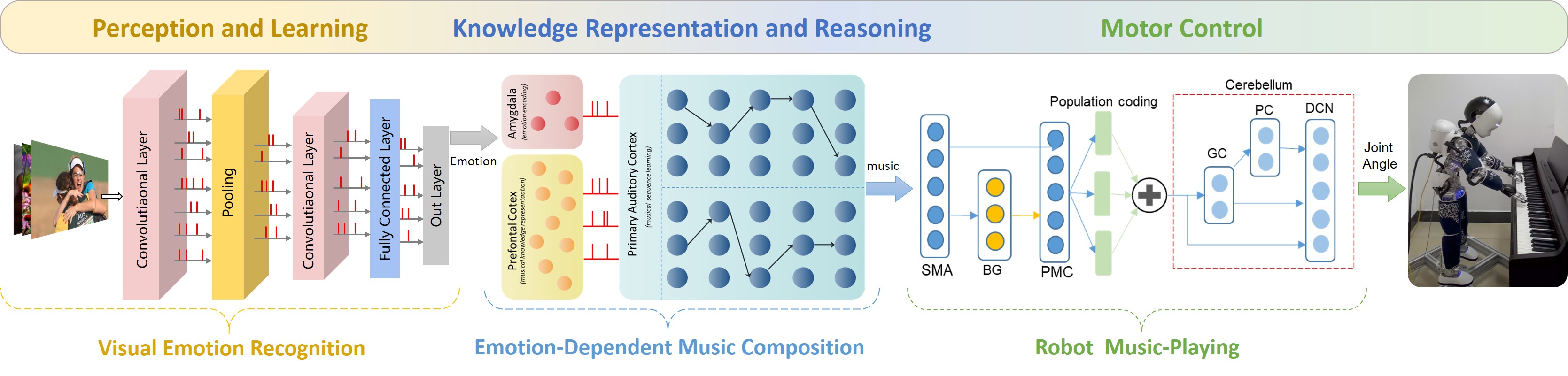}
\caption{The procedure of multi-cognitive function coordinated emotion dependent music composition and playing by humanoid robot based on BORN.}
\label{rmp}
\end{figure*}

\emph{1) Visual Emotion Recognition: }
For emotion recognition, inspired by the ventral visual pathway, we construct a deep convolutional spiking neural network with LIF neuron model and surrogate gradient provided by BrainCog. The structure of the network is set as 32C3-32C3-MP-32C3-32C3-300-7. 32C3 means the output channel is set with 32, and the kernel size is set as 3. MP denotes the max pooling. The mean firing rate is used to make the final prediction. We use the Adam optimizer, and the mean square error loss. The initial learning rate is set with 0.001, and it will decay to 1/10 of the previous value every 40 epochs, for a total of 100 epochs. We use the Emotion6 dataset~\cite{peng2015mixed} to train and test our model. The Emotion6 dataset is composed of 6 emotions such as anger, disgust, fear, joy, sadness, surprise, and each type of emotion consists of 330 samples. On this basis, we extend the original Emotion6 dataset with exciting emotion which we collect online. 80\% of the images are used as the training set, and the remaining 20\% are used as the test set. 

\emph{2) Emotion-dependent Music Composition: }
Listening to the the music can make us emotional, while, when people feel happy or sad, they always express their feeling with music. Amygdala plays a key role in human emotion. Inspired by this mechanism, we constructed a simple spiking neural network to simulate this important area to represent different types of emotions and learn the relationships with other brain areas related to music. As shown in Fig.~\ref{rmp}, the amygdala network is composed of several LIF neurons supported through BrainCog, the connections from this cluster are projected to the musical sequence memory networks. 

During the learning process, amygdala, PFC, and the musical sequence memory networks cooperate with each other and form complex neural circuits. Here, connections are updated by the STDP learning rule. The dataset used here also contains 331 MIDI files of classical piano works~\cite{Krueger2018}, and it is important to note that a part of these music works are labeled with different types of emotional categories (such as happy, depressed, passionate and beautiful). 

At the generation process, given the beginning notes and specific emotion type, the model can generate a series of notes and form a melody with particular emotion finally.

\emph{3) Robot Music-Playing: }
A humanoid robot iCub is used to validate the abilities of robotic music composition and playing depending on the result of visual emotion recognition. The iCub robot has a total of 53 degrees of freedoms throughout the body. In the piano playing task, we used 6 degrees of freedoms of the head, 3 degrees of freedoms of the torso, and 16 degrees of freedoms for each of the left and right arm (including the left and right hand). Besides, we mainly control the index fingers to press the keys; in the multi-fingered playing mode, we mainly control the thumbs, the index fingers, and the middle fingers to press the keys. During playing, the robot controls the movement of the hand in sequence according to the generated sequence of different musical notes, and presses the keys with corresponding fingers, thereby completing the performance. For each note to be played, the corresponding playing arm needs to complete the entire process of moving, waiting, pressing the key, holding, and releasing the key according to the beat. During the playing process, we also control the movements of the robot's head and the non-playing hand to match the performance.

We have constructed a multi-brain area coordinated robot motor control SNN model based on the brain motor control circuit. The SNN model is built with LIF neurons and implements SMA, PMC, basal ganglia and cerebellum functions. The  music notes is first processed by SMA, PMC and basal ganglia networks to generate high-level target movement directions, and the output of PMC is encoded by population neurons to target movement directions. The fusion of population codings of movement directions is further processed by the cerebellum model for low level motor control. The cerebellum SNN module consists of Granular Cells (GCs), Purkinje Cells (PCs) and Deep Cerebellar Nuclei (DCN), which implements the three level residual learning in motor control. The DCN network generates the final joint control outputs for the robot arms to perform the playing movement.

\section{Conclusion}
BrainCog aims to provide a community based open source platform for developing spiking neural network based AI models and cognitive brain simulators. It integrates multi-scale biological plausible computational units and plasticity principles. Different from existing platforms, BrainCog incorporates and provides task ready SNN models for AI, and supports brain function and structure simulations at multiple scales. With the basic and functional components provided in the current version of BrainCog, we have shown how a variety of models and applications can already be implemented for both brain-inspired AI and brain simulations. Based on BrainCog, we are also committed to building BORN into a powerful SNN-based AI engine that incorporates multi-scale plasticity principles to realize brain-inspired cognitive functions towards human level. Powered by 9 years development of BrainCog modules, components and applications, and inspired by biological mechanisms and natural evolution, continuous efforts on BORN will enable it to be a general purpose AI engine. We have already started the efforts to extend BrainCog and BORN to support high-level cognition such as theory of mind~\cite{zhao2022brain}, consciousness~\cite{zeng2018toward}, and morality~\cite{zhao2022brain}, and it definitely takes the world to build true and general purpose AI for human and ecology good. Join us on this explorations to create the future for human-AI symbiotic society. 
\section*{Acknowledgements}
This work is supported by the National Key Research and Development Program (Grant No. 2020AAA0104305), the Strategic Priority Research Program of the Chinese Academy of Sciences (Grant No. XDB32070100).

\bibliographystyle{IEEEtran}
\bibliography{IEEEabrv,mybibfile}



%




\end{document}